%% file: main.tex
\documentclass[]{x2lab}
\usepackage{csquotes}
\usepackage{afterpage}
\usepackage{amsmath,amssymb,amsfonts}
\usepackage{mathtools}
\usepackage{amsthm}
\usepackage{xspace}
\usepackage{colortbl}
\usepackage{enumitem}
\usepackage{wrapfig}
\usepackage{nicefrac}
\usepackage{makecell}
\usepackage{pifont}
\usepackage{svg}
\usepackage{array}
\usepackage{tabularx}
\usepackage{longtable}
\usepackage{xltabular}
\usepackage{float}
\usepackage{placeins}
\usepackage[utf8]{inputenc}
\usepackage{times}
\usepackage{booktabs,multirow,graphicx}

\graphicspath{{figures/}{figs/}}

\definecolor{yamlBlue}{RGB}{75,133,249}    
\definecolor{yamlIndigo}{RGB}{125,113,230}  
\definecolor{yamlMagenta}{RGB}{185,95,199}  
\definecolor{yamlPink}{RGB}{236,83,165}     
\definecolor{yamlPunct}{RGB}{90,90,110}     
\definecolor{yamlFrame}{RGB}{223,226,235}   
\definecolor{yamlText}{RGB}{40,42,54}       
\lstdefinelanguage{yaml}{
  keywords={true,false,null,yes,no,on,off},
  keywordstyle=\color{yamlMagenta}\bfseries,
  sensitive=false,
  comment=[l]{\#},
  commentstyle=\color{yamlPink}\itshape,
  morestring=[b]",
  morestring=[b]',
  stringstyle=\color{yamlIndigo},
  literate=%
    {:}{{{\color{yamlPunct}:}}}1
    {-}{{{\color{yamlPunct}-}}}1
    {\{}{{{\color{yamlPunct}\{}}}1
    {\}}{{{\color{yamlPunct}\}}}}1
    {[}{{{\color{yamlPunct}[}}}1
    {]}{{{\color{yamlPunct}]}}}1
    {,}{{{\color{yamlPunct},}}}1
    {@}{{{\color{yamlBlue}@}}}1,
}
\definecolor{fbApp}{HTML}{c8e7fa}
\definecolor{fbPurple3}{HTML}{f0ebf5}
\definecolor{citecolor}{HTML}{0071BC}
\definecolor{linkcolor}{HTML}{ED1C24}

\titleformat*{\paragraph}{\rmfamily\bfseries}

\title{\begin{center}
X2Real Technical Report\\[-0.5cm]
{\large\mdseries\itshape An e\textbf{X}tensive simulation benchmark for \textbf{real}-world generalist policies}
{\author{X Square Robot}}
\end{center}}

\abstract{Generalist robot manipulation policies have developed rapidly, yet their reliable evaluation remains challenging due to fundamental flaws in existing simulation benchmarks: prominent sim-to-real gaps, narrow task coverage, and unfair evaluation caused by ambiguous training-test pipelines. Prior works only partially resolve these issues and lack simultaneous faithfulness, diversity, and fairness, while static benchmark designs fail to sustain long-term policy development. We presents X2Real, an evolvable simulation benchmark for faithfully evaluating the real-world performance of robotic manipulation policies based on Nvidia Isaac Lab-Arena. Following three core principles—faithfulness, diversity, and fairness—X2Real calibrates simulation visual and physical properties to align with real hardware, achieving a 0.84 linear correlation between simulated and real-robot evaluation results. It features a comprehensive taxonomy with 10 capability dimensions and 44 hierarchical long-horizon tasks, covering basic manipulation skills and advanced capacities such as visual grounding, language understanding, and bimanual control. We further adopt multi-axis domain randomization and strictly disjoint training-evaluation pipelines to mitigate benchmark exploitation and ensure credible evaluation. Powered by a custom physical domain-specific language, the Mana simulation ecosystem supports modular task design and iterative performance analysis, alongside a nearly 300-hour annotated simulation trajectory dataset. X2Real offers a faithful, diverse, and fair evolving evaluation infrastructure, effectively bridging the sim-to-real evaluation gap and supporting the advancement of generalist robotic manipulation policies.}

\metadata[Code]{\url{https://github.com/X-Square-Robot/x2real}}
\metadata[Project Page]{\url{https://x2robot.com/en/pages/x2real}}

\input{sec/assets_scene/preamble}

\usetikzlibrary{shadows}

\begin{document}
\maketitle

\input{sec/1_intro}
\input{sec/2_benchmark}

\input{sec/3_platform}
\input{sec/5_experiments}

\input{sec/6_related}

\input{sec/7_discussion}

\input{sec/contributors}

\clearpage
\newpage
\bibliographystyle{unsrtnat}
\bibliography{main,sec/assets_scene/references}

\clearpage
\appendix
\FloatBarrier
\input{sec/appendix}

\end{document}

%% file: sec/assets_scene/preamble.tex
\usepackage{tikz}
\usetikzlibrary{arrows.meta,fit,positioning,shapes.geometric}
\newcolumntype{Y}{>{\raggedright\arraybackslash}X}

%% file: sec/1_intro.tex
\section{Introduction}\label{sec:intro}

\begin{figure}[th]
    \centering
    \includegraphics[width=\linewidth]{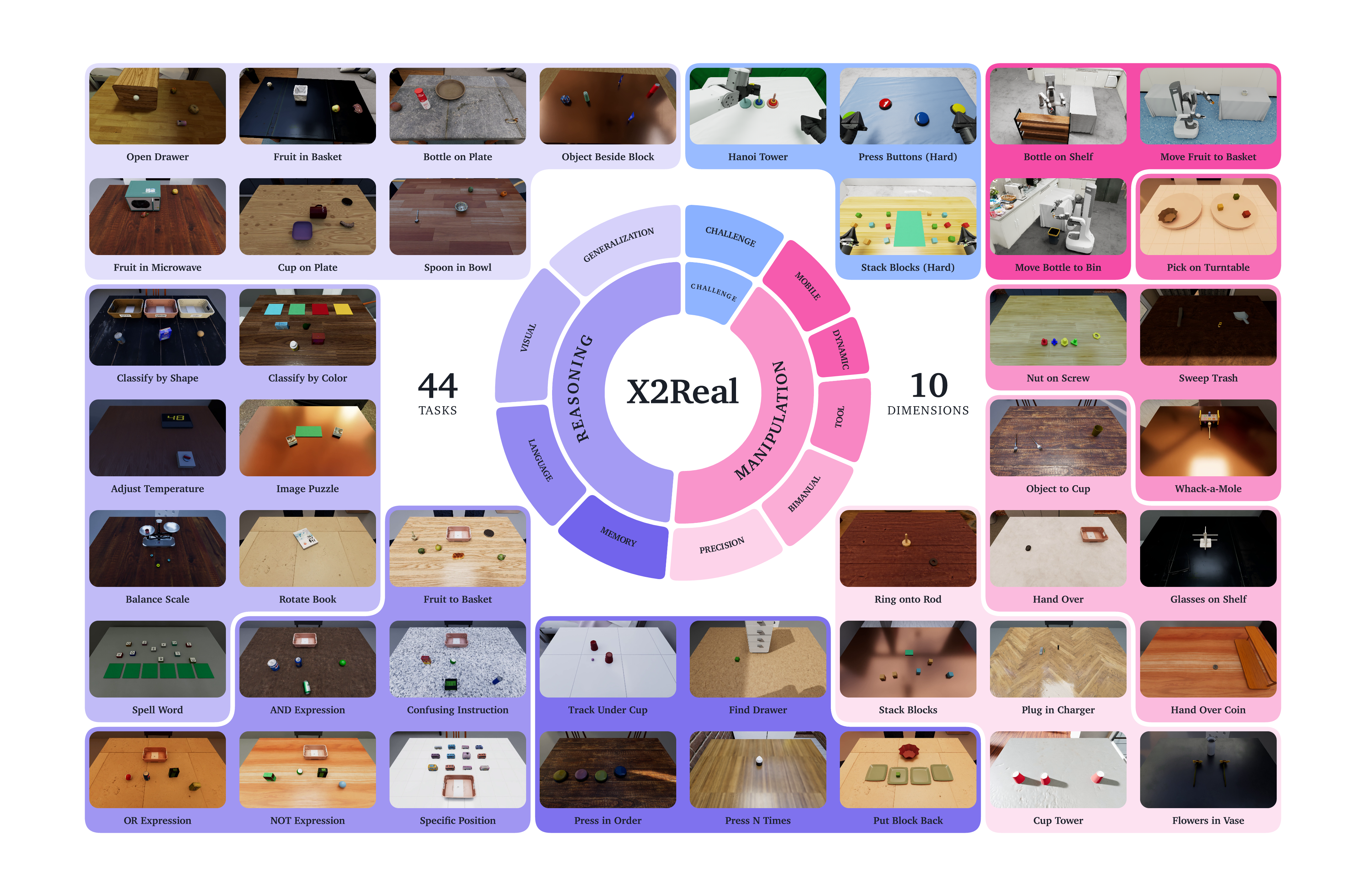}
    \caption{\textbf{X2Real Task Taxonomy.} X2Real features a comprehensive task taxonomy spanning 10 core capability dimensions and 44 hierarchical long-horizon tasks. It leverages full multi‑axis domain randomization with strictly disjoint training‑evaluation pipelines to remove benchmark biases and enhance real‑to‑sim generalization fidelity.}
    \label{fig:task-distribution}
\end{figure}

As generalist robot manipulation policies like RT-2~\cite{zitkovich2023rt2}, $\pi_{0.5}$~\cite{black2025pi05}, DreamZero~\cite{ye2026dreamzero}, wall-oss-0.5~\cite{yu2026walloss05}, wall-wm~\cite{li2026wallwm}, Cosmos3~\cite{nvidia2026cosmos3} advance rapidly, evaluating what they can actually do becomes increasingly important and difficult.
Real-robot evaluation~\cite{atreya2025roboarena,sun2026maniparena} is the gold standard, but it is costly, slow to yield feedback, and hard to reproduce.
Simulation benchmarks sidestep these problems by running robot–environment interactions in parallel, at low cost and with full reproducibility.
Yet three caveats keep current simulation benchmarks from serving as a faithful proxy for real-world ability.

{\bf Sim2real gap.}
A simulation score is faithful only if it predicts real-robot performance, yet discrepancies in visual appearance, physical dynamics, and robot control create a gap between the two, a.k.a, the simulation-to-real gap (sim2real gap).
More specifically, we care about the linear correlation between the scores of the same model evaluated in simulation and on a real robot; ideally, this correlation should approach $1.0$.
Without such predictivity, a high simulation score may reflect only how well a policy exploits the simulator, rather than how well it will perform once deployed.

{\bf Narrow evaluation distribution.}
A simulation task is inherently multi-dimensional: it determines which policy capabilities are exercised, which atomic manipulation skills are required, and—crucially—whether it probes a policy's true competence or merely its (over)fitting to a fixed setup.
For an evaluation to effectively reflect a policy's capability, its tasks must span a diverse distribution along every such dimension: assets, layouts, instructions, task designs, and so on.
Constrained by the availability of assets, simulators, and demonstration data, however, prior benchmarks typically cover only a handful of largely pick-and-place tasks built on a fixed set of assets.
Such a scope was adequate for early policies, but these benchmarks saturate quickly as policies improve.

{\bf Benchmark exploitation.}
Ideally, a policy would be evaluated fully zero-shot, with every task unseen.
In practice, current policies still generalize poorly zero-shot and typically require some post-training to perform these tasks, so a benchmark must also supply fine-tuning data.
This calls for a strict separation between the data used for training and the setup used for evaluation. Such separation is hard to maintain in simulation: because a simulated environment is deterministic and easily reproduced, a policy can be optimized toward the specific evaluation setup, and in the extreme the leaderboard can be gamed by fixing random seeds or memorizing particular scenes.
A narrow task distribution only widens this room for exploitation.

Together, these three issues—the sim2real gap, narrow evaluation distribution, and benchmark exploitation—determine whether a simulated evaluation can reflect a policy's capability in a faithful, effective, and fair manner, respectively.
Prior work has largely addressed only one of them at a time, directly or indirectly. SIMPLER studies sim–real correspondence through paired simulation
and real-robot evaluation~\cite{li2025simpler}, followed by PolaRis~\cite{jain2026polaris} and REALM~\cite{sedlacek2026realm} and other works, but their task suites cover only a limit distribution of robotic capabilities. BEHAVIOR-1K~\cite{li2023behavior1k}, Robocasa365~\cite{nasiriany2026robocasa365},  Robotwin2.0~\cite{chen2025robotwin2} and many other works cover a wide range of tasks, but do not ground their simulation score on the real performance. Recently as our concurrent work, Robodojo~\cite{chen2026robodojo} provides a comprehensive sim-and-real benchmark on a diverse task distribution, but regrettably does not include an explicit sim2real analysis.
Moreover, even once these issues are resolved, any static benchmark will eventually saturate as policies improve; what is ultimately needed is an infrastructure that can continually evolve its environments and tasks to serve the long-term development of embodied intelligence.

To tackle these problems, we propose {\bf X2Real}, an e\textbf{X}tensive simulation benchmark aimed to test policies' capabilities under \textbf{real‑world} deployment. Guided by three core design principles—faithfulness, diversity, and fairness—X2Real mitigates the aforementioned limitations through hardware‑aligned simulation calibration, a systematic capability taxonomy, comprehensive domain randomization, and a strictly separated training‑evaluation data pipeline.

To reduce the sim2real gap, we calibrate multiple robot embodiments visually and physically, and optimize low‑level control parameters to achieve millimeter‑level trajectory‑replay consistency against real‑world hardware. We achieve an $\mathbf{0.84}$ linear correlation coefficient for simulation and reality evaluation scores on a model trained only on real data. Moving beyond the limited scope of conventional pick‑and‑place benchmarks, X2Real covers hierarchical reasoning and manipulation skills including visual perception, language grounding, memory, and precise bimanual control. In total, the benchmark comprises 44 simulation tasks organized across 10 capability dimensions, as shown in Fig.~\ref{fig:task-distribution}. These tasks encompass fundamental atomic skills such as pick, place, twist, push, and pull, while further increasing difficulty by composing these primitives into long‑horizon challenging sequences. 

\begin{figure}[t]
    \centering
    \includegraphics[width=\linewidth]{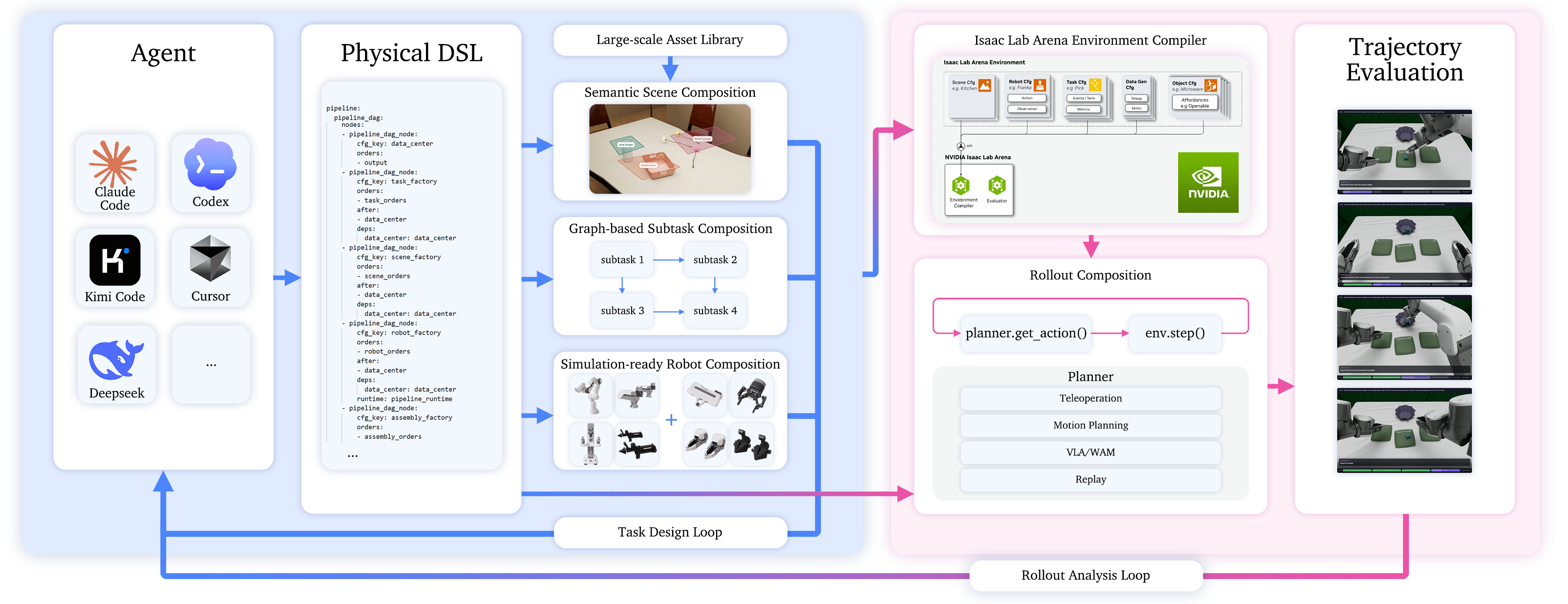}
    \caption{\textbf{Mana System Overview.} Physical DSL enables agents to drive task design and rollout analysis in closed loops, from semantic scene, subtask, and robot composition to environment compilation, policy rollout, and trajectory evaluation.}
    \label{fig:pipeline}
\end{figure}

Unlike previous benchmarks that mostly only perform visual randomization, our framework enables task-customized randomization across three orthogonal axes: \textbf{appearance, task setting, and embodiment}. This design ensures that evaluation results reliably reflect the policy’s true manipulation capabilities. To counter benchmark exploitation, we enforce strict separation between training and evaluation environments. All demonstration trajectories are collected within a green‑booth setup adapted from ManipArena~\cite{sun2026maniparena} with constrained randomization, which is kept distinct from evaluation configurations. Via teleoperation and automatic synthesis, we assemble a high‑quality dataset of nearly \textbf{300 hours} of simulation trajectories, annotated with frame‑level subtask progress labels.

To realize X2Real’s demanding requirements for high task diversity, physical faithfulness, we construct an agent‑driven unified simulation ecosystem \textbf{Mana} based on Nvidia Isaac Lab-Arena, depicted in Fig.~\ref{fig:pipeline}. At its core lies our physical domain-specific language (Physical DSL), a declarative abstraction layer that unifies the specifications of scenes, robot embodiments, subtask dependency graphs, and multi-axis domain-randomization rules into modular, reusable primitives. It enables the agent to interact with the underlying simulation stack in a safer, more controllable, and fully declarative fashion. The DSL drives two main loops: the \textbf{Task Design Loop} for benchmark authoring and the \textbf{Rollout Analysis Loop} for rollout execution and iterative refinement.
Within the Task Design Loop, scene, subtask‑graph and robot‑embodiment definitions are assembled and compiled into runnable simulation environments via the Isaac Lab Arena Environment Compiler.
In the Rollout Analysis Loop, multiple execution backends generate frame‑annotated trajectories; evaluation signals flow back to agents to analysis model performance. These together make X2Real an evolving benchmark for future development of generalist policies.

To support standardized evaluation across heterogeneous policy architectures, X2Real further introduces \textbf{Policy Space}, a unified runtime interface that decouples policy-specific inference pipelines from the simulation and evaluation infrastructure. Policy Space standardizes observation mapping, action adaptation, inference scheduling, history and internal-state updates, and episode lifecycle management, while allowing each policy to retain its native preprocessing and action representation. This abstraction enables policies with substantially different runtime requirements to be evaluated under exactly the same task definitions, observations, randomization rules, and evaluation criteria. Using this protocol, we benchmark four representative fine-tuned generalist policies across the full X2Real task suite. The results reveal substantial capability differences across reasoning and manipulation dimensions, as well as a pronounced degradation from in-distribution (ID) to out-of-distribution (OOD) evaluation. For example, even the strongest evaluated policy decreases from an overall ID success rate of \textbf{53.9\%} to an OOD success rate of \textbf{34.3\%}, with particularly large gaps on tasks requiring generalization, language reasoning, memory tracking, and fine-grained manipulation. These results demonstrate that current generalist policies remain far from saturating X2Real and highlight the importance of evaluating not only task completion, but also robustness across diverse, unseen deployment conditions.

%% file: sec/2_benchmark.tex
\section{Benchmark Overview}\label{sec:benchmark}

\subsection{Multi-dimensional Task Suite}

We systematically evaluate the overall competencies of generalist robotic manipulation policies by dividing policy behaviors into two canonical, complementary dimensions: Reasoning, which governs task decision-making (i.e., what to do), and Manipulation, which governs physical execution (i.e., how to do it). The overall taxonomy is visualized in Fig.~\ref{fig:task-distribution}. The Reasoning dimension further decomposes into four standardized sub-capabilities:
\begin{enumerate}
    \item Generalization: It characterizes the ability of a policy to stably perform basic task operations under diverse environmental perturbations.
    \item Visual Understanding: It characterizes the ability of a policy to perceive and distinguish visual task cues, including object shapes, colors, and numbers, etc.
    \item Language Understanding: It characterizes the ability of a policy to accurately follow structured linguistic instructions, such as parsing logical descriptions and interpreting row-column positional specifications.
    \item Memory: It characterizes the ability of a policy to retain and utilize historical information for completing context-dependent sequential tasks.
\end{enumerate}
Accordingly, the Manipulation dimension consists of five execution-oriented core capabilities:
\begin{enumerate}
    \item Precision Operation: It characterizes the ability of a policy to control robot end-effectors for stable, high-precision motion execution.
    \item Bimanual Coordination: It characterizes the ability of a policy to implement collaborative motion control for dual robotic arms.
    \item Tool Usage: It characterizes the ability of a policy to operate daily functional tools (e.g., hammers, brooms, screws) to satisfy task demands.
    \item Dynamic Operation: It characterizes the ability of a policy to interact steadily and effectively with moving objects.
    \item Mobile Operation: It characterizes the ability of a policy to accomplish manipulation tasks that couple mobile-base navigation with object interaction across spatially separated workspaces.
\end{enumerate}

Beyond standard capability assessment, we design three challenging tasks to probe the performance ceiling and quantify the upper competence bound of generalist manipulation policies:
\begin{enumerate}
    \item Stack Blocks (Hard): The robot stacks 15 uniformly sized blocks into the tallest feasible structure. This task evaluates the upper precision limit of policy motion execution.
    \item Press Buttons (Hard): The robot presses buttons strictly according to a predefined color sequence with more than ten ordered cues. This task examines the upper bound of policy long-term memory capacity.
    \item Hanoi Tower: The robot solves the classic 5-layer Tower of Hanoi problem. This task assesses the high-level reasoning and sequential planning capabilities of the policy.
\end{enumerate}

Our full task suite is documented in the Appendix~\ref{app:task-details}. These tasks span diverse low‑level atomic manipulation skills, such as picking, placing, twisting, pushing, and pulling. All tasks decompose into hierarchical subtasks that form a Directed Acyclic Graph (DAG). A rule-based evaluator determines the success of each individual subtask. This design supports fine-grained episode-level progress monitoring in addition to conventional final success rate evaluation. Detailed configurations of the task construction pipeline are provided in Sec.~\ref{sec:task-skill}. 

\begin{figure}[H]
    \centering
    \includegraphics[width=0.8\linewidth]{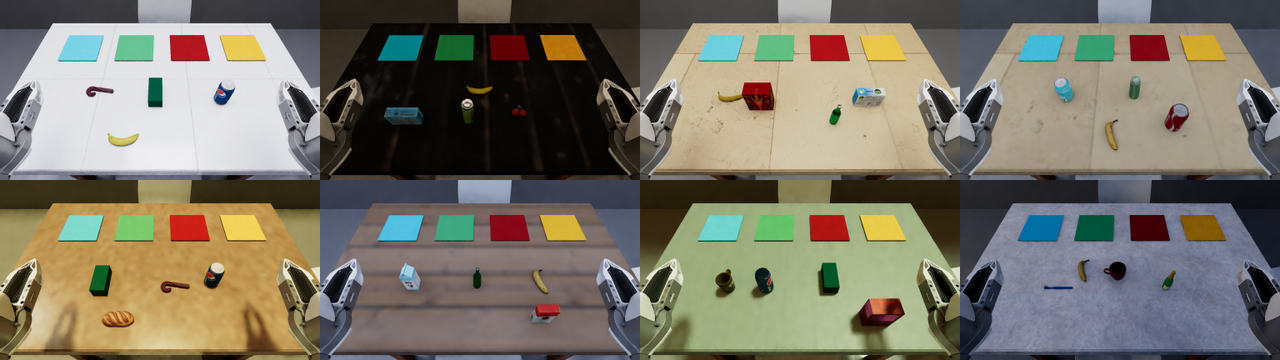}
    \caption{\textbf{Domain Randomization Example.} The lighting condition, table material, task assets can be randomized for the Classify by Color task.}
    \label{fig:domain_randomization}
\end{figure}

\subsection{Diverse Evaluation Distribution}

To mitigate the narrow evaluation distribution and benchmark exploitation limitations prevalent in existing simulation benchmarks, we equip all designed tasks with multi-dimensional domain randomization. Our randomization scheme falls into three orthogonal axes to guarantee diverse, robust, and unbiased evaluation:
\begin{enumerate}
    \item Appearance: It covers background scenes, lighting conditions, and table surface textures.
    \item Task Setting: It covers task assets, object placement, distractor layout, linguistic instructions, and puzzle solution configurations.
    \item Embodiment: It covers gripper types, camera intrinsic and extrinsic parameters, and table heights.
\end{enumerate}
This multi-axis diversity design ensures that each task reliably targets its corresponding policy capability and avoids overfitting to fixed simulation configurations. We take the "Classify by Color" task as a representative case, as shown in Fig.~\ref{fig:domain_randomization}: the robot places red, blue, green, and yellow objects onto color-aligned target papers. By dynamically randomizing scene appearance and object assets, the task strictly verifies visual understanding robustness. Policies must extract discriminative color features for zero-shot classification instead of relying on superficial cues such as object shape, function, or placement position.

We adopt hierarchical randomized settings for category-specific evaluation. Generalization-oriented tasks enable all three axes of randomization; other Reasoning tasks adopt appearance and task-setting randomization; Manipulation tasks adopt appearance-only randomization. This stratified design balances evaluation diversity and task specificity.

\begin{figure}[H]
    \centering
    \includegraphics[width=0.8\linewidth]{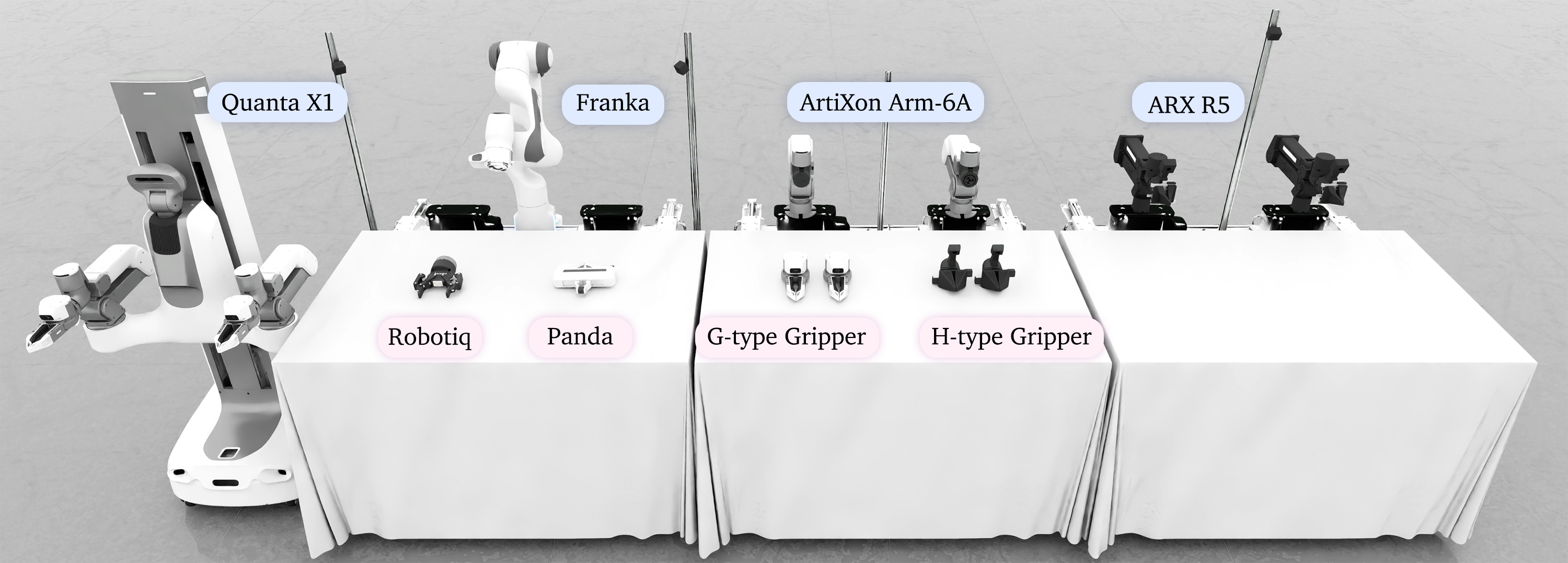}
    \caption{\textbf{X2Real Embodiments.} X2Real includes mobile, single-arm, dual-arm robots with replaceable grippers.}
    \label{fig:embodiments}
\end{figure}

We support diverse robot embodiments with replaceable gripper configurations, covering multiple commonly used and self-developed robotic platforms. The supported systems include the single-arm Franka equipped with Robotiq Gripper and Panda Hand, the self-developed dual-arm ArtiXon Arm-6A assembled with G-type Gripper and H-type Gripper, the dual-arm ARX R5, as well as the self-developed mobile manipulation platform Quanta X1. We calibrate the visual appearance and physical parameters of each simulated embodiment strictly against their real-world counterparts to reduce simulation bias.

We conduct the majority of data collection and experimental validation on the self-developed ArtiXon Arm-6A and Quanta X1 platforms. To further narrow the real2sim gap, we optimize the low-level control parameters of robotic arms. This optimization guarantees millimeter-level positional deviation between simulated and real robot arms under identical control action inputs. We provide detailed implementation and calibration specifics in Sec.~\ref{sec:exp-sim2real}.

\subsection{Training-evaluation Separation}

\begin{figure}[H]
    \centering
    \includegraphics[width=\linewidth]{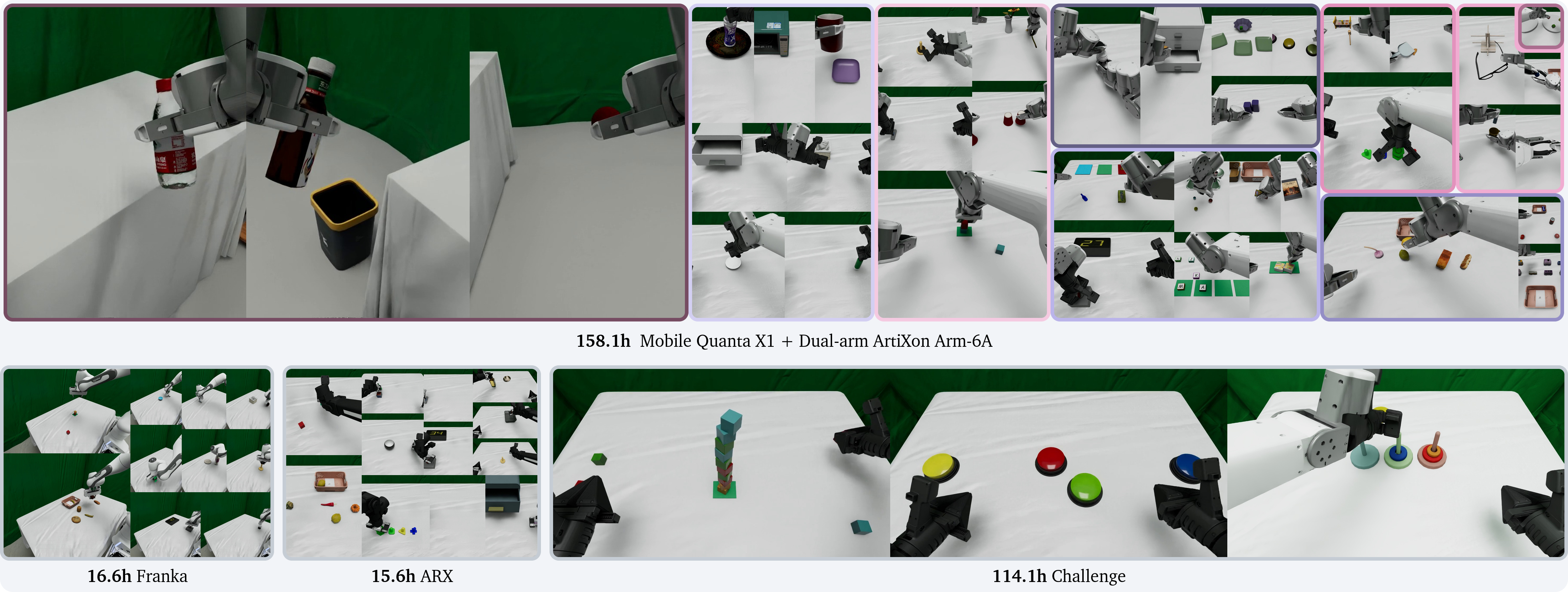}
    \caption{\textbf{X2Real Dataset.} we collect nearly 300 hours high-quality demonstration data on different embodiments. The image area of each task is proportional to its data length.}
    \label{fig:dataset-vis}
\end{figure}

To strictly decouple training configurations from evaluation setups and avoid benchmark exploitation, we build a digital twin of the green-screen simulation environment aligned with ManipArena~\cite{sun2026maniparena} and conduct all data collection within this standardized virtual setting, as visualized in Fig.~\ref{fig:dataset-vis}. We collect 300 demonstration trajectories for each desktop manipulation task, 1000 trajectories for each mobile operation task, and over 20000 trajectories for the three challenging tasks. For data source distribution, approximately 80\% of the ArtiXon Arm data comes from human teleoperation, while the remaining 20\% is generated via automatic data synthesis. In contrast, all data from the Quanta X1 mobile platform, Franka, ARX R5, and all challenging tasks are fully synthesized automatically. In total, we collect 158.1 hours of trajectory data from Quanta X1 and ArtiXon Arm as the primary post-training dataset for our benchmark. We further supplement 16.6 hours of Franka data, 15.6 hours of ARX R5 data, and 114.1 hours of challenging task data, yielding a total dataset duration of nearly 300 hours.

\begin{figure}[H]
    \centering
    \includegraphics[width=\linewidth]{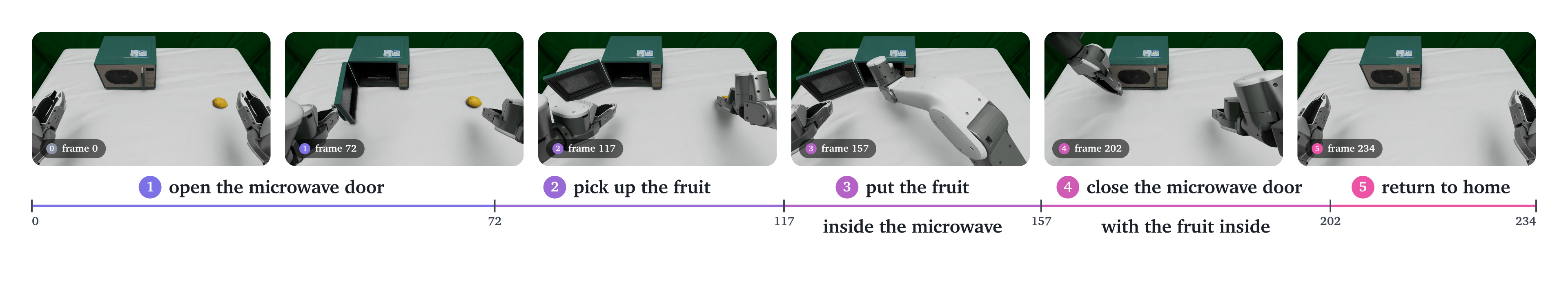}
    \caption{\textbf{Frame-level Labeling.} The Fruit in Microwave task contains 5 subtasks. Our system can automatically label corresponding frame duration during data collection.}
    \label{fig:subtask-labeling}
\end{figure}

We enable the subtask-level success checker throughout both teleoperation and automatic synthesis pipelines, which supports automated temporal labeling for all collected trajectories. As illustrated in Fig.~\ref{fig:subtask-labeling}, each trajectory not only retains raw camera recordings, robotic action sequences, and linguistic instructions, but also contains fine-grained frame-level subtask annotations that precisely document the complete task execution process.

Importantly, X2Real goes beyond a static separation between training and evaluation configurations. Powered by the procedural generation capabilities of the Mana ecosystem, it can continuously instantiate new combinations of assets, layouts, task parameters, instructions, and embodiment configurations under predefined randomization rules. As a result, policies cannot simply memorize a finite collection of evaluation setups or optimize against fixed scene configurations. More importantly, the evaluation distribution itself can evolve over time by introducing new randomized configurations and task variants, reducing benchmark-specific overfitting as policy capabilities improve.

\subsection{Unified Evaluation Protocol}

\begin{figure*}[h]
    \centering
    \makebox[\linewidth][c]{%
        \resizebox{0.96\linewidth}{!}{%
            \input{final_figure_table/policy_space_protocol_tikz.tex}%
        }%
    }
    \caption{
    \textbf{Policy Space Overview}. Policy Space coordinates observation mapping,
    action adaptation, and the episode lifecycle. Within each policy service,
    a model adapter converts mapped observations into model-specific inputs for
    policy inference. Privileged simulator state is accessible only to the evaluator.
    }
    \label{fig:policy_space}
\end{figure*}
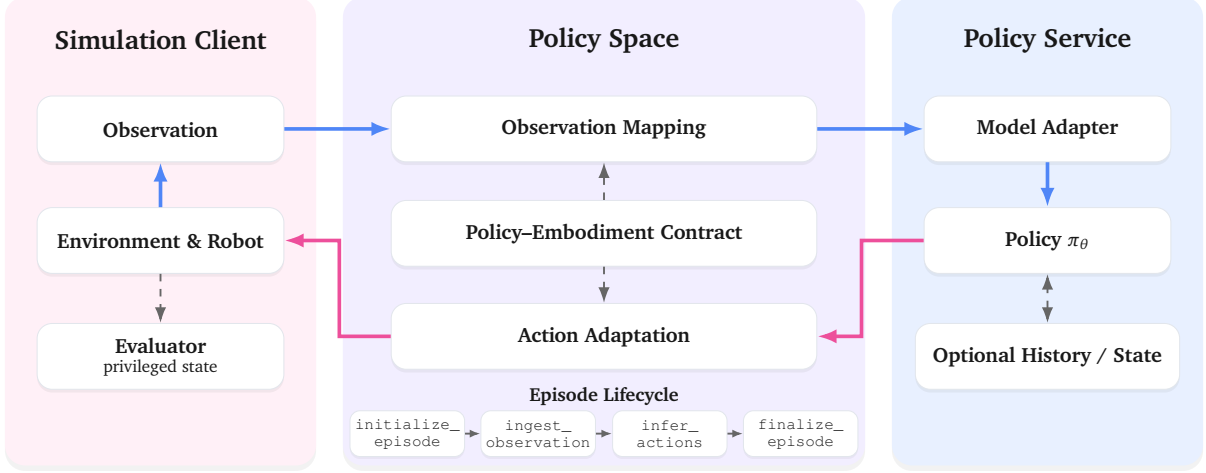

Generalist robot policies often differ in their runtime dependencies, observation formats, history and state requirements, action representations, and inference schedules. To evaluate these policies under a unified protocol, X2Real adopts a dependency-isolated architecture~\citep{atreya2025roboarena,chen2026robodojo,starvla2026,choi2026vlaeval}, in which the simulation client and each policy service run in separate processes and communicate through a lightweight remote interface. Building on this architecture, we introduce \textbf{Policy Space}, which defines a unified runtime contract between the simulation client and policy services across robot embodiments. As shown in Fig.~\ref{fig:policy_space}, Policy Space coordinates observation mapping, action adaptation, and the episode lifecycle. Policy-specific preprocessing and inference remain within each policy service, allowing policies with different native interfaces and inference requirements to be evaluated under the same protocol.

X2Real uses an episode as the basic unit of closed-loop evaluation. Each episode is instantiated from an evaluation configuration that specifies key settings such as the task definition, robot embodiment, language instruction, randomization rules, and sampling seed. It starts with environment initialization and ends when the task succeeds, a failure condition is triggered, or the time limit is reached. For each policy and task, X2Real runs multiple episodes under randomized conditions and reports the success rate, mean progress score, and subtask completion rates.

At each control step, the environment produces an observation. Based on the current embodiment configuration, Policy Space maps the observation into the common format used by policy services. The observation content may vary with the task and embodiment. When inference is requested, the model adapter converts the mapped observation into the policy's native input format and passes it to the policy for inference. The policy output is then returned to Policy Space, which converts it into commands executable by the current robot.

Policy Space uses configurable modules to accommodate policy- and embodiment-specific processing beyond the common observation--action flow. Examples include coordinate and action-representation transformations, embodiment-specific unit conversion, Real-Time Chunking (RTC)~\citep{black2025rtc}, and history updates for history-dependent policies~\citep{ye2026dreamzero}. Related modular interfaces are used in existing robot-learning toolchains and policy-integration frameworks~\citep{cadene2026lerobot,starvla2026,xpolicylab2026}.

Policy Space defines the episode lifecycle through four operations. \texttt{initialize\_episode} prepares the policy for a new episode; \texttt{ingest\_observation} passes the current observation to the policy service to update any required history or internal state; \texttt{infer\_actions} requests the next policy output; and \texttt{finalize\_episode} marks the end of the episode and clears episode-specific state. The evaluation configuration also specifies when observations are passed to the policy service, when inference is requested, and how policy outputs are executed. The same lifecycle applies to policies with different history and state requirements.

Policy inference is separated from task evaluation. Policies receive only the observations provided through Policy Space, while the evaluator uses privileged simulator state that is not exposed to the policy. During each episode, the evaluator records stage completion, task success, progress, and failure information based on the task definition. The resulting evaluation metrics are detailed in Sec.~\ref{sec:dag-guided-evaluation}. For each evaluation setting, all policies use the same task definition, randomization rules, observation content, and evaluation criteria.

%% file: final_figure_table/policy_space_protocol_tikz.tex
\begingroup%
\definecolor{psPaperBlue}{HTML}{4F86F7}%
\definecolor{psPaperPink}{HTML}{ED4F9B}%
\definecolor{psPanelBlue}{HTML}{E8F0FF}%
\definecolor{psPanelPink}{HTML}{FFF0F7}%
\definecolor{psPanelLavender}{HTML}{F2EEFF}%
\definecolor{psInk}{HTML}{181818}%
\definecolor{psMuted}{HTML}{666666}%
\definecolor{psSoftLine}{HTML}{E2E6EC}%
\tikzset{
  psEvery/.style={font=\rmfamily\fontsize{8.2}{9.4}\selectfont, text=psInk},
  psPanel/.style={rounded corners=8pt, draw=none,
    drop shadow={shadow xshift=0pt,shadow yshift=-2.2pt,opacity=.10}},
  psCard/.style={rounded corners=6pt, draw=psSoftLine, line width=.35pt,
    fill=white, minimum height=.92cm, align=center,
    inner xsep=7pt, inner ysep=5pt,
    drop shadow={shadow xshift=0pt,shadow yshift=-1.3pt,opacity=.10}},
  psLifecycle/.style={rounded corners=4pt, draw=psSoftLine,
    line width=.35pt, fill=white, minimum width=1.60cm,
    minimum height=.64cm, align=center, inner xsep=2pt, inner ysep=3pt,
    font=\ttfamily\fontsize{6.3}{7.0}\selectfont,
    drop shadow={shadow xshift=0pt,shadow yshift=-1pt,opacity=.09}},
  psObsFlow/.style={-{Latex[length=2.7mm,width=2.0mm]},
    draw=psPaperBlue, line width=1.45pt},
  psActFlow/.style={-{Latex[length=2.7mm,width=2.0mm]},
    draw=psPaperPink, line width=1.45pt},
  psAuxFlow/.style={-{Latex[length=2.2mm,width=1.6mm]},
    draw=psMuted, line width=.75pt, dashed},
  psAuxBiFlow/.style={{Latex[length=2.2mm,width=1.6mm]}-{Latex[length=2.2mm,width=1.6mm]},
    draw=psMuted, line width=.75pt, dashed},
  psLifecycleFlow/.style={-{Latex[length=1.5mm,width=1.15mm]},
    draw=psMuted, line width=.75pt}
}%
\begin{tikzpicture}[x=1cm,y=1cm,every node/.style={psEvery}]
  \node[psPanel, fill=psPanelPink, minimum width=4.35cm,
    minimum height=6.55cm] at (0,1.05) {};
  \node[psPanel, fill=psPanelLavender, minimum width=7.30cm,
    minimum height=6.55cm] at (6.20,1.05) {};
  \node[psPanel, fill=psPanelBlue, minimum width=4.35cm,
    minimum height=6.55cm] at (12.40,1.05) {};

  \node[font=\rmfamily\bfseries\fontsize{10.5}{12}\selectfont]
    at (0,3.73) {Simulation Client};
  \node[font=\rmfamily\bfseries\fontsize{10.5}{12}\selectfont]
    at (6.20,3.73) {Policy Space};
  \node[font=\rmfamily\bfseries\fontsize{10.5}{12}\selectfont]
    at (12.40,3.73) {Policy Service};

  \node[psCard, minimum width=3.45cm] (psObservation) at (0,2.48)
    {\textbf{Observation}};
  \node[psCard, minimum width=3.45cm] (psEnvironment) at (0,0.92)
    {\textbf{Environment \& Robot}};
  \node[psCard, minimum width=3.45cm] (psEvaluator) at (0,-0.70)
    {\textbf{Evaluator}\\[-2pt]
     {\fontsize{6.8}{7.5}\selectfont privileged state}};

  \node[psCard, minimum width=5.95cm] (psMapping) at (6.20,2.48)
    {\textbf{Observation Mapping}};
  \node[psCard, minimum width=5.95cm] (psContract) at (6.20,1.02)
    {\textbf{Policy--Embodiment Contract}};
  \node[psCard, minimum width=5.95cm] (psAdaptation) at (6.20,-0.42)
    {\textbf{Action Adaptation}};

  \node[psCard, minimum width=3.45cm] (psAdapter) at (12.40,2.48)
    {\textbf{Model Adapter}};
  \node[psCard, minimum width=3.45cm] (psPolicy) at (12.40,0.92)
    {\textbf{Policy $\pi_\theta$}};
  \node[psCard, minimum width=3.45cm] (psContext) at (12.40,-0.70)
    {\textbf{Optional History / State}};

  \draw[psObsFlow] (psEnvironment.north) -- (psObservation.south);
  \draw[psObsFlow] (psObservation.east) -- (psMapping.west);
  \draw[psObsFlow] (psMapping.east) -- (psAdapter.west);
  \draw[psObsFlow] (psAdapter.south) -- (psPolicy.north);

  \draw[psActFlow] (psPolicy.west) -- (9.80,0.92) --
    (9.80,-0.42) -- (psAdaptation.east);
  \draw[psActFlow] (psAdaptation.west) -- (2.50,-0.42) --
    (2.50,0.92) -- (psEnvironment.east);

  \draw[psAuxBiFlow] (psContext.north) -- (psPolicy.south);
  \draw[psAuxFlow] (psEnvironment.south) -- (psEvaluator.north);
  \draw[psAuxFlow] (psContract.north) -- (psMapping.south);
  \draw[psAuxFlow] (psContract.south) -- (psAdaptation.north);

  \node[font=\rmfamily\bfseries\fontsize{7.5}{8.5}\selectfont]
    at (6.20,-1.25) {Episode Lifecycle};
  \node[psLifecycle] (psInit) at (3.45,-1.78)
    {initialize\_\\episode};
  \node[psLifecycle] (psIngest) at (5.28,-1.78)
    {ingest\_\\observation};
  \node[psLifecycle] (psInfer) at (7.12,-1.78)
    {infer\_\\actions};
  \node[psLifecycle] (psFinalize) at (8.95,-1.78)
    {finalize\_\\episode};
  \draw[psLifecycleFlow] (psInit) -- (psIngest);
  \draw[psLifecycleFlow] (psIngest) -- (psInfer);
  \draw[psLifecycleFlow] (psInfer) -- (psFinalize);
\end{tikzpicture}%
\endgroup%

%% file: sec/3_platform.tex
\section{Mana Simulation Platform}\label{sec:platform}

\subsection{Agent-driven System Design}

Large‑language‑model (LLM) agents exhibit strong capabilities in coding and mathematical reasoning~\cite{yang2024sweagent,gou2024tora}. Nevertheless, their potential for physical AI research remains largely untapped. Ideally, such agents could assist the full pipeline of embodied research: on one hand supporting benchmark and evaluation workflows, including task design, environment composition, and policy rollout; on the other hand facilitating model evolution via trajectory generation, data augmentation, and fine‑grained performance analysis. However, the inherent complexity of low‑level simulation stacks prevents LLM agents from realizing these capabilities at scale. Direct interaction with simulators requires handling sprawling low‑level APIs, physical parameter tuning, robot embodiment configurations, scene assembly, and hand‑crafted task success logic. Unmediated LLM calls to such interfaces frequently produce invalid configurations, runtime errors, and brittle task implementations, making large‑scale autonomous iteration impractical. Isaac Lab Arena~\cite{isaaclab-arena2025} simplifies low‑level simulation interfaces by factoring environments into three composable building blocks: scene, task, and embodiment, which can be flexibly assembled to construct diverse simulation scenarios. However, it lacks built‑in asset management facilities and integrated data pipelines, and offers no agent‑centric framework to enable model‑evolution workflows. To unlock this agent‑simulation closed‑loop workflow, we present Mana, a comprehensive agent‑driven simulation infrastructure.

As shown in Fig.~\ref{fig:pipeline}, Mana is organized around a
YAML-based physical Domain-Specific Language (DSL). Rather than replacing Arena's
scene--task--embodiment factorization, the DSL adopts it as the runtime abstraction and supplies the pieces Arena leaves out: asset registration and placement, a way to name and compose the three factors, and a unified rollout path. Everything an agent touches is expressed as a modular YAML primitive, and a compile-and-check harness sits behind these primitives: it expands \texttt{include} references, lowers the result through a \emph{Translator} into factory orders, and validates the whole
pipeline before any simulator process starts. An agent can therefore assemble a valid environment by editing declarative text, without ever calling a low-level simulator API. We describe this design in four steps---the authoring surface, compilation into factories and orders, assembly, and rollout---and then return to the two loops the agent runs on top of it.

\subsubsection{Authoring Surface}

The DSL is organized at two levels. A \emph{pipeline} file is the
top-level program: it declares which factories are active, the order in
which they compile, and the streams of orders they consume. The substance
of an environment, however, lives in \emph{fragments}---independently
authored YAML files for scenes, tasks, robot embodiments, policies, and
collection presets---which the pipeline pulls in with \texttt{include}.
Because inclusion expands a referenced fragment and then lets the
including file override individual fields, a new variant is written as a
small overlay rather than a duplicated file, as in Listing~\ref{lst:overlay}.

\begin{lstlisting}[caption={A scene variant authored as an overlay.},label={lst:overlay}]
include: scene/pick_place.yaml
definition:
  scene_type: pick_place_ood
\end{lstlisting}

Keeping independent concerns in independent fragments is what makes an
edit local: changing one fragment does not perturb the others. A scene
fragment, for instance, describes only the world---its assets, a
background with the surfaces it must expose, and a layout---and says
nothing about the robot or the success criteria (Listing~\ref{lst:scene}).
Each object carries an identity (a concrete asset or a pool of
candidates), a set of \texttt{semantic\_tags} that later act as role
handles, and a placement, which may be a fixed pose or a sampled region
on a surface. Optional variant blocks randomize assets, materials, or
lighting, and a layout solver discards colliding placements before the
scene is admitted, so an authored scene is guaranteed to be physically
realizable.

\begin{lstlisting}[caption={A scene fragment: identity, semantic tags, and placement.},label={lst:scene}]
objects:
  - id: mug
    asset: {type: pool, candidates: [{id: mugs, assets: [mug_a, mug_b]}]}
    semantic_tags: [pickable]
layout:
  nodes:
    mug:
      object: mug
      placement: {type: surface, surface: tabletop, ranges: {x: [0.10, 0.20], y: [-0.10, 0.10]}}
\end{lstlisting}

A task fragment describes the logic as a directed acyclic graph.
Graph-level fields hold the instruction and the reset-time sampling,
while each node is an atomic subtask equipped with role slots,
declarative success terms, optional sensors, and a partial score; the
edges impose a partial order over these nodes (Listing~\ref{lst:task}).
Since node files can themselves be included, a single stage is reused
across many graphs. Success terms are named functions resolved at compile
time rather than hand-written checkers, and a task refers to the scene
only through tags such as \texttt{@pickable}, never through concrete
asset paths. Atomic skills (\texttt{pick}, \texttt{place}, \texttt{push},
and related primitives) may additionally attach on the data-generation
path, where a skill or motion planner consumes them; evaluation itself
relies only on the success terms.

\begin{lstlisting}[caption={A task fragment: a DAG whose nodes carry declarative success terms.},label={lst:task}]
graph_task_generator:
  nodes:
    - {id: place, include: task/pick_place/stages/place.yaml, score: 3}
    - {id: return, include: task/pick_place/stages/return.yaml, score: 1}
  edges:
    - {src: place, dst: return}
# place.yaml
success:
  _func: object_near_destination
  object_cfg: '@pickable'
  destination_cfg: '@container'
\end{lstlisting}

A robot fragment, finally, specifies an embodiment---its kinematics,
gripper variant, cameras, and actuator law---while policy and collection
fragments specify how a model is served and how a teacher is attached.
Crucially, no fragment ever embeds a complete environment; each describes
exactly one concern.




\subsubsection{Factories and Orders}

These authoring files are never interpreted directly at runtime. Instead,
the Translator expands every \texttt{include} and lowers the pipeline into
two artifacts: factory configurations and \emph{orders}. A factory is
simply a lifecycle---setup, receive, fulfill, cleanup---whereas an order
is a named request to generate one thing: a particular scene, robot, task,
assembly, or rollout (Listing~\ref{lst:orders}). This separation is what
keeps agent edits safe: because factories consume already-lowered orders
rather than raw YAML, an agent extends the system by writing another
order, not by patching factory code.

\begin{lstlisting}[caption={Order streams reference fragments by name.},label={lst:orders}]
scene_factory: {type: default_scene_factory}
scene_orders:
  - include: scene/pick_place.yaml          # scene_type: pick_place
  - include: scene/pick_place_ood.yaml      # overlay, scene_type: pick_place_ood
robot_orders:
  - {include: robot/arm.yaml, name: ARM-G, gripper_variant: G}
task_orders:
  - include: task/pick_place/vnext_dag.yaml
\end{lstlisting}

Fulfilling an order follows a single chain: the factory batches and caches
work, a generator composes one complete object, and components implement
the atomic capabilities---background, objects, lighting, termination,
events, metrics---so that extending the language usually means adding a
component rather than branching a factory. The scene, robot, and task
factories act as independent part suppliers; once shared data services are
up, they can run in parallel and store their outputs by name in a shared
data center.

\subsubsection{Assembly}

An assembly order is a triple of generator names. Given such a triple, the
assembly factory fetches the three parts from the data center, binds each
task role to the scene assets whose tags satisfy it (so \texttt{@pickable}
resolves to objects tagged \texttt{pickable}), and emits an environment for
the Arena compiler to instantiate (Listing~\ref{lst:assembly}). Episode
length and step limits remain properties of the task, so assembly stays
purely compositional and never owns the MDP.

\begin{lstlisting}[caption={An assembly order binds scene, robot, and task by name.},label={lst:assembly}]
assembly_orders:
  - assembly_generator:
      name: arm_G_pick_place_env
      scene_generator_name: pick_place
      robot_generator_name: ARM-G
      task_generator_name: pick_place
\end{lstlisting}

Because binding is by name, a large matrix of environments follows from a
small set of parts without duplication: one scene order can appear in many
assemblies, and one task graph can be paired with several gripper variants.
The pipeline DAG fixes the only legal compile order---data services, then
the three part factories, then assembly, then rollout---and a program that
violates it, for example by assembling before its parts exist, is rejected
at dry-run, before the simulator is ever launched.

\subsubsection{Unified Rollout}

A rollout order reuses an environment rather than rebuilding it: it names
an existing \texttt{env\_ref} and the teacher that will drive it, be that
teleoperation, a motion or skill planner, policy inference, or trajectory
replay. All teachers share the same get-action-and-step loop, so a single
assembly can carry both a data-generation order and an evaluation order
(Listing~\ref{lst:rollout}). Both emit trajectories annotated with per-node
DAG scores and instruction logs, which makes fine-grained analysis an
intrinsic property of the task graph rather than a separate instrumentation
layer.

\begin{lstlisting}[caption={One assembly, two rollout modes: data generation and evaluation.},label={lst:rollout}]
rollout:
  - {env_ref: arm_G_pick_place_env, modes: [gendata], include: datacollection/teleop.yaml}
  - {env_ref: arm_G_pick_place_env, modes: [benchmark], planner: {PolicyPlanner: {policy_cfg: {include: policy/vla.yaml}}}}
\end{lstlisting}

\subsubsection{Agent-driven Loops}

Equipped with this DSL, an agent drives the two loops named in the
opening---benchmark construction and model evolution---entirely through
the harness rather than through simulator APIs. The agent remains the
author of each program; the harness merely guarantees that every edit is
compilable and every run attributable.

In the task-design loop, the agent writes or patches fragments, appends or
rewires orders, validates the pipeline, and dry-runs the compile graph.
Such edits are naturally local: including a different scene overlay,
retargeting an assembly to another robot, adding a node to a task DAG, or
attaching a new teacher to an existing \texttt{env\_ref}. The Translator
then lowers the edited program and the assembly factory instantiates fresh
environments. Should the schema prove insufficient, the agent can still
extend a component directly, but the common path never requires touching a
simulator API.

In the rollout-analysis loop, the agent launches data generation or
evaluation on selected environments, reads the node-level metadata and
failure trajectories they produce, and feeds these observations
back---either as further edits to the DSL or as sliced demonstrations for
augmentation. The human contributes only high-level intent at the pipeline
scale, while the agent carries out the iteration within the language
itself.

\subsection{Simulation-ready Assets and Semantic Scene Construction}

\input{sec/assets_scene/content}

\subsection{Atomic Task \& Skill Composition}
\label{sec:task-skill}
\input{sec/atomic_task_skill_composition/content}

\subsection{Efficiency Optimization}
\label{sec:runtime-optimization}

\begin{figure}[t]
    \centering
    \includegraphics[width=\linewidth]{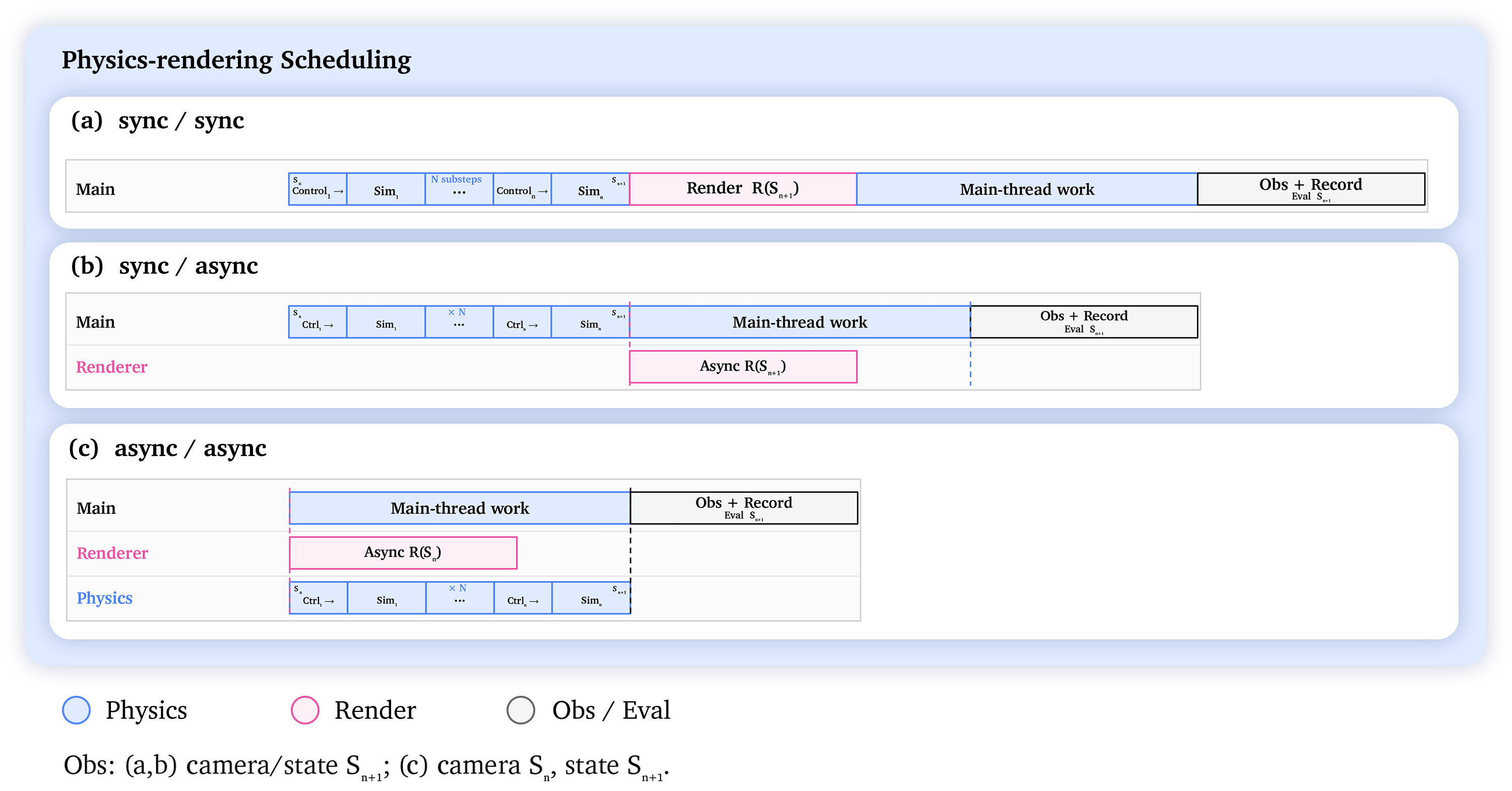}
    \caption{\textbf{Physics-render scheduling in Mana.} (a) \textit{sync/sync} serializes physics and rendering. (b) \textit{sync/async} overlaps $R(S_{n+1})$ with main-thread work. (c) \textit{async/async} overlaps $R(S_n)$ with physics advancing to $S_{n+1}$; camera observations use $S_n$, while state observations and evaluation use $S_{n+1}$. }
    \label{fig:runtime-scheduling}
\end{figure}

Built on the Isaac ecosystem, Mana treats each environment step as a heterogeneous CPU--GPU pipeline. Isaac Lab's default manager-based reinforcement-learning environment serializes action processing, the decimated physics loop, rendering, observation, and evaluation. Within each of the $N$ physics substeps, actuator computation precedes simulation. Mana controls physics stepping and rendering independently, yielding the three execution modes in Fig.~\ref{fig:runtime-scheduling}. The \textit{async/async} mode is the default Mana configuration, while \textit{sync/sync} and \textit{sync/async} provide controlled baselines.

\paragraph{Optimized Camera Pipeline.}
All three scheduling modes use the same optimized camera pipeline. It tracks camera updates on the CPU, stores rendered outputs in CUDA double buffers, and moves image layout conversion and post-processing to a dedicated GPU stream. This design reduces per-frame CPU--GPU synchronization. In a controlled \textit{sync/sync} ablation with the same environment and camera configuration, it reduced mean latency by 28.41\% relative to Isaac Lab's default camera pipeline.

\paragraph{CPU Physics.}
For rigid-body manipulation, Mana places physics and actuator computation on the CPU. This profile enables continuous collision detection (CCD) to reduce high-speed tunneling and improve collision robustness. It also reserves the GPU for multi-camera rendering, allowing physics and rendering to overlap without competing for GPU resources. The dual-asynchronous path is currently validated only for CPU PhysX rigid-body workloads; deformable bodies, cloth, and particles remain outside its scope.

\paragraph{Physics-render Scheduling.}
The \textit{sync/sync} baseline completes the $N$ actuator--simulation substeps, renders the resulting state $S_{n+1}$, executes the remaining main-thread work, and finally evaluates and records observations. The \textit{sync/async} mode keeps physics synchronous but submits $R(S_{n+1})$ asynchronously, allowing rendering to overlap with main-thread work. The environment waits for both paths before returning, so camera and state observations correspond to $S_{n+1}$.

The dual-asynchronous design extends Mana's earlier two-thread CPU-physics/rendering pipeline with an explicit frozen-state boundary. In \textit{async/async}, a persistent CPU worker executes the complete $N$-substep actuator and physics sequence. In parallel, the renderer consumes a frozen snapshot $S_n$, and the main thread performs independent work. At the synchronization boundary, the caller joins the worker and renderer, consumes and records the $S_n$ camera result, publishes $S_{n+1}$, and completes state observation, evaluation, and post-step recording. Camera observations therefore incur a deliberate one-step delay, whereas state observations and evaluation use the current state; no rendering work crosses the environment-step boundary.

\paragraph{Performance.}
Let $T_{\mathrm{physics}}=\sum_{i=1}^{N}(T_{\mathrm{actuator},i}+T_{\mathrm{simulation},i})$ denote the complete CPU physics sequence, and let $T_{\mathrm{render}}$, $T_{\mathrm{main}}$, and $T_{\mathrm{obs}}$ denote rendering, main-thread work, and final observation/evaluation. The critical paths of \textit{sync/sync}, \textit{sync/async}, and \textit{async/async}, denoted $T_A$, $T_B$, and $T_C$, are approximated by
\[
\begin{aligned}
T_A &\approx T_{\mathrm{physics}} + T_{\mathrm{render}} + T_{\mathrm{main}} + T_{\mathrm{obs}}, \\
T_B &\approx T_{\mathrm{physics}} + \max\!\left(T_{\mathrm{render}},T_{\mathrm{main}}\right) + T_{\mathrm{obs}} + T_{\mathrm{sync},B}, \\
T_C &\approx \max\!\left(T_{\mathrm{physics}},T_{\mathrm{render}},T_{\mathrm{main}}\right) + T_{\mathrm{obs}} + T_{\mathrm{sync},C},
\end{aligned}
\]
where $T_{\mathrm{sync},B}$ and $T_{\mathrm{sync},C}$ denote mode-specific submission and synchronization overheads. These expressions summarize the overlap structure; all reported latencies are measured end to end.

We evaluate Mana's three scheduling modes on the same workstation (AMD Ryzen 9 9950X CPU and NVIDIA GeForce RTX 5090 GPU), using the same seed and three cameras: two $640\times480$ wrist cameras and one $1280\times720$ head camera. Over one complete EX001-G episode from the X2Real benchmark, the mean per-step latencies of \textit{sync/sync}, \textit{sync/async}, and \textit{async/async} are 46.79, 24.68, and 22.43~ms, respectively. Relative to \textit{sync/sync}, the two asynchronous modes reduce latency by 47.24\% and 52.06\%; \textit{async/async} further reduces it by 9.13\% relative to \textit{sync/async}. For tasks that tolerate the one-step camera delay, these controlled single-episode results indicate that the default dual-asynchronous schedule improves data-collection and policy-evaluation efficiency without changing the physics time step or control frequency; they do not estimate cross-task throughput.

%% file: sec/assets_scene/content.tex

\label{sec:assets-scenes-runtime}
\suppressfloats[t]

\input{sec/assets_scene/sections/00_overview}
\input{sec/assets_scene/sections/10_asset_pipeline}
\input{sec/assets_scene/sections/20_scene_layout}
\input{sec/assets_scene/sections/30_runtime_diversity}

\FloatBarrier

%% file: sec/assets_scene/sections/00_overview.tex
X2Real connects three content-side capabilities: simulation-ready asset
preparation, semantic scene construction, and task-conditioned variation with
state-aligned replay.  Their factorization supports evaluation fidelity by
reducing content-side simulation-to-real discrepancies, evaluation breadth
through task-role coverage and constraint-preserving composition, and benchmark
integrity through explicit identities, controlled content hold-outs, and
resampling.

Asset preparation, including AI-generated 3D (AIG3D) synthesis, creates
simulation-ready packages; the scene program and compiler resolve semantic
placement intent into compiled scenes; reset-time variation and replay diversify
episodes and observations.

\begin{figure}[H]
  \centering
  \input{sec/assets_scene/figures/fig_asset_collection_spectrum}
  \caption{\textbf{Task-oriented tabletop asset spectrum.}
  Representative packages span 44 rigid categories and nine articulated
  interaction groups.  Rigid examples are arranged in four visual groups for
  readability; the collection taxonomy contains ten semantic families.}
  \label{fig:asset-collection-previews}
\end{figure}

%% file: sec/assets_scene/figures/fig_asset_collection_spectrum.tex
\includegraphics[width=\linewidth]{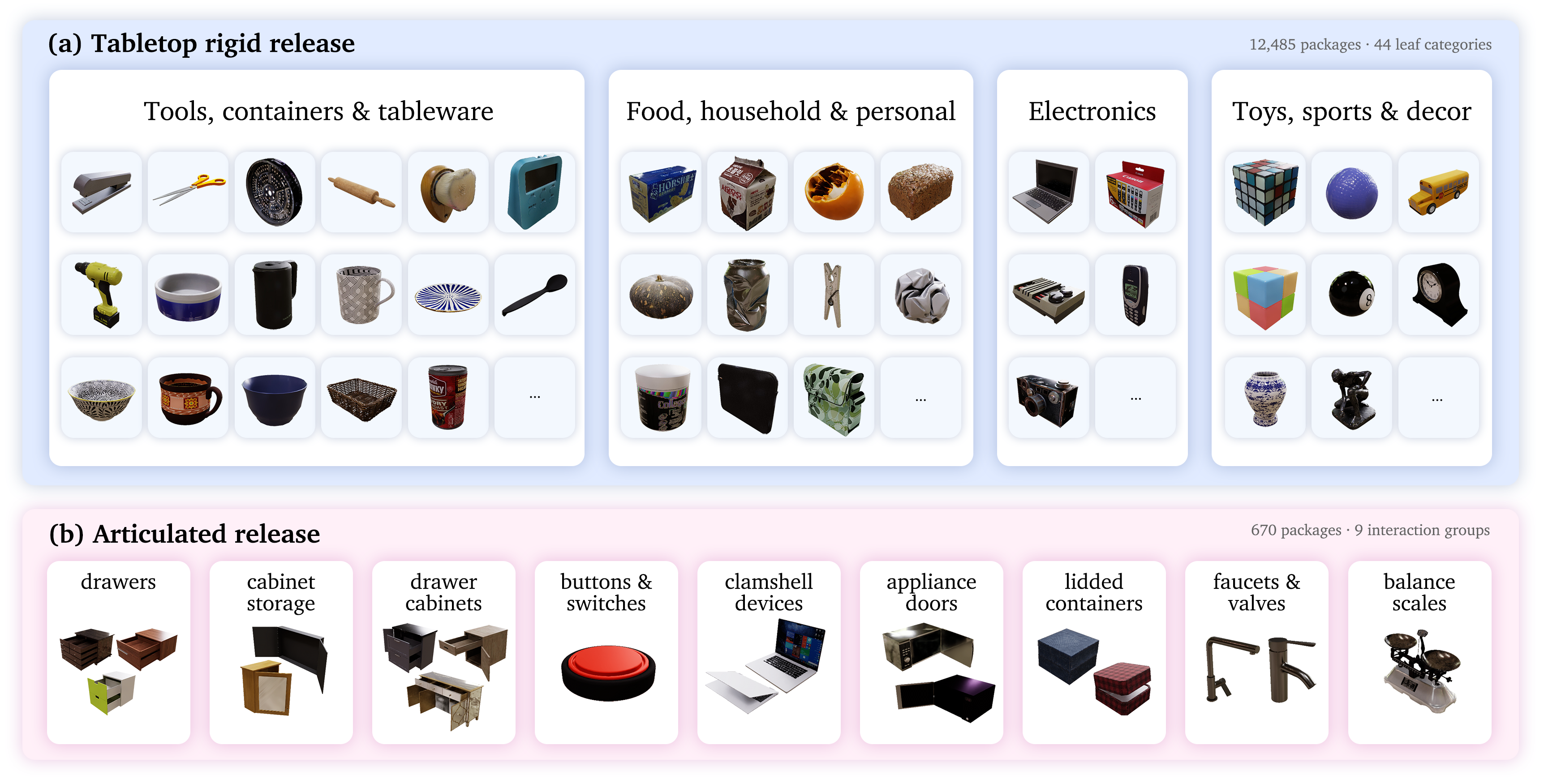}

%% file: sec/assets_scene/sections/10_asset_pipeline.tex
\subsubsection{Task-oriented Simulation-ready Assets}
\label{sec:sim-ready-assets}

A simulation-ready asset is a reusable package that bundles visual, metric,
semantic, collision, inertial, kinematic, and physical-material records and has
passed the simulator checks required by its declared roles.  These packages
determine interaction coverage and support reliable instantiation of
heterogeneous content in task scenes.

X2Real organizes the tabletop collection around \emph{task roles and interaction
modes}; raw catalog size and generic taxonomies alone do not establish
manipulation coverage.  Manipulated objects, receptacles, tools, fixtures, and
articulated mechanisms must support the benchmark interactions, while category
breadth provides visual and semantic diversity within each role.

Existing ecosystems emphasize complementary parts of the asset--task interface,
including reusable task definitions, realistic content, metric or grasp
annotations, and language-conditioned interaction metadata
\cite{liu2023libero,nasiriany2024robocasa,nasiriany2026robocasa365,kim2026molmospaces,chen2025robotwin2,chen2026robodojo}.  X2Real integrates these
signals with validated metric scale, part- and task-level interaction records,
physical-material authoring, and simulator-side acceptance in a common
simulation-ready asset package.

The collection combines Objaverse/CanoVerse, Omni6DPose, BlenderKit, GSO,
ManiTwin-100K, HSSD, ProcTHOR, ABO, Poly Haven, and ReplicaCAD with task-specific
and generated additions
\cite{deitke2023objaverse,jin2026canoverse,zhang2024omni6dpose,blenderkit2026,downs2022google,wang2026manitwin,khanna2024hssd,deitke2022procthor,collins2022abo,polyhaven2026,szot2021habitat2}.
Each accepted unit is released as one \emph{benchmark package}.
The indexed rigid catalog contains 96,480 source records; the tabletop release
contains 12,485 rigid packages spanning 44 leaf categories and 10 semantic
families, and the articulated collection contains 670 packages across nine
interaction groups.  AIG3D additions are included in the tabletop rigid total,
while deformable assets form a separate collection.  Counts are package-level,
so geometrically similar content from different sources may remain distinct
packages.
Figure~\ref{fig:asset-collection-previews} shows the visual and interaction
breadth of the released collections; detailed provenance appears in
Appendix~\ref{app:asset-preparation-details}.

To turn collected content into simulation-ready packages, X2Real uses seven
dependency-ordered stages that account for interactions among geometry,
appearance, scale, semantics, and physics
(Fig.~\ref{fig:asset-preparation-pipeline}).  The pipeline preserves trustworthy
source evidence, repairs recoverable defects, and revalidates affected downstream
records before release.

\noindent\textbf{Source intake and scope screening.}
Sourced and generated content enters a common OpenUSD representation.  Intake
screens parseability, benchmark-domain fit, supported roles, and object coherence
while preserving reliable geometry, texture, and material evidence.  Compound
scenes are decomposed when possible; ambiguous content is reviewed, and malformed
or out-of-scope content is excluded.

\begin{figure}[!htbp]
  \centering
  \input{sec/assets_scene/figures/fig_asset_preparation_pipeline}
  \caption{\textbf{Dependency-ordered simulation-ready asset preparation.}
  Seven stages transform sourced or generated content into a
  benchmark release package; failed candidates return for targeted repair and
  dependent-record revalidation.}
  \label{fig:asset-preparation-pipeline}
\end{figure}
\FloatBarrier

\begin{figure}[!t]
  \centering
  \input{sec/assets_scene/figures/fig_aig3d_asset_synthesis}
  \caption{\textbf{Coverage-directed AIG3D asset synthesis.}
  Text or image evidence conditions detailed reconstruction,
  simulation-oriented retopology, and PBR regeneration on the final mesh; four
  examples show the resulting long-tail geometry and appearance.}
  \label{fig:aig3d-asset-synthesis}
\end{figure}

\noindent\textbf{Geometric canonicalization and quality repair.}
The pipeline resolves source hierarchy and transforms, normalizes
axes and reliable source units, and checks metric extent, topology, component
structure, surface orientation, and complexity at mesh and merged-object levels.
Repaired assets are rechecked for bounds, origin, orientation, and components.  A
normalized annotation frame supports visual reasoning without changing physical
scale.

\noindent\textbf{Appearance normalization and PBR recovery.}
The pipeline maps source materials to a common PBR model according to optical
behavior while preserving trustworthy maps and scalar values.  Missing
or degraded appearance first draws from category-conditioned candidates;
generative PBR synthesis is used when these are insufficient.  Standardized views
then verify the whole-object rendering-material record and its asset bindings,
which remain separate from physical-material records.

\noindent\textbf{Object-level semantic and metric grounding.}
Standardized multiview observations support instance description,
open-vocabulary retrieval, task-role matching, and semantic orientation.  A
multimodal annotator combines these views with category context and source
metadata, using controlled role vocabularies and free-form text for long-tail
distinctions.  Candidate rotations relative to a category reference pose
establish directional meaning \cite{jin2026canoverse}.  When source scale is
unreliable, category- and function-matched GSO, ABO, and YCB anchors condition a
relative-scale estimate followed by plausibility checks
\cite{downs2022google,collins2022abo,calli2017ycb}.  Semantic pose supplies
directional meaning; metric scale controls clearance, placement domains, and
physical proxies.

\noindent\textbf{Part semantics and task-conditioned interaction grounding.}
Object labels alone cannot identify the regions required by contact-rich tasks.
PartSAM proposes candidate 3D regions from surface evidence
\cite{zhu2025partsam}; multiview observations, 3D containment, and directional
relations reconcile them into semantically consistent parts.  The pipeline
represents an affordance as a relation among a part, a task, and an admissible
interaction---for example, placement, cap removal, or pouring.  Unsupported
assignments are rejected or reviewed, while interaction records bind contact
regions and orientation requirements to scene and task roles so one asset can
expose different regions for grasping, opening, pouring, or placement.

\noindent\textbf{Physical, kinematic, and contact-material authoring.}
Collision construction first tests analytic boxes, spheres, and capsules,
followed by a convex hull; V-HACD or CoACD decomposition handles task-relevant
concavities that require multiple parts
\cite{mamou2009approximate,wei2022coacd}.  Mass and inertia follow a
confidence-ranked hierarchy from reliable watertight volume to convex- or
bounding-volume estimates.  For articulated assets, link--joint topology,
frames, signed axes, limits, collision shapes, inertials, and drives are
normalized in one object frame.  Visual PBR evidence, multiview appearance,
category, and part semantics determine a coarse physical-material class.
Simulator-specific contact-pair priors then assign physical properties and bind
the resulting record to the corresponding collision parts.

\noindent\textbf{Role-specific simulator acceptance and targeted repair.}
A package enters the benchmark release after the simulator exercises the
interactions required by its declared role.  Common checks cover metric extent,
render--collision alignment, collider validity, and settling.
Manipulable rigid objects are grasped, transported, released, and checked after
placement; articulated objects and fixtures additionally exercise authored
joints, limits, collision behavior, and task-relevant contacts.  The same
role-specific acceptance procedure applies to sourced and AIG3D candidates.
Failed records are repaired and retested or excluded.

Implementation details for asset conversion, annotation, physical authoring,
and acceptance appear in Appendix~\ref{app:asset-preparation-details}.

\paragraph{Coverage-directed AIG3D Asset Synthesis}

This synthesis branch fills long-tail gaps in the task-role coverage of sourced
collections, including crushed packaging, peels, cores, and refuse with unusual
form, state, or appearance.  Text requests are converted into reference images,
while single- or multiview observations can condition reconstruction directly.
AIG3D reconstructs high-detail geometry and PBR appearance before
simulation-oriented remeshing and retopology, then reuses the input evidence to
regenerate PBR materials on the final mesh.  This ordering preserves silhouette
and local structure before imposing the simulation mesh budget.

Generated outputs complete the same preparation and acceptance path as sourced
content.  Accepted additions join the tabletop rigid release and become
selectable for scene compilation (Section~\ref{sec:semantic-scene-layout}).
Figure~\ref{fig:aig3d-asset-synthesis} summarizes the synthesis path and examples;
implementation details appear in
Appendix~\ref{app:asset-preparation-details}.

\FloatBarrier

%% file: sec/assets_scene/figures/fig_asset_preparation_pipeline.tex
\includegraphics[width=.9\linewidth]{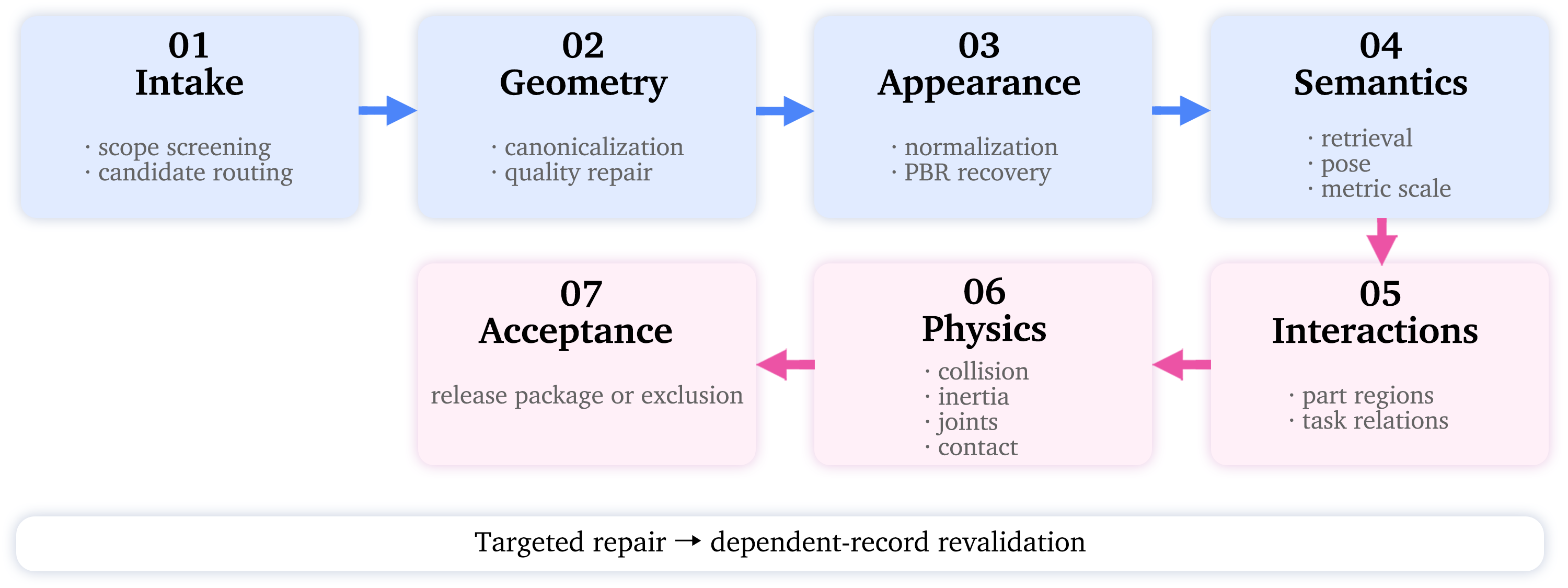}

%% file: sec/assets_scene/figures/fig_aig3d_asset_synthesis.tex
\includegraphics[width=.95\linewidth]{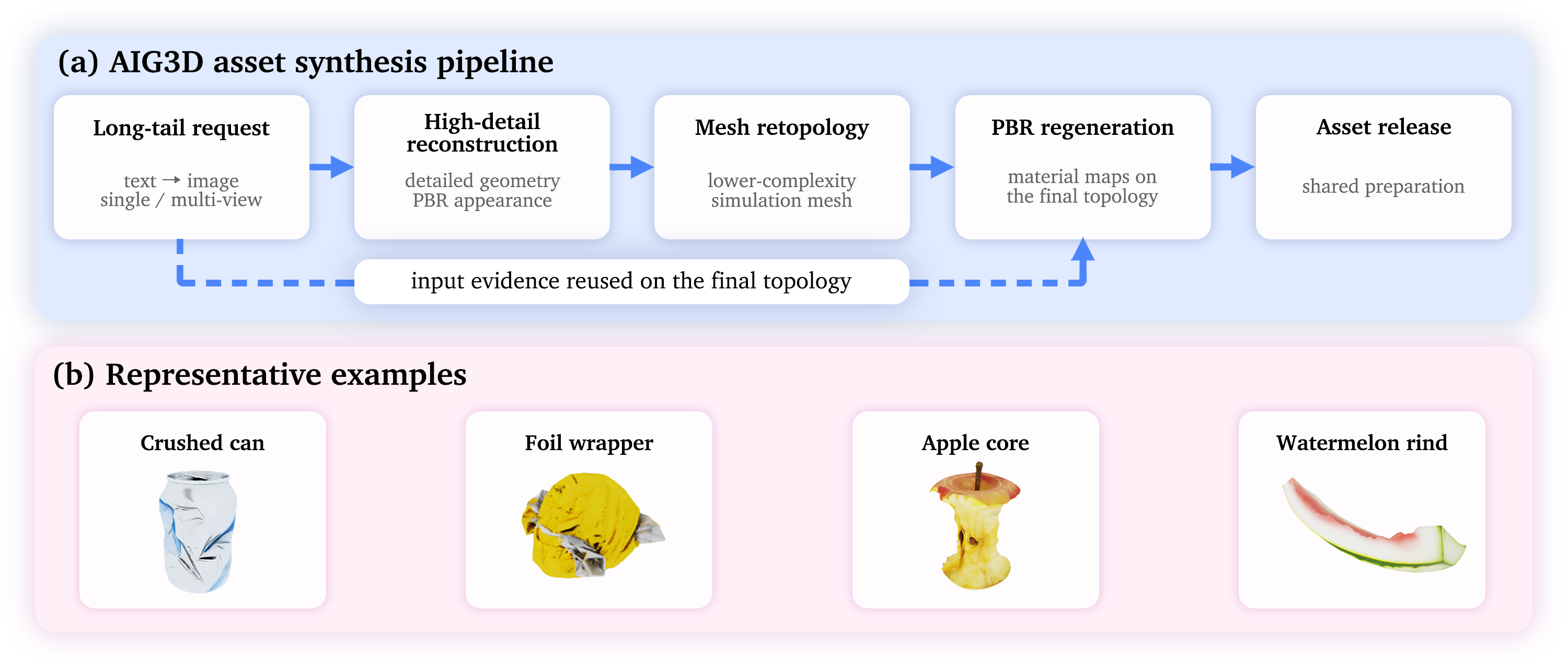}

%% file: sec/assets_scene/sections/20_scene_layout.tex
\subsubsection{Semantic Scene Specification and Construction}
\label{sec:semantic-scene-layout}

A world-space pose that is valid for one tabletop or receptacle can become
invalid when the support is resized or the object is replaced.  A fixed
transform does not preserve which surface supports the object, which volume
contains it, or which spatial relation the task requires.  Scene definitions
must therefore preserve this intent as assets, supports, receptacles,
backgrounds, and scales change.

The \emph{scene specification} defines the abstract semantic model, and a
\emph{semantic scene program} is its concrete authoring artifact.  The
\emph{X2Real scene compiler} resolves the program against selected geometry and
produces either a \emph{compiled scene} satisfying the declared static geometric
and relational constraints or an infeasibility diagnostic.
Figure~\ref{fig:scene-generation-results} traces the program from authoring to
the compiled scene.

\begin{figure}[!htbp]
  \centering
  \input{sec/assets_scene/figures/fig_vnext_schema_authoring}
  \caption{\textbf{Semantic scene program and compilation.}
  A semantic scene program can be edited visually and compiled against selected
  geometry into a concrete scene.}
  \label{fig:scene-generation-results}
\end{figure}

The scene program separates object-local geometry, placement ownership, grouped
composition, cross-domain relations, and geometry-changing alternatives because
each contributes a different constraint to compilation.

\paragraph{Scene Program Constructs}
Together, these constructs encode task intent in a form that the compiler can
resolve against selected geometry.

Objects and object-local anchors define the scene's reusable entities and
reference frames.  Each background or object has a stable scene identity, task
role, and binding to either a fixed asset or an authored, role-compatible
candidate set.
Object-local surfaces and volumes expose usable geometry such as tabletops,
shelves, trays, drawers, and receptacle interiors, so dependent placements resolve
in the support's current local frame.  An instance may refine scale and semantic
tags without changing the asset package.

Placement ownership assigns each placed node exactly one owner---the node itself
or a joint group---to prevent conflicting poses.  Surface, volume, and discrete
pose domains encode admissible planar, contained, or complete task-specific
configurations rather than already sampled poses.

Grouped composition represents multi-object arrangements that cannot be reduced
to independent placements.  A \texttt{region} distributes objects over a shared
support,
\texttt{scatter} represents denser or repeated collections, \texttt{volume}
composes objects inside a receptacle, and \texttt{slot} binds nodes to authored
locations; the group owns the correlated placement decision.

Spatial relations express directional constraints across placement owners,
separate from support and containment.  Relations such as \texttt{left\_of},
\texttt{right\_of}, \texttt{front\_of}, and \texttt{behind} retain their meaning as
support sizes and object footprints change.

Scene variants declare asset, background, and slot alternatives within one
program.
Geometry-changing variants are resolved before placement because their
footprints, support fit, and collision shapes alter the admissible domains; only
compatible alternatives are carried into runtime-plan construction.

\paragraph{Geometry-aware Compilation}
The compiler turns these declarations into concrete placements through two
complementary mechanisms: dependency analysis determines coupling and solve
order, while transactional solving realizes a consistent assignment.

Placement coupling and dependency analysis constructs a typed interaction graph
over node and group domains.
Footprint-expanded overlap edges define jointly solved components because one
placement can remove space from another.  Directional relations remain explicit
constraints and scheduling dependencies without merging independent components,
while object-local support and containment references add parent--child
dependencies and yield an owner-first schedule.

Transactional component solving first evaluates each coupled component through a
temporary assignment.  Forward checking rejects choices that empty dependent
domains, while incremental validation tests
activated support, containment, group, directional, collision, and clearance
constraints.  The compiler commits a complete assignment atomically; otherwise
it rolls back all provisional poses and explores another alternative through
bounded backtracking.  Final
validation precedes export, and exhausted search returns a diagnostic tied to the
unsatisfied constraint and placement context.

\paragraph{Visual and Language-guided Authoring}
The visual editor and language-guided workflow both edit the same semantic scene
program and invoke the same compiler.  The editor exposes anchors, domains,
groups, and relations in perspective and top-down views while previewing the
compiled scene (Fig.~\ref{fig:scene-generation-results}(b,c)).  Following prior
language-guided scene-construction systems
\cite{yang2024holodeck,xia2026sage}, the X2Real authoring agent inspects asset
metadata, visual previews, USD geometry, renders from selected cameras, and
compiler diagnostics, then uses this evidence to revise asset selection,
ownership, domains, groups, or relations.

\FloatBarrier

%% file: sec/assets_scene/figures/fig_vnext_schema_authoring.tex
\includegraphics[width=.95\linewidth]{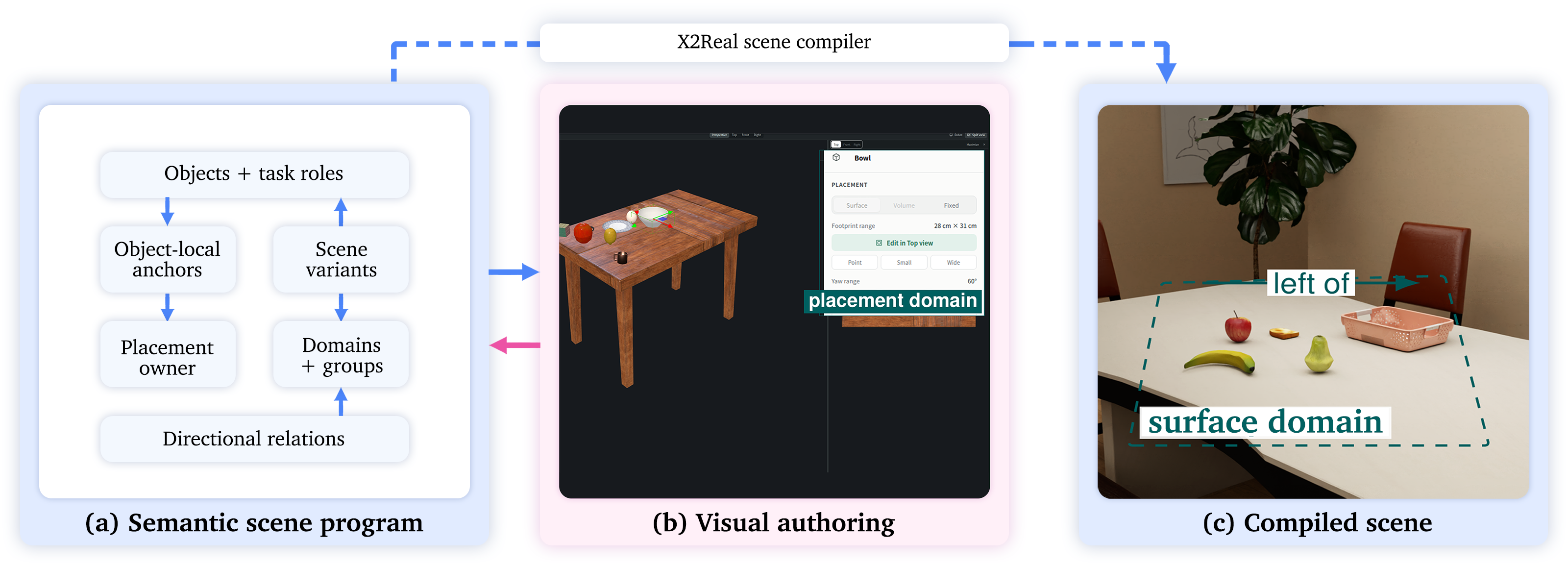}

%% file: sec/assets_scene/sections/30_runtime_diversity.tex
\subsubsection{Task-conditioned Runtime Variation and State-aligned Replay}
\label{sec:runtime-diversity}

X2Real supports two forms of content variation with different preservation
boundaries.  Reset-time variation samples compiled alternatives while
preserving authored task structure, whereas replay recomposes static context
while preserving recorded states and timestamps.  The reset-time path follows a
compile-once, sample-many principle: offline compilation resolves admissible
alternatives, and each episode reset samples from the resulting plan.
Figure~\ref{fig:runtime-sampling} summarizes this path.

\begin{figure}[!t]
  \centering
  \input{sec/assets_scene/figures/fig_runtime_sampling_plan}
  \caption{\textbf{Runtime variation compilation and sampling.}
  The compiler consolidates feasible alternatives into a reset-time plan;
  examples show coupled layouts and environment-lighting variation.}
  \label{fig:runtime-sampling}
\end{figure}

For episode reset, the compiler emits a \texttt{RuntimeSamplingPlan} containing
coupled discrete layouts, continuous object-local surface and volume domains,
allowed orientations, dependency order, and compatible asset, paired-material,
and lighting variants.  Coupled layouts capture cross-object choices that must be
selected jointly, while local domains preserve residual pose freedom within each
choice.

The compiler populates the plan by generating candidates from concrete geometry
and feasible domains and coupling the choices that interact
(Section~\ref{sec:semantic-scene-layout}).  Within each support, a
geometry-conditioned generator proposes layouts for placement domains whose
feasible regions overlap.  The compiler rejects static-constraint violations,
removes geometric duplicates, and retains a compact pool balancing geometric
quality and layout diversity.  Overlapping footprint-expanded domains form joint
case pools, object-local support dependencies define parent-before-child
sampling, and task-specific restrictions filter or replace scene-level
alternatives.

At reset, the runtime realizes the plan by selecting a coupled layout, sampling
continuous poses in dependency order, and applying the associated asset,
paired-material, and lighting variants.  Geometry-changing choices are resolved
during plan construction.  Appearance adjustments modify bounded rendering
channels while preserving the physical-material record.  Paired substitutions
atomically select authored rendering and physical-material records and update
applicable mass, inertia, friction, and support-contact properties.

\FloatBarrier
\paragraph{Swept-volume-aware Replay Augmentation}

A recorded episode supplies robot and task-object states, benchmark-camera
transforms, and timestamps, while a selected scene variant supplies alternative
static assets, appearance, and lighting.  Replay uses these inputs to
recompose observations for training-data augmentation \cite{liu2026pipette}.

\begin{figure}[!t]
  \centering
  \input{sec/assets_scene/figures/fig_replay_pipeline}
  \caption{\textbf{State-aligned replay augmentation.}
  Recorded motion defines a swept envelope for static-context recomposition while
  preserving the state-time stream; panels (b) and (c) show source and recomposed
  views from the same recorded instant.}
  \label{fig:controlled-replay-results}
\end{figure}

Forward kinematics reconstructs robot-link motion, while recorded poses provide
the task-object trajectories.  Geometric proxies, adaptive temporal sampling,
and error-bounded path
compression capture the trajectory where endpoint interpolation is insufficient.
Inflating these paths by spatial extent, approximation error, and clearance yields
a swept exclusion envelope.  Inserted objects must lie on valid support, outside
the envelope and existing obstacles, and satisfy pairwise clearance; stable
orientations and support offsets are used when available.

The procedure composes the recorded robot, task-object, and camera transforms with
the new static scene layer at their original timestamps.  This preserves
kinematic state--time alignment while changing the observation context for
training-data augmentation.
Figure~\ref{fig:controlled-replay-results} illustrates the recomposition;
implementation details for runtime-plan construction and replay geometry appear in
Appendix~\ref{app:runtime-replay-details}.
\FloatBarrier

%% file: sec/assets_scene/figures/fig_runtime_sampling_plan.tex
\includegraphics[width=.95\linewidth]{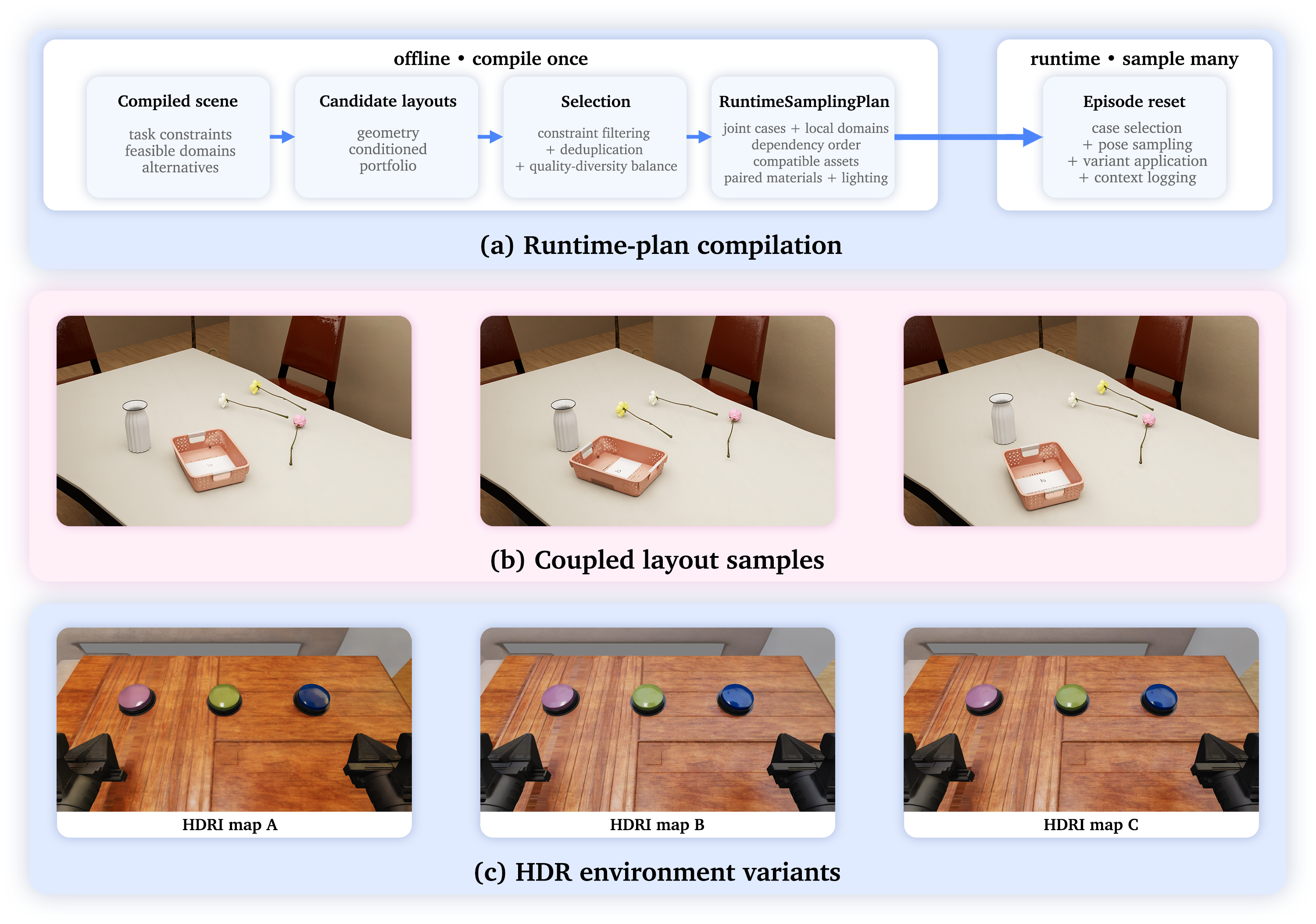}

%% file: sec/assets_scene/figures/fig_replay_pipeline.tex
\includegraphics[width=.95\linewidth]{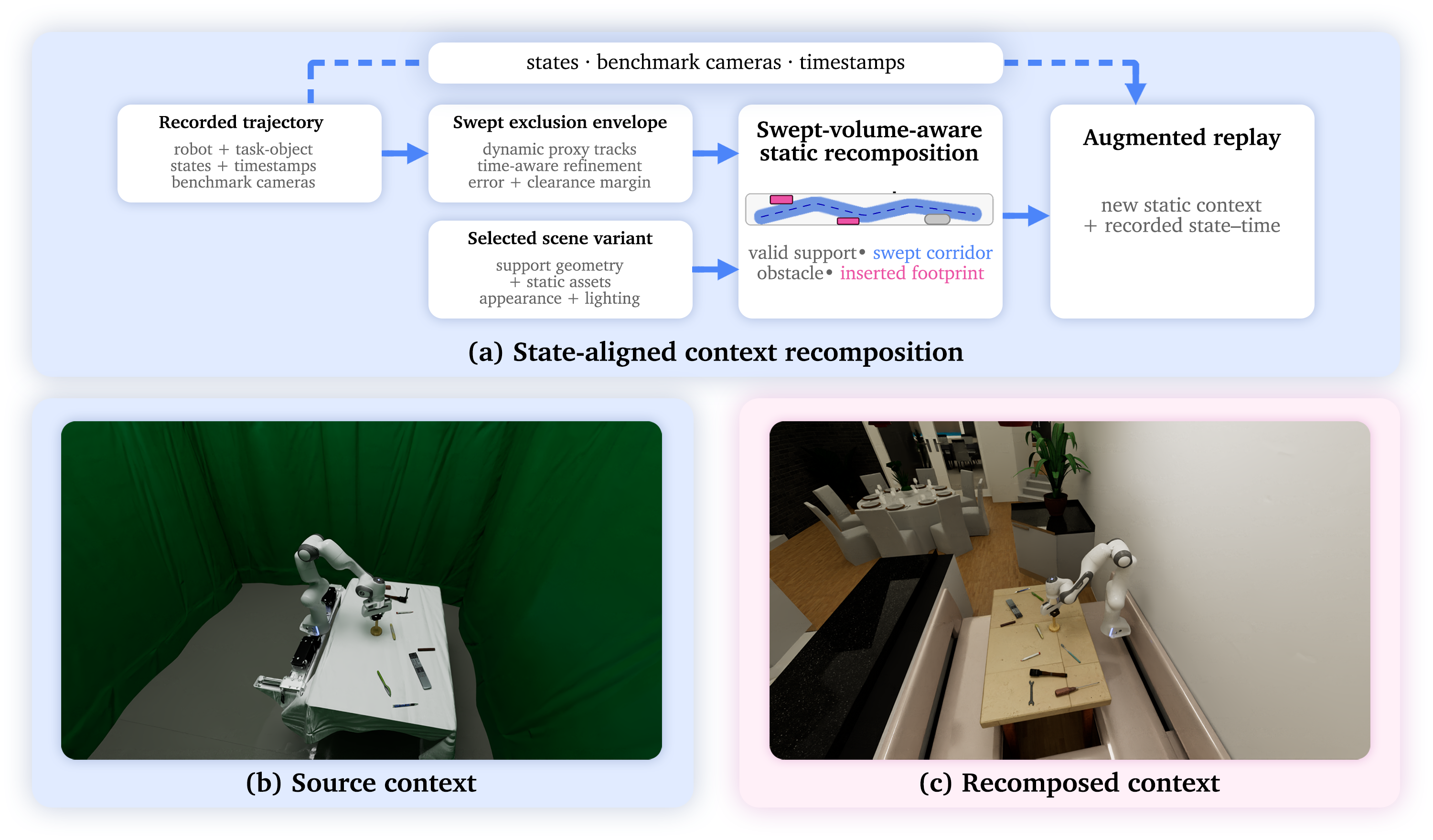}

%% file: sec/atomic_task_skill_composition/content.tex

Robot manipulation tasks can typically be decomposed into multiple stages with explicit semantics and dependency relations. For example, placing a medicine bottle in a drawer and then closing the drawer involves several interdependent stages, including opening the drawer, picking up the bottle, placing it inside, and closing the drawer. However, conventional task definitions and evaluation protocols usually provide only a final success signal for the entire task, without a unified description of the intermediate process. This makes it difficult to determine which operations a policy has completed, how far it has progressed, or at which stage it failed. Collected demonstration trajectories also lack directly usable stage boundaries and subtask labels. Although these annotations can be obtained through manual frame-wise labeling or post-hoc parsing with an auxiliary model, both approaches are costly and make it difficult to maintain consistent and verifiable process semantics across tasks. In addition, automatic demonstration collection uses atomic skills such as picking, placing, and pushing to complete stage-level subtasks, and composes them according to their dependencies into complete demonstration trajectories.

To address this limitation, we represent each manipulation task as an executable \textbf{Directed Acyclic Graph} (DAG) composed of atomic tasks. Each node defines a verifiable atomic task, i.e., a local physical outcome that the robot must achieve, while each edge specifies a dependency between atomic tasks. The same task graph enables the system to record stage-completion events online and generate subtask labels for teleoperated demonstrations. It also allows DAG nodes to be paired with atomic skills for motion-planning-based demonstration synthesis, with the generated outcomes verified by node success conditions. During evaluation, the graph further provides subtask completion, task progress, and failure diagnostics beyond final task success.

\subsubsection{Executable Task DAG}
\label{sec:executable-task-dag}

\begin{figure}[t]
    \centering
    \includegraphics[width=\linewidth]{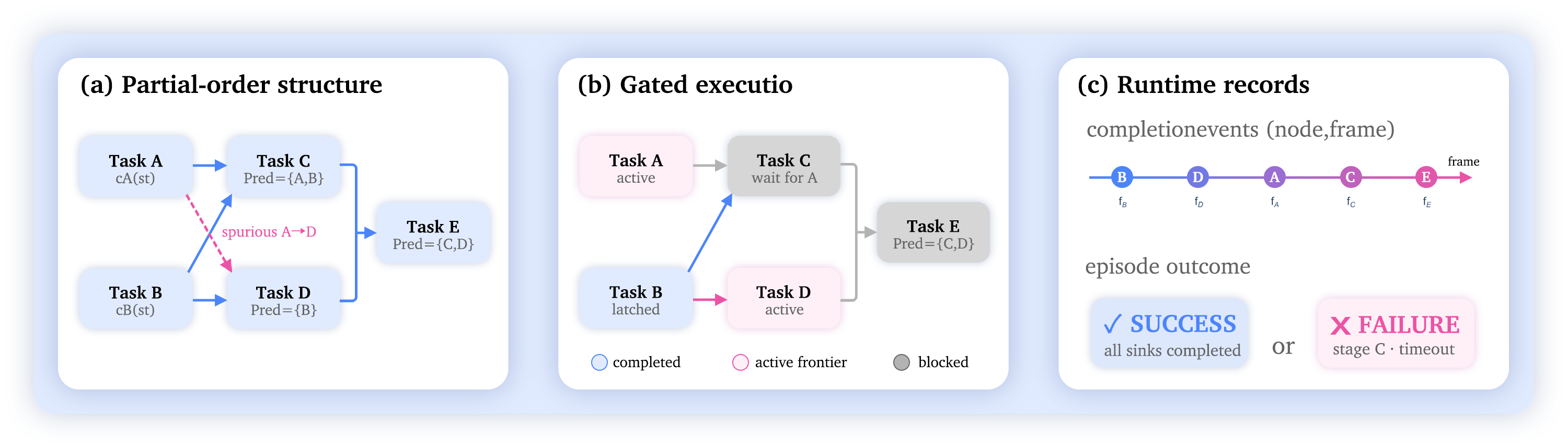}
    \caption{\textbf{Executable Task DAG.} (a) The Task DAG represents only the necessary partial-order dependencies and does not impose an order on independent nodes. (b) At runtime, a node becomes active only after all of its predecessors have completed, and its completion state is latched when its success predicate is first satisfied. (c) Node-completion events record the node ID and first completion frame, supporting task-success determination, progress measurement, and failure attribution. The success and failure records illustrate two alternative episode outcomes rather than simultaneous results from the same execution.}
    \label{fig:executable-task-dag}
\end{figure}

We represent a manipulation task as a directed acyclic graph $G=(V,E)$. Each node $v\in V$ defines an atomic task and is associated with a simulator-state-based success predicate $c_v(s_t)$, which evaluates physical conditions such as object poses, contact relations, grasp states, or joint states. Each edge $(u,v)\in E$ indicates that node $v$ depends on the completion of node $u$. Unlike a fixed step list, a DAG specifies only the necessary precedence constraints and imposes no order between independent stages. As illustrated in Fig.~\ref{fig:executable-task-dag}(a), nodes $A$ and $B$ can be completed independently; node $C$ waits for both $A$ and $B$, whereas node $D$ depends only on $B$; node $E$ becomes active only after both $C$ and $D$ are complete. Consequently, when $B$ is complete but $A$ is not, $D$ can already be executed while $C$ must still wait for $A$. A fixed linear sequence would impose unnecessary ordering constraints. Recursively organizing the same task with only sequential and parallel stages would likewise require either introducing an additional $A\rightarrow D$ constraint or duplicating the shared predecessor $B$. A DAG represents these dependencies directly.

During execution, a node success predicate is not evaluated unconditionally. The system activates a node and evaluates its predicate only after all of its predecessors have completed. Let $d_v^t$ indicate whether node $v$ has been completed by time $t$. Its gate and latched completion state are defined as

\begin{equation}
g_v^t=\bigwedge_{u\in\mathrm{Pred}(v)}d_u^t,
\qquad
d_v^t=d_v^{t-1}\lor\left(g_v^t\land c_v(s_t)\right).
\label{eq:task-dag-gate-latch}
\end{equation}

Here, $g_v^t$ indicates whether the dependency conditions of node $v$ have been satisfied. At time $t$, the unfinished nodes with open gates form the active node set, and the system evaluates their success predicates in parallel at every simulation step. The set of nodes newly completed at the current frame is

\begin{equation}
\Delta D_t=
\left\{
v\in V
\mid
\neg d_v^{t-1}\land g_v^t\land c_v(s_t)
\right\}.
\label{eq:newly-completed-nodes}
\end{equation}

For every $v\in\Delta D_t$, the system emits a node-completion event $(v,t)$. In Fig.~\ref{fig:executable-task-dag}(b), when $B$ is complete but $A$ is not, $D$ becomes active because it depends only on $B$, whereas $C$ continues to wait for $A$. Gating prevents a downstream stage from being marked complete merely because its predicate happens to hold in the initial state. Once a node is completed, its state is latched, converting a transient physical condition into a persistent stage milestone.

Completing an individual node indicates only that one stage objective has been achieved and does not terminate the episode by itself. After a node is completed, the system activates its successors according to the graph dependencies. The overall task is considered successful only when all sink nodes have completed and no stage-level failure has been triggered:

\begin{equation}
S^t=
\neg F^t
\land
\bigwedge_{v\in\mathrm{Sink}(G)} d_v^t,
\label{eq:task-dag-success}
\end{equation}

where $F^t$ denotes a node failure predicate, stage timeout, or another task-level failure condition. Because a sink node can become active only after all of its predecessors have completed, this criterion also guarantees completion of the dependency stages leading to the final objectives.

The Task DAG further converts changes in node state into recordable runtime events. When a latched node state first changes from incomplete to complete, the system records the node ID and first completion frame. The sequence $B\rightarrow D\rightarrow A\rightarrow C\rightarrow E$ in Fig.~\ref{fig:executable-task-dag}(c) is one valid runtime completion order induced by the same DAG, rather than a fixed sequence prescribed by the task. When a node failure predicate or stage timeout is triggered, the system additionally records the first failed node, the failure frame, and the corresponding reason. Each episode therefore contains not only a final success signal, but also node-level completion states, first completion frames, the failed stage, and the failure reason. These runtime records are subsequently used for stage-level demonstration annotation and fine-grained policy evaluation.

\subsubsection{DAG-guided Demonstration Collection}
\label{sec:dag-guided-demonstration-collection}

X2Real supports two sources of Task-DAG-guided demonstrations: real-time teleoperation and atomic-skill-based automatic synthesis. The two collection pipelines share the same scene generation, task definitions, DAG runtime, evaluation module, trajectory recorder, and exported data format. Their primary difference lies in how control trajectories are produced. The teleoperation pipeline maps control signals from a teleoperation device to robot control targets, whereas the automatic pipeline uses atomic skills bound to DAG nodes to generate targets and a motion planner to solve the corresponding execution trajectories. Consequently, both sources produce demonstrations with consistent observation--action structures, stage semantics, and node-level evaluation results.

\paragraph{Teleoperated Demonstrations}
X2Real provides a unified teleoperation interface supporting physical master arms, VR controllers, and keyboard input. The human demonstrations in the benchmark are collected through physical master--slave teleoperation. An operator uses the physical master arms to control a simulated dual-arm robot. The collection loop receives master-device states asynchronously and sends dual-arm commands to the simulated robot at a control frequency of $30\,\mathrm{Hz}$ while providing real-time visual feedback to the operator. Within the same control loop, the system synchronously records camera observations, robot states, actions, and DAG runtime states, producing complete trajectories aligned by simulation step.

During trajectory execution, the DAG runtime uses privileged simulator state to evaluate the success predicates of the currently active nodes online. When the completion state of node $v$ first changes from false to true, the system records the stage-completion event

\begin{equation}
e_v=(v,f_v),
\qquad
f_v=\min\{t\mid d_v^t=1\},
\label{eq:node-completion-event}
\end{equation}

where $f_v$ is the first completion frame of node $v$. Let $\pi_\tau=(v_1,\ldots,v_K)$ denote the temporally ordered node-completion events observed in a demonstration, and let $f_{v_0}=0$. For a trajectory in which stages are executed sequentially, the segment between the completion of the preceding node and the first completion of node $v_k$ is assigned the subtask label $v_k$:

\begin{equation}
z_t=v_k,
\qquad
f_{v_{k-1}}<t\leq f_{v_k}.
\label{eq:dag-subtask-label}
\end{equation}

\begin{figure}[t]
    \centering
    \includegraphics[width=\linewidth]{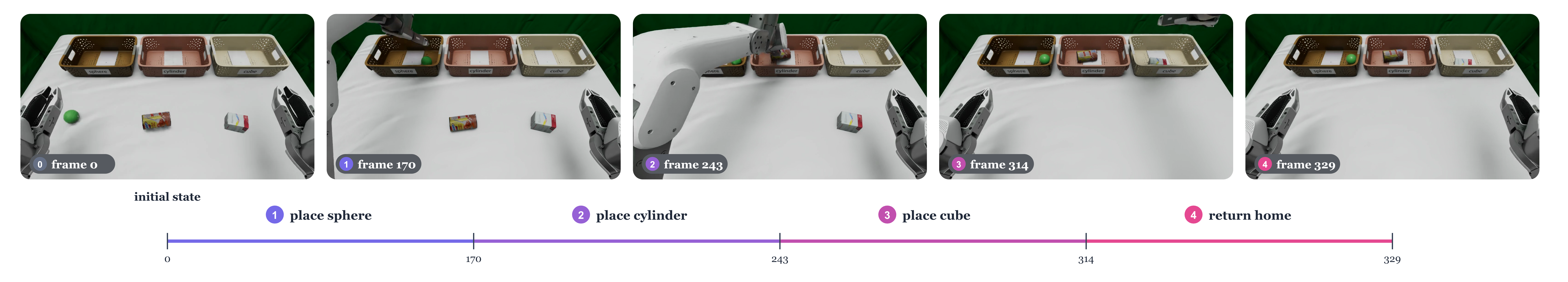}
    \caption{\textbf{DAG-guided Teleoperation Labeling.} The first completion frames of the Task DAG nodes divide a Classify by Shape demonstration into trajectory segments with atomic-task labels.}
    \label{fig:dag-teleoperation-labeling}
\end{figure}

Each trajectory segment is labeled by the atomic task newly completed at its endpoint and thus reflects the physical outcome actually achieved by that segment. Because the first completion frames are determined online from simulator state and node success predicates, neither manual boundary annotation nor post-hoc parsing with an auxiliary model is required. In the Classify by Shape example in Fig.~\ref{fig:dag-teleoperation-labeling}, the three nodes for placing the sphere, cylinder, and cube have no dependencies between them and may therefore be completed in any order, whereas the return-home node waits for all three placement nodes. The Task DAG does not prescribe the placement order; the figure shows one valid order produced by this particular demonstration.

\paragraph{Automatic Data Synthesis}
X2Real constructs a closed-loop data-generation process driven by the Task DAG. At runtime, the DAG determines the currently active nodes from task progress. The atomic skill sequence bound to a selected node generates sparse motion targets from the current scene state, and a motion planner expands these targets into an executable trajectory. The node success predicate then independently verifies the resulting physical outcome. Upon successful verification, the system latches the node completion state and activates successor nodes whose dependency conditions have become satisfied. If skill generation, motion planning, or predicate verification fails, the system resamples the target or motion path. Through this closed loop, the same Task DAG can produce demonstrations with different node orders, skill parameters, and motion trajectories while preserving the same task semantics.

Specifically, each node $v$ is associated with an ordered sequence of parameterized atomic skills,

\begin{equation}
\mathcal{K}_v=
\left[k_{v,1}(\theta_{v,1}),\ldots,k_{v,m_v}(\theta_{v,m_v})\right],
\label{eq:node-skill-sequence}
\end{equation}

where $m_v$ is the number of skills associated with node $v$, and $k_{v,i}$ and $\theta_{v,i}$ denote the $i$-th atomic skill and its parameters, respectively. The node and its success predicate specify the physical outcome to be achieved, i.e., \emph{what} constitutes completion, whereas the skill sequence $\mathcal{K}_v$ describes \emph{how} the robot attempts to achieve that outcome. The relationship is not one-to-one: a node may compose multiple sequential skills, while the same skill can be reused across nodes and tasks through different object roles and parameters. Different target samples and motion plans can also yield multiple valid realizations of the same node.

For node scheduling, the collector computes the active node set from the completed nodes and selects a node $v$ whose skill sequence will be instantiated. Nodes without dependencies between them can be active simultaneously, allowing the collector to choose the next node and thereby produce different valid execution orders. A successor node becomes active only after all of its predecessors have completed.

After selecting a node, the collector instantiates the atomic skills in $\mathcal{K}_v$ sequentially. An atomic skill exposes a unified target-generation interface but does not directly solve inverse kinematics or issue low-level control commands. It reads the current robot state together with privileged simulator state for objects, joints, and target regions, and generates a sparse target sequence with stage semantics. Each target may specify an end-effector position and orientation, gripper state, execution phase, arm selection, and optional auxiliary targets. A collision-aware motion planner then expands the sparse targets into a dense joint-control trajectory. Our current implementation uses cuRobo for inverse kinematics and trajectory optimization, after which an executor tracks the planned trajectory in closed loop. For example, \texttt{pick} generates targets for approach, descent, gripper closure, and lifting, whereas \texttt{place} generates targets for pre-placement, placement, release, and retraction. Different execution phases may use different motion-planning strategies and motion speeds.

Skill parameters are decoupled from specific scenes through semantic roles defined by the task. For example, \texttt{@pickable} and \texttt{@destination} denote the manipulated object and target region in the current episode, respectively, and are automatically bound to concrete asset instances after scene resampling. The same \texttt{pick} and \texttt{place} skills can therefore be reused across tasks with different object and target parameters. Within a node, multiple atomic skills can form a sequential skill sequence, such as \texttt{pick} $\rightarrow$ \texttt{place} for achieving the outcome that an object lies within a target region. Across the task, the skill sequences associated with different nodes are scheduled according to the Task DAG dependencies to complete the full task.

Completing motion planning does not imply completion of the atomic task. After executing a candidate trajectory, the system evaluates the node's own success predicate against the resulting physical state. The node and its trajectory segment are accepted only when the predicate is first satisfied and the completion state is latched. If the target is unreachable, motion planning fails, a collision occurs during execution, or the node predicate remains unsatisfied after execution, the collector records the failed stage and reason and resamples the skill target or motion path. Atomic skills and the motion planner therefore propose candidate behaviors for \emph{how} to perform the task, while the Task DAG independently determines whether the desired physical outcome has actually been achieved. This separation between generation and verification ensures that automatically synthesized demonstrations are validated using the same geometric, contact, and joint-state conditions used in subsequent policy evaluation, rather than being declared successful by the generator itself.

This closed loop also provides behavior-level data augmentation. For the same node, the system can resample scene layouts and object instances, grasp candidates and approach directions, placement poses, arm assignments, skill parameters, and collision-free motion paths. For tasks with multiple simultaneously active nodes, it can also sample different valid node orders. Each sample undergoes a new reachability check and motion-planning process, and its outcome is verified by the same node predicate. The resulting demonstrations may therefore differ in robot states, action sequences, trajectory durations, and intermediate paths while sharing the same atomic-task semantics.

For tasks containing multiple stages, the Task DAG additionally supports stage-wise collection, failure recovery, and trajectory stitching. After a node is completed, the system saves its trajectory segment, node-completion event, and resulting scene checkpoint. If a subsequent node fails, the collector restores the most recent successful checkpoint and resamples and replans only the current failed node, rather than replaying the completed prefix from the initial state. Successful segments retain lineage information such as their node IDs, parent samples, and completion progress. The system finally removes seam frames introduced by checkpoint restoration and stitches the segments according to the actual execution order to form a complete demonstration. This mechanism localizes the cost of failures in later stages and allows a successful prefix to connect to different successor targets and motion paths, yielding traceable failure-recovery and trajectory-recomposition data.

\begin{figure}[t]
    \centering
    \includegraphics[width=\linewidth]{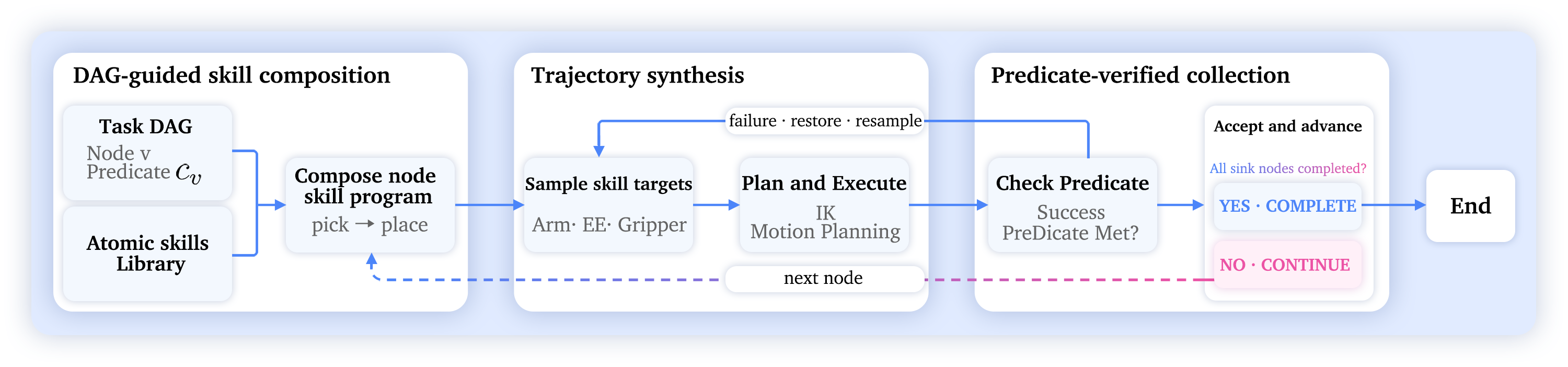}
    \caption{\textbf{DAG-guided Data Synthesis.} The system selects an active node, instantiates its skill sequence to generate sparse targets, and plans and executes the resulting trajectory. The node success predicate independently verifies the outcome. On success, the system latches the node state, saves the trajectory segment and scene checkpoint, and updates the Task DAG; on failure, it restores the latest valid checkpoint and resamples the target or motion path. Once all sink nodes have completed, the stage-wise verified segments form a complete demonstration.}
    \label{fig:dag-automatic-synthesis}
\end{figure}

\subsubsection{DAG-guided Evaluation}
\label{sec:dag-guided-evaluation}

Conventional robot-manipulation evaluation typically reduces an episode to a binary final outcome. Although this metric directly indicates whether a policy completes the entire task, it cannot distinguish between failures that occur before a critical operation begins and those that occur near the end after most stages have been completed. Nor can it identify the specific stage that limits overall performance. During a policy rollout, X2Real executes the same Task DAG used in the task definition and directly derives fine-grained evaluation metrics from node completion states, first completion events, and failure events, without additional manual annotation or post-hoc video parsing. Final task success remains the primary metric, while node-level metrics measure partial progress and diagnose failures.

\paragraph{Final Task Success.}
For an episode $\tau$, let $T_\tau$ denote its termination time. Following the definition in Sec.~\ref{sec:executable-task-dag}, the episode is successful only if all sink nodes have completed and no failure condition has been triggered:

\begin{equation}
S_\tau=
\neg F_\tau^{T_\tau}
\land
\bigwedge_{v\in\mathrm{Sink}(G_\tau)}
d_{v,\tau}^{T_\tau}.
\label{eq:episode-task-success}
\end{equation}

Over a set of $N$ evaluation episodes, the final task success rate is

\begin{equation}
\mathrm{SuccessRate}
=\frac{1}{N}\sum_{\tau=1}^{N} S_\tau.
\label{eq:final-task-success-rate}
\end{equation}

This metric requires the policy to complete every dependency stage leading to the sink nodes. It therefore cannot be triggered by a final-state condition that happens to hold at initialization, nor does partial node completion cause premature success.

\paragraph{Subtask Completion Rate.}
Let $\mathcal{I}_v$ denote the set of evaluation episodes that contain node $v$. The completion rate of subtask $v$ is

\begin{equation}
C_v=
\frac{1}{|\mathcal{I}_v|}
\sum_{\tau\in\mathcal{I}_v}
d_{v,\tau}^{T_\tau}.
\label{eq:subtask-completion-rate}
\end{equation}

$C_v$ is the fraction of rollouts in which the policy achieves the physical outcome defined by node $v$. Compared with reporting only overall success, per-node completion rates reveal which manipulation stages the policy can complete reliably and which nodes constitute the largest performance bottlenecks. We report this metric separately for each node to compare the reliability of different stages within the same task.

\paragraph{Task Progress.}
X2Real assigns a stage score $w_v$ to every node in a Task DAG such that the scores within each task sum to 10:

\begin{equation}
\sum_{v\in V_\tau}w_v=10.
\label{eq:task-dag-score-sum}
\end{equation}

Node scores are assigned according to each stage's contribution and semantic importance to the overall task objective: nodes representing critical task outcomes receive higher scores, while auxiliary stages receive lower scores. All X2Real tasks use this ten-point node-scoring scheme, providing a common progress scale across tasks with different graph structures and numbers of nodes. For episode $\tau$, the cumulative task score and normalized task progress at time $t$ are defined as

\begin{equation}
Q_\tau^t=
\sum_{v\in V_\tau}w_vd_{v,\tau}^t,
\qquad
P_\tau^t=\frac{Q_\tau^t}{10}.
\label{eq:task-progress}
\end{equation}

Because node completion states are latched, both the cumulative score $Q_\tau^t$ and task progress $P_\tau^t$ are monotonically nondecreasing throughout execution. At termination, $Q_\tau^{T_\tau}\in[0,10]$ provides a partial-completion score for an unsuccessful episode, whereas an episode that completes every node receives a score of 10. The subtask completion rate $C_v$ measures the reliability of a semantic node across multiple rollouts, while $P_\tau^t$ describes the important stages completed within an individual rollout and their cumulative contribution at a given time.

\paragraph{Failure Case Analysis.}
For an unsuccessful episode, the DAG runtime records the node associated with the first failure event, its frame index, and the corresponding reason. Failure reasons arise from node-level failure predicates, stage timeouts, or other task-level failure conditions. If an episode terminates because of a task-level timeout without emitting a node-level failure event, the system additionally stores the set of currently active nodes that remain incomplete at termination. These records support a two-dimensional distribution over failure node and failure reason. Together with first completion frames, they distinguish episodes that never reached a stage, reached but failed to complete it, or completed it before failing at a successor stage.

%% file: sec/5_experiments.tex
\section{Experiments and Results}\label{sec:experiments}

\subsection{Sim2real Alignment}

\label{sec:exp-sim2real}

\subsubsection{System Identification}
To reduce the discrepancy in end-effector tracking behavior between the real and simulated robots, we calibrate the control parameters of the simulator by matching end-effector trajectories. Real-robot trajectories are collected during teleoperation and form a dataset. The corresponding end-effector pose commands are replayed in simulation in a time-synchronized manner, converted into joint targets via inverse kinematics, and executed using the Stable Proportional--Derivative (StablePD) controller~\citep{tan2011stable}. The proportional gain \(K_p\) and derivative gain \(K_d\) are shared across all 12 bimanual arm joints. 

\begin{figure}[th]
    \centering
    \includegraphics[width=0.7\linewidth]{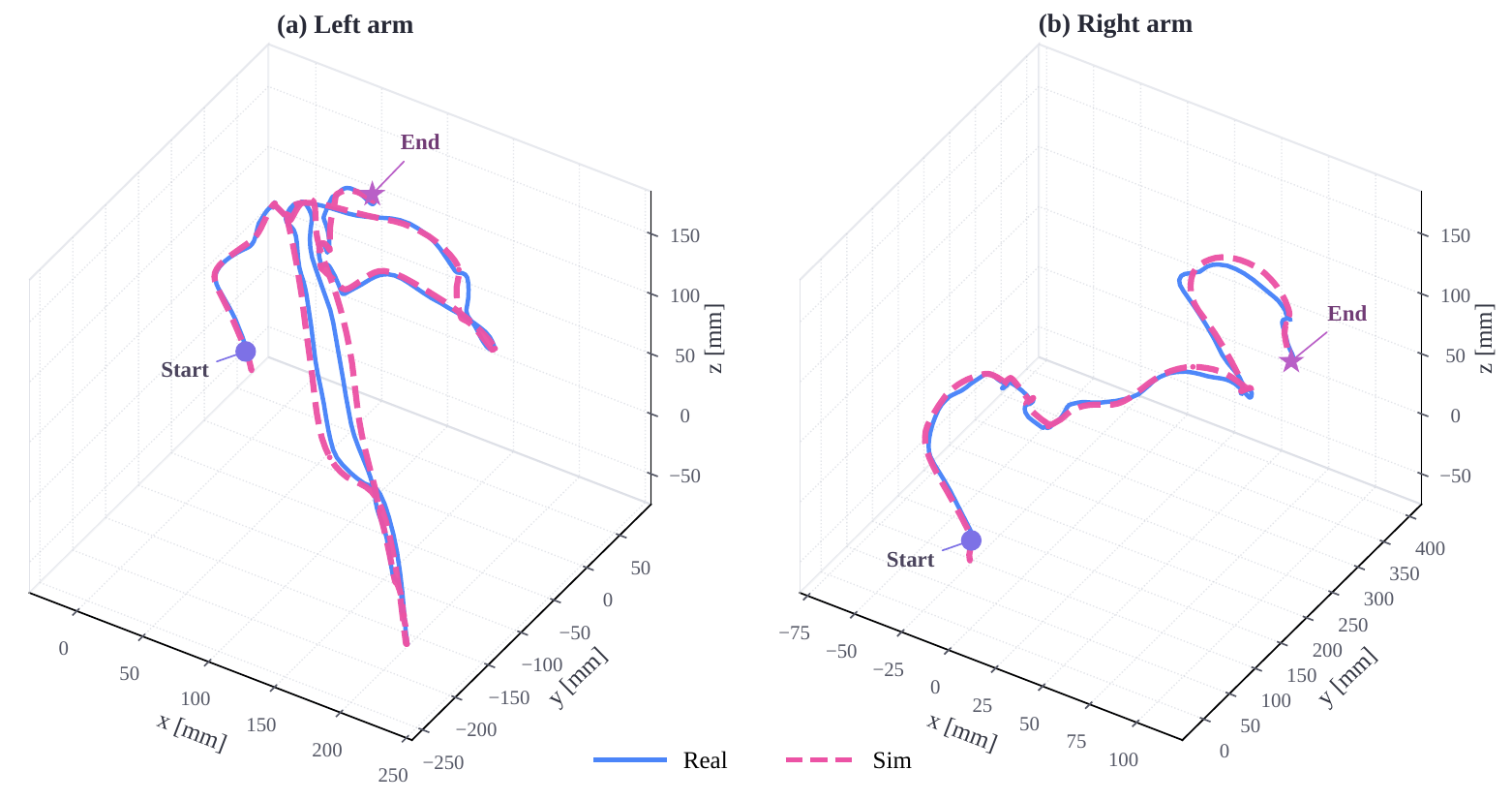}
    \caption{\textbf{Sim2real Trajectory Alignment.} Representative real and simulated end-effector trajectories on a held-out test trajectory. 
    }
    \label{fig:real2sim-ee-example}
\end{figure}

We optimize the shared gains using the Covariance Matrix Adaptation Evolution Strategy (CMA-ES)~\citep{hansen2001cmaes}. The collected trajectories are partitioned into calibration, validation, and test sets for parameter search, model selection, and final evaluation, respectively. Given a calibration set of $n$ trajectories, we solve: 
\begin{equation} 
\left(K_p^*,K_d^*\right) = \arg\min_{K_p,K_d} \frac{1}{n}\sum_{i=1}^{n}\mathcal{L}_i. 
\end{equation} 
The per-trajectory loss combines average and tail tracking errors: 
\begin{equation} 
\mathcal{L}_i = w_p\,\mathrm{RMSE}(e_{p,i}) + w_R\,\mathrm{RMSE}(e_{R,i}) + w_{p,95}\,P_{95}(e_{p,i}) + w_{R,95}\,P_{95}(e_{R,i}), 
\end{equation} 
where $w_p, w_R$ weight position and orientation root mean square error (RMSE), and $w_{p,95}, w_{R,95}$ weight their 95th-percentile errors. The percentile terms explicitly penalize intermittent large deviations that are not well captured by RMSE. The frame-wise error signals are defined as 
\begin{equation} 
e_{p,i}=10^3\|\mathbf{p}^{\mathrm{real}}-\mathbf{p}^{\mathrm{sim}}\|_2, \quad e_{R,i}=\frac{180}{\pi}\left\|\left(\operatorname{Log}\!\left((\mathbf{R}^{\mathrm{real}})^\top\mathbf{R}^{\mathrm{sim}}\right)\right)^\vee\right\|_2, \end{equation} 
where $\mathbf{p}$ and $\mathbf{R}$ are end-effector position and orientation, and $\operatorname{Log}(\cdot)^\vee$ maps relative rotations to axis-angle vectors~\citep{barfoot2017state}. The scaling factors $10^3$ and $180/\pi$ convert meters to millimeters and radians to degrees, respectively. RMSE and $P_{95}$ are computed over all frames and both arms. We set $(w_p,w_R,w_{p,95},w_{R,95})=(1,5,0.1,0.5)$ in all experiments. CMA-ES optimizes in the log-parameter space with $K_p\in[100,10\,000]$ and $K_d\in[5,500]$. Since the measured input–response delays of the real and simulated systems are comparable, trajectories are compared at synchronized frame indices without temporal alignment or delay compensation. Figure~\ref{fig:real2sim-ee-example} shows representative tracking on a held-out test trajectory.

A similar gain-only real-to-sim system identification approach is also adopted in RobotArena$\infty$~\citep{jangir2026robotarena}. Broader system-identification studies motivate a staged extension to effective actuator parameters such as armature, damping, and friction~\citep{mehta2021calibrating,bronars2026tune,bjelonic2026bridging}, while avoiding the simultaneous optimization of structurally confounded quantities~\citep{bjelonic2026bridging}. However, gains that minimize trajectory-matching error may not be optimal for downstream manipulation~\citep{bronars2026tune}: contact-rich tasks couple actuator dynamics with object properties, compliance, impacts, and unobserved forces. Task-relevant excitation and force sensing are therefore promising directions for manipulation-oriented real-to-sim system identification.

\subsubsection{Sim2real Correlation}

\begin{figure}[H]
    \centering
    \includegraphics[width=\linewidth]{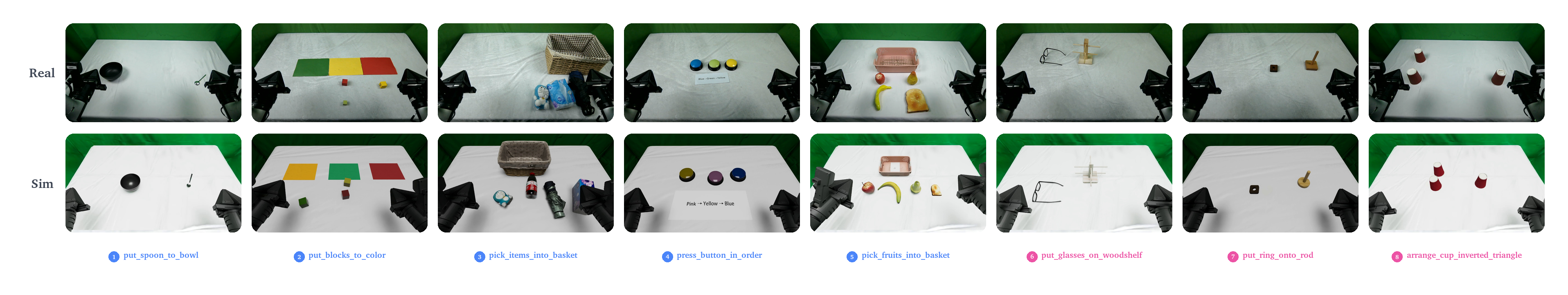}
    \caption{\textbf{Sim2real Tasks.} We create digital twins (bottom row) of 8 tasks from ManipArena~\cite{sun2026maniparena} (top row).}
    \label{fig:real2sim-tasks}
\end{figure}

To verify that our realistic simulation environment can serve as a faithful proxy for real-world evaluation, we replicate the green booth scene from ManipArena~\cite{sun2026maniparena} in simulation by manually aligning all the setting like light sources, assets, and embodiments. As shown in Fig.~\ref{fig:real2sim-tasks}, we choose 8 tasks from ManipArena that mainly contain rigid object interactions. The first 5 tasks are general manipulation tasks, including putting spoons into the bowl, putting three blocks onto matching colors, placing all items into the basket, pressing the buttons in given sequence, selecting all the fruits and placing them into the basket. The last three are harder manipulation tasks requiring control precision, including putting the glasses on the wood shelf by bimanual coordination, sliding the ring through the rod, stacking three cups into a triangle tower. 

\begin{figure}[H]
    \centering
    \includegraphics[width=0.7\linewidth]{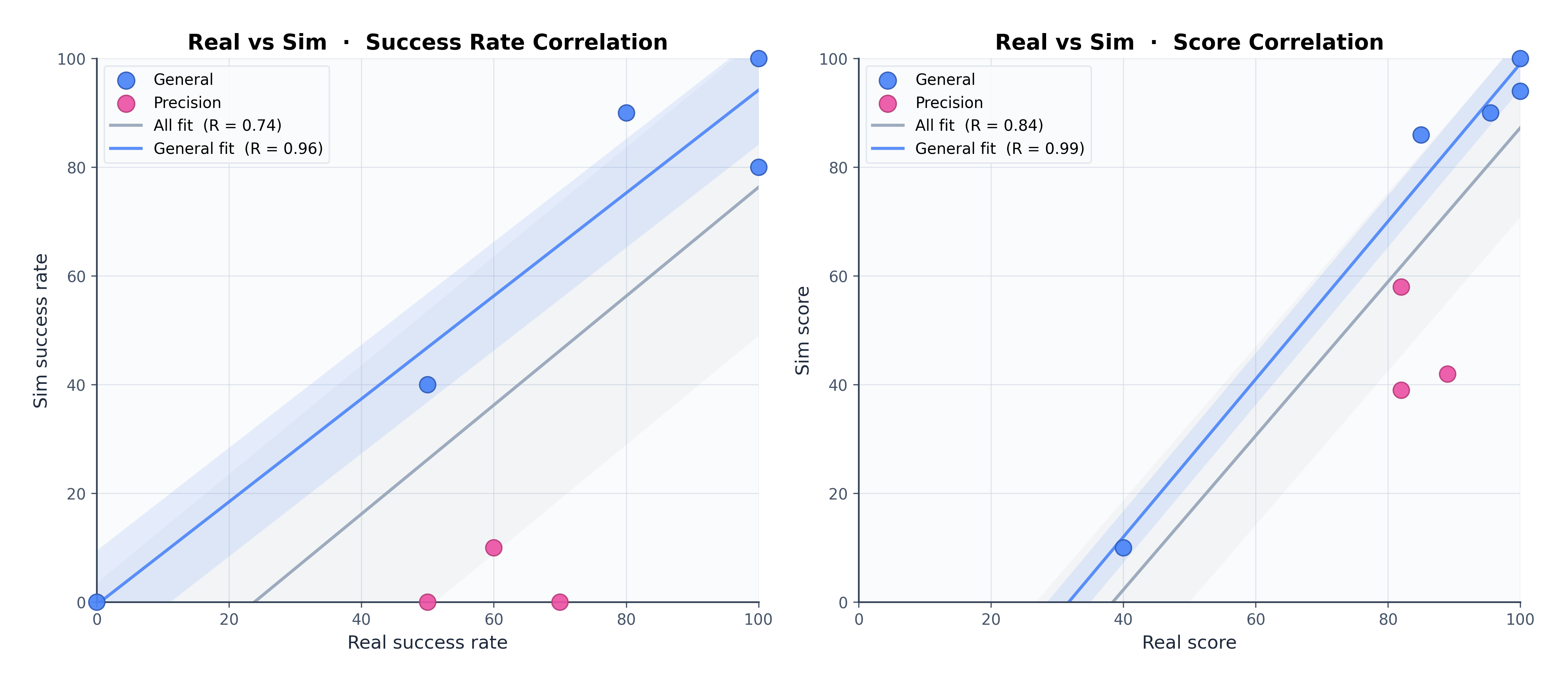}
    \caption{\textbf{Sim2real Correlation.} Wall-oss-0.5 trained on zero simulation data achieves high sim2real linear correlation in success rate and progress score.}
    \label{fig:real2sim-correlation}
\end{figure}

We select the wall-oss-0.5 model as our base model for its best performance on ManipArena. After finetuning it with real demonstration data, we directly evaluate it in our simulation environment, and compare the results with real evaluation depicted in Fig.~\ref{fig:real2sim-correlation}. Note that zero simulation data is used in the pretraining and finetuning process.


From Fig.~\ref{fig:real2sim-correlation}, we observe a strong correlation between simulation evaluation and real evaluation. For general manipulation task only, we achieve $0.96$ linear correlation coefficient for success rate and $0.99$ for progress score. However the policy struggles on three precision tasks in simulation with low success rates. Taking all tasks into account, our simulation environment achieves $0.74$ linear correlation coefficient in success rate and $0.84$ in progress score. Taking a closer look at each evaluation episode we can better understand why this gap appears.

\begin{figure}[th]
    \centering
    \begin{subfigure}{\linewidth}
        \centering
        \includegraphics[width=\linewidth]{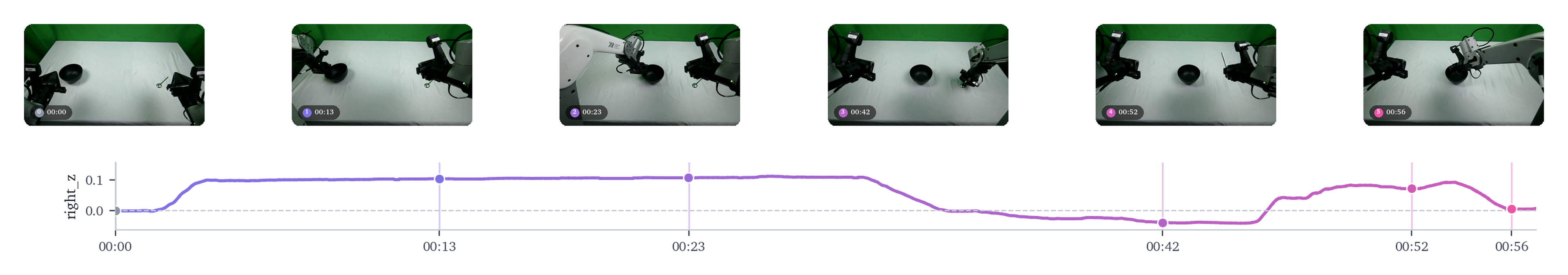}
        \caption{The real success trajectory of putting a spoon in a bowl with statistics of right gripper height.}
    \end{subfigure}
    \begin{subfigure}{\linewidth}
        \centering
        \includegraphics[width=\linewidth]{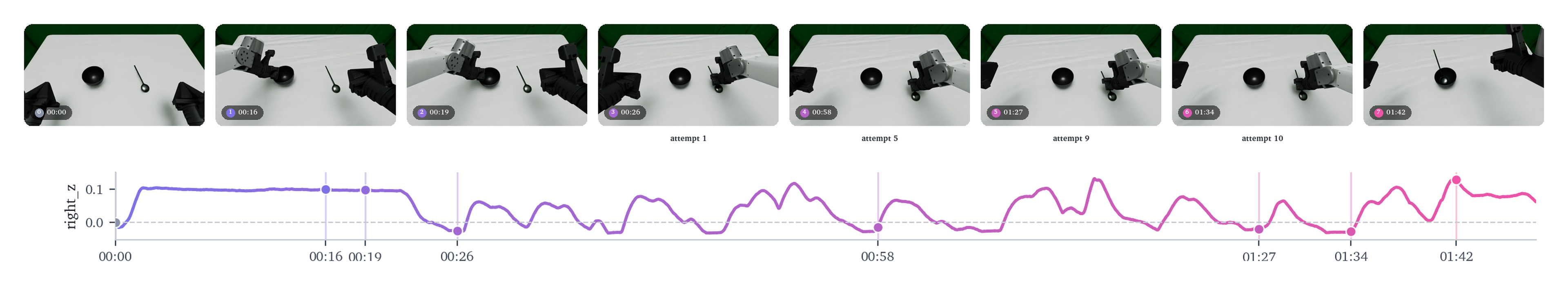}
        \caption{The simulated success trajectory of putting a spoon in a bowl with statistics of right gripper height.}
    \end{subfigure}
    \begin{subfigure}{\linewidth}
        \centering
        \includegraphics[width=\linewidth]{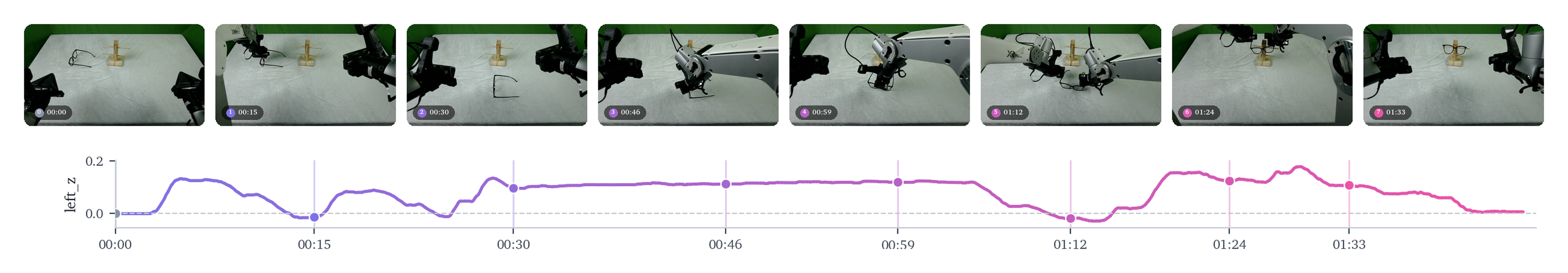}
        \caption{The real success trajectory of putting glasses on the shelf with statistics of left gripper height.}
    \end{subfigure}
    \begin{subfigure}{\linewidth}
        \centering
        \includegraphics[width=\linewidth]{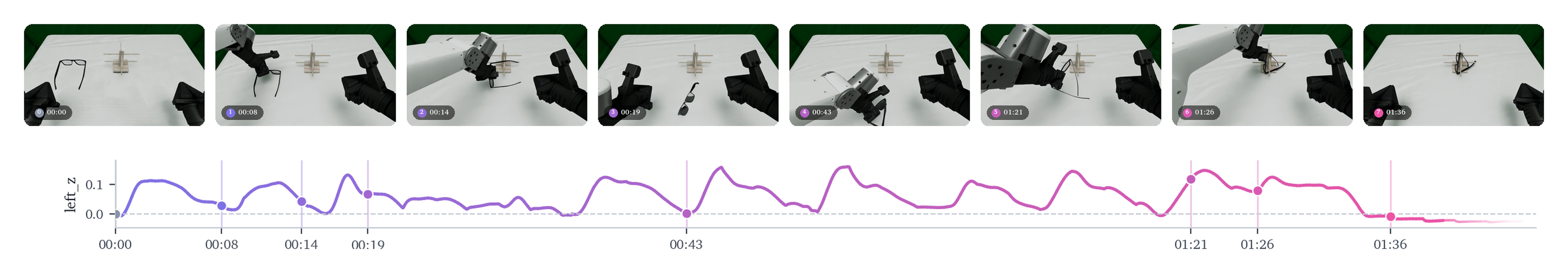}
        \caption{The simulated failure trajectory of putting glasses on the shelf with statistics of left gripper height.}
    \end{subfigure}
    \caption{\textbf{Sim2real Trajectory Analysis.} The policy trained on real data can overcome sim2real gap through multiple attempts in general manipulation tasks, but deteriorate in precision tasks due to the high failure penalty.}
    \label{fig:real2sim-task-detail}
\end{figure}

As shown in Fig.~\ref{fig:real2sim-task-detail}, we show example real and simulation evaluation trajectories for two tasks: putting the spoon into the bowl and putting glasses onto the wood shelf. For the first task, the policy achieve high success rate in both real world ($100\%$) and simulation ($80\%$), but in different ways. In real world, the robot can smoothly pick up the bowl and put it onto the center of the table, then pick and place the spoon in the first try. However in simulation, after it successfully place the bowl, the robot try for 10 times before it finally pick up the spoon, which can be visualized both from the camera and from the height of right gripper. This phenomenon does not appear solemnly but in nearly every episode. For the glasses-on-wood-shelf task, the policy has a $70\%$ success rate in reality, but $0\%$ in simulation. From the camera and the height trajectory of the left gripper in simulation, the robot still tries to finish the task through multiple attempts, but once the robot fail to properly place the glasses, the legs of the glasses might be folded, resulting into an unprecedented out-of-distribution scenario that the policy cannot handle. For other precision tasks, the failure case is similar, the policy can no longer achieve high success rate through multiple re-tries. 

These evidences prove that our simulation environment is adequate for evaluating real robot's performance, given the total score correlation $0.84$. But there still exist a real2sim gap due to the simulator, controller, or rendering, that makes real-world policies can't work perfectly in simulation as in reality. 

\subsection{Model Evaluation}

\subsubsection{Tabletop Manipulation}

We evaluate four generalist robot policies on all tabletop manipulation tasks in X2Real: Wall-OSS-0.5~\cite{yu2026walloss05}, DreamZero~\cite{ye2026dreamzero}, $\pi_{0.5}$~\cite{black2025pi05}, and our internal model wall-oss-preview. All models are post-trained using the same task demonstrations collected on the ArtiXon Arm-6A embodiment and are evaluated through the unified Policy Space protocol described above. To ensure a controlled comparison, each model is fine-tuned for approximately one epoch using only the raw observations, actions, and language instructions, without access to the frame-level subtask annotations used by the evaluator. We otherwise follow the native training and action representations of each policy:
\begin{enumerate}
\item \textbf{Wall-OSS-0.5}: EEF control, batch size 256, 20k iterations.
\item \textbf{DreamZero}: joint-space control, batch size 64, 64k iterations.
\item \textbf{$\pi_{0.5}$}: EEF control, batch size 64, 80k iterations.
\item \textbf{wall-oss-preview}: EEF control, batch size 144, 25k iterations.
\end{enumerate}
The complete results under both in-distribution (ID) and out-of-distribution (OOD) evaluation settings are reported in Tab.~\ref{tab:x2real-results}.

For each task, we report both the success rate and the progress score, with the latter measuring how far a policy progresses through the task dependency graph even when the full task is not completed. To avoid over-weighting capability dimensions containing more tasks, the final benchmark score is computed using hierarchical macro-averaging. We first average episode-level scores within each task, then average equally across tasks within each subcategory, across subcategories within each major category, and finally across the Reasoning and Manipulation categories. The same aggregation procedure is applied to success rates. Consequently, each capability dimension contributes equally to the final result regardless of the number of tasks it contains, as shown in Eq.~\ref{eq:sr-score}.

\begin{equation}
\label{eq:sr-score}
\mathrm{SR(Score)}
=
100 \times
\underset{\mathrm{Reasoning/Manipulation}}{\mathrm{Avg}}
\left[
\underset{\mathrm{List}}{\mathrm{Avg}}
\left[
\underset{\mathrm{Task}}{\mathrm{Avg}}
\left[
\underset{\mathrm{Setting}}{\mathrm{Avg}}
\left[
\underset{\mathrm{Episode}}{\mathrm{Avg}}
\left(
\mathrm{SR(Score)}
\right)
\right]
\right]
\right]
\right].
\end{equation}

Several observations emerge from the benchmark results. First, \textbf{OOD evaluation consistently exposes a substantial generalization gap}. All four policies obtain lower aggregate performance under OOD randomization than under ID evaluation. For example, wall-oss-preview drops from an overall success rate of $53.9\%$ to $34.3\%$, while its progress score decreases from $67.3$ to $56.6$. Similar degradation is observed across the other policies, although the magnitude varies substantially across individual tasks. This suggests that performance measured under training-like configurations can considerably overestimate robustness to unseen deployment conditions.

Second, \textbf{success rate alone does not fully characterize policy capability}. Policies frequently make meaningful progress without completing the entire long-horizon task. For example, DreamZero achieves only $4.85\%$ overall ID success rate but obtains a substantially higher progress score of $34.21$. This discrepancy is particularly common on long-horizon and precision-sensitive tasks, where failure at a late stage turns an otherwise largely correct trajectory into a binary failure. The progress metric therefore provides complementary information about where a policy fails along the task execution process.

Third, the results reveal several \textbf{shared capability bottlenecks across model families}, with a particularly clear gap between partial task progress and successful completion. On many challenging tasks, policies obtain non-trivial DAG scores despite very low success rates. For example, contact-rich manipulation tasks such as \texttt{ring\_to\_rod}, \texttt{plug\_in\_charger}, and \texttt{screw} often exhibit near-zero success while still achieving measurable intermediate progress. This suggests that current policies can frequently infer plausible action sequences and complete early subtasks, but struggle to reliably execute the final, precision-sensitive stages of long-horizon behaviors. A second bottleneck appears in tasks requiring tightly coupled perception, reasoning, and action, such as \texttt{adjust\_temperature}, \texttt{alphabet\_word}, and several memory-tracking tasks, where performance degrades substantially under OOD conditions. In contrast, much stronger results are observed on coarse object-centric manipulation and familiar semantic compositions, including \texttt{put\_glasses\_on\_woodshelf}, \texttt{sweep\_trash}, and several language reasoning tasks under ID settings. Taken together, these results indicate that current generalist policies are often capable of producing partially correct behavior, but remain substantially less reliable at converting such progress into successful task completion when precise contact control, persistent state tracking, or closed-loop adaptation is required.

\begin{table}
\centering
\caption{\textbf{X2Real evaluation results.} Each cell reports ID/OOD.}
\label{tab:x2real-results}
\setlength{\tabcolsep}{4pt}
\renewcommand{\arraystretch}{1.05}
\footnotesize
\resizebox{\linewidth}{!}{%
\begin{tabular}{lcccccccc}
\toprule
Task & \multicolumn{2}{c}{wall-oss-preview} & \multicolumn{2}{c}{$\pi_{0.5}$} & \multicolumn{2}{c}{DreamZero} & \multicolumn{2}{c}{Wall-OSS-0.5} \\
\cmidrule(lr){2-3}\cmidrule(lr){4-5}\cmidrule(lr){6-7}\cmidrule(lr){8-9}
& SR (\%) & Score & SR (\%) & Score & SR (\%) & Score & SR (\%) & Score \\
\midrule

\multicolumn{9}{l}{\textbf{Reasoning}} \\
\multicolumn{9}{l}{\textit{Generalization}} \\
\quad\texttt{atom\_objects\_beside\_block} & 92.50/75.00 & 94.50/84.75 & 50.00/7.32 & 70.25/53.30 & 5.00/2.50 & 35.80/47.00 & 15.00/7.50 & 50.00/44.00 \\
\quad\texttt{atom\_bottle\_on\_plate} & 70.00/55.00 & 84.50/78.50 & 22.50/15.00 & 58.50/52.00 & 25.00/5.00 & 47.80/67.50 & 0.00/0.00 & 41.25/32.50 \\
\quad\texttt{atom\_fruit\_in\_basket} & 97.50/87.50 & 97.50/96.50 & 65.00/67.50 & 67.00/86.50 & 65.00/17.50 & 56.00/80.50 & 17.50/20.00 & 21.50/34.25 \\
\quad\texttt{atom\_spoon} & 82.50/80.00 & 89.50/87.25 & 12.50/32.50 & 31.50/58.50 & 0.00/12.50 & 36.00/48.30 & 35.00/12.50 & 68.75/38.75 \\
\quad\texttt{atom\_drawer} & 100.00/55.00 & 100.00/86.00 & 40.00/5.00 & 62.50/27.50 & 0.00/17.50 & 15.20/54.70 & 2.50/0.00 & 15.50/11.00 \\
\quad\texttt{cup\_on\_plate} & 90.00/0.00 & 96.00/27.00 & 37.50/0.00 & 77.80/25.50 & 12.50/0.00 & 75.30/17.30 & 0.00/0.00 & 41.25/25.50 \\
\quad\texttt{open\_door} & 90.00/2.50 & 94.00/36.00 & 62.50/0.00 & 77.20/33.00 & 0.00/0.00 & 63.00/9.50 & 0.00/0.00 & 17.00/4.50 \\
\quad\textbf{average} & \textbf{88.93/50.71} & \textbf{93.71/70.86} & \textbf{41.43/18.19} & \textbf{63.54/48.04} & \textbf{15.36/7.86} & \textbf{47.01/46.40} & \textbf{10.00/5.71} & \textbf{36.46/27.21} \\

\multicolumn{9}{l}{\textit{Visual Reasoning}} \\
\quad\texttt{classify\_objects\_color} & 92.50/62.50 & 96.00/83.50 & 35.00/20.00 & 66.50/54.50 & 0.00/0.00 & 19.50/20.50 & 0.00/0.00 & 5.00/6.00 \\
\quad\texttt{classify\_object\_shape} & 42.50/30.00 & 56.00/55.50 & 30.00/5.00 & 60.00/36.50 & 5.00/0.00 & 41.20/21.70 & 0.00/0.00 & 6.00/0.70 \\
\quad\texttt{adjust\_temperature} & 12.50/2.50 & 58.50/14.50 & 0.00/2.50 & 18.00/14.50 & 0.00/0.00 & 2.00/0.00 & 5.00/0.00 & 11.00/4.00 \\
\quad\texttt{image\_puzzle} & 35.00/22.50 & 61.75/52.25 & 7.50/5.00 & 45.00/42.25 & 0.00/0.00 & 22.70/28.50 & 0.00/0.00 & 11.75/10.00 \\
\quad\texttt{adjust\_balance} & 56.25/18.33 & 69.25/35.92 & 31.25/6.67 & 44.88/14.08 & 0.00/0.00 & 32.25/12.77 & 0.00/0.00 & 7.25/1.25 \\
\quad\texttt{rotate\_book} & 80.00/37.50 & 80.00/41.50 & 60.00/15.00 & 70.00/31.00 & 2.50/0.00 & 26.00/18.00 & 0.00/0.00 & 4.00/0.00 \\
\quad\texttt{alphabet\_word} & 16.25/1.88 & 33.75/10.88 & 0.00/0.00 & 3.15/0.94 & 0.00/0.00 & 1.45/0.62 & 0.00/0.00 & 0.38/0.25 \\
\quad\textbf{average} & \textbf{47.86/25.03} & \textbf{65.04/42.01} & \textbf{23.39/7.74} & \textbf{43.93/27.68} & \textbf{1.07/0.00} & \textbf{20.73/14.58} & \textbf{0.71/0.00} & \textbf{6.48/3.17} \\

\multicolumn{9}{l}{\textit{Language Reasoning}} \\
\quad\texttt{fruit\_to\_basket} & 52.50/32.50 & 91.50/69.25 & 27.50/20.00 & 50.30/40.80 & 0.00/2.50 & 54.00/37.70 & 2.44/0.00 & 13.17/8.20 \\
\quad\texttt{and\_expression} & 87.50/52.50 & 95.00/92.50 & 75.00/37.50 & 77.50/62.50 & 2.50/0.00 & 45.00/52.00 & 15.00/15.00 & 19.00/21.00 \\
\quad\texttt{or\_expression} & 87.50/57.50 & 95.00/62.50 & 75.00/37.50 & 77.50/44.50 & 2.50/7.50 & 45.00/59.00 & 15.00/7.32 & 19.00/7.32 \\
\quad\texttt{negative\_expression} & 87.50/2.50 & 95.00/37.50 & 75.00/0.00 & 77.50/0.00 & 2.50/2.50 & 45.00/9.50 & 15.00/0.00 & 19.00/0.00 \\
\quad\texttt{confusion\_instruction} & 87.50/75.00 & 95.00/82.50 & 75.00/52.50 & 77.50/70.00 & 2.50/2.50 & 45.00/47.00 & 15.00/5.00 & 19.00/9.00 \\
\quad\texttt{specific\_position} & 92.50/77.50 & 92.50/77.50 & 0.00/2.50 & 0.00/2.50 & 10.00/15.00 & 38.00/34.00 & 12.50/17.50 & 24.50/19.50 \\
\quad\textbf{average} & \textbf{82.50/49.58} & \textbf{94.00/70.29} & \textbf{54.58/25.00} & \textbf{60.05/36.72} & \textbf{3.33/5.00} & \textbf{45.33/39.87} & \textbf{12.49/7.47} & \textbf{18.95/10.84} \\

\multicolumn{9}{l}{\textit{Memory Tracking}} \\
\quad\texttt{track\_object\_under\_cup} & 17.50/5.00 & 29.50/20.00 & 2.50/5.00 & 19.00/11.00 & 0.00/0.00 & 27.70/22.50 & 0.00/2.50 & 0.75/7.75 \\
\quad\texttt{buttons\_random\_order} & 38.75/11.25 & 54.87/32.88 & 31.25/18.75 & 46.75/37.25 & 0.00/0.00 & 23.65/16.60 & 8.75/2.50 & 26.25/24.00 \\
\quad\texttt{hit\_times} & 65.00/55.00 & 81.75/81.00 & 25.00/27.50 & 50.00/55.50 & 0.00/0.00 & 78.80/78.80 & 7.50/17.50 & 36.50/43.50 \\
\quad\texttt{find\_drawer} & 80.00/5.00 & 92.50/40.75 & 17.50/0.00 & 62.00/10.00 & 0.00/0.00 & 11.50/9.00 & 0.00/0.00 & 11.00/1.00 \\
\quad\texttt{put\_block\_back} & 60.00/25.00 & 75.75/49.00 & 12.50/17.50 & 53.00/48.50 & 10.00/5.00 & 50.50/49.00 & 2.50/0.00 & 10.00/2.50 \\
\quad\textbf{average} & \textbf{52.25/20.25} & \textbf{66.88/44.73} & \textbf{17.75/13.75} & \textbf{46.15/32.45} & \textbf{2.00/1.00} & \textbf{38.43/35.18} & \textbf{3.75/4.50} & \textbf{16.90/15.75} \\

\quad\textbf{total average} & \textbf{67.88/36.39} & \textbf{79.91/56.97} & \textbf{21.79/14.61} & \textbf{40.50/33.62} & \textbf{5.44/3.46} & \textbf{37.88/34.01} & \textbf{6.74/4.42} & \textbf{19.70/14.24} \\

\midrule

\multicolumn{9}{l}{\textbf{Manipulation}} \\
\multicolumn{9}{l}{\textit{Fine Manipulation}} \\
\quad\texttt{stack\_blocks} & 70.00/10.00 & 90.00/52.25 & 2.50/0.00 & 33.00/24.50 & 0.00/0.00 & 26.50/0.50 & 0.00/0.00 & 1.00/0.50 \\
\quad\texttt{triangle\_cup} & 20.00/5.00 & 64.50/61.25 & 2.50/0.00 & 52.25/49.75 & 15.00/10.00 & 41.50/36.00 & 0.00/0.00 & 5.75/3.50 \\
\quad\texttt{ring\_to\_rod} & 17.50/20.00 & 80.50/78.25 & 2.50/0.00 & 39.00/43.00 & 2.50/0.00 & 56.70/37.50 & 2.50/0.00 & 6.50/0.00 \\
\quad\texttt{plug\_in\_charger} & 0.00/0.00 & 13.50/20.25 & 0.00/0.00 & 9.00/6.80 & 0.00/0.00 & 0.70/0.70 & 0.00/0.00 & 0.75/0.75 \\
\quad\texttt{insert\_flower} & 62.50/45.00 & 93.00/92.00 & 25.00/12.50 & 75.75/69.50 & 0.00/2.50 & 31.00/38.00 & 0.00/2.50 & 16.50/20.25 \\
\quad\textbf{average} & \textbf{34.00/16.00} & \textbf{68.30/60.80} & \textbf{6.50/2.50} & \textbf{41.80/38.71} & \textbf{3.50/2.50} & \textbf{31.28/22.54} & \textbf{0.50/0.50} & \textbf{6.10/5.00} \\

\multicolumn{9}{l}{\textit{Bimanual Collaboration}} \\
\quad\texttt{hand\_over\_coin} & 40.00/25.00 & 64.75/58.00 & 0.00/0.00 & 31.50/31.50 & 5.00/5.00 & 46.50/45.80 & 0.00/0.00 & 24.00/21.70 \\
\quad\texttt{hand\_over} & 65.00/47.50 & 75.50/67.75 & 7.14/10.00 & 34.50/36.20 & 12.50/10.00 & 45.00/40.50 & 7.50/5.00 & 31.50/30.50 \\
\quad\texttt{put\_glasses\_on\_woodshelf} & 95.00/50.00 & 99.50/89.00 & 72.50/47.50 & 94.50/84.50 & 0.00/0.00 & 57.00/57.20 & 10.00/5.00 & 53.30/33.50 \\
\quad\texttt{object\_to\_cup} & 52.50/17.50 & 74.00/43.75 & 2.50/0.00 & 29.80/19.50 & 0.00/0.00 & 25.00/18.00 & 0.00/0.00 & 15.25/11.75 \\
\quad\textbf{average} & \textbf{63.12/35.00} & \textbf{78.44/64.62} & \textbf{20.54/14.38} & \textbf{47.58/42.93} & \textbf{4.38/3.75} & \textbf{43.38/40.38} & \textbf{4.38/2.50} & \textbf{31.01/24.36} \\

\multicolumn{9}{l}{\textit{Tool Use}} \\
\quad\texttt{Whack\_a\_Mole} & 95.00/67.50 & 96.00/73.50 & 50.00/15.00 & 70.50/49.00 & 25.00/2.50 & 50.80/20.50 & 0.00/0.00 & 13.00/7.50 \\
\quad\texttt{sweep\_trash} & 92.50/67.50 & 97.50/87.25 & 55.00/22.50 & 74.00/45.50 & 2.50/5.00 & 45.30/51.70 & 2.50/0.00 & 30.25/22.50 \\
\quad\texttt{screw} & 0.00/0.00 & 18.50/18.50 & 0.00/0.00 & 19.00/18.00 & 0.00/0.00 & 20.00/20.00 & 0.00/0.00 & 19.50/20.00 \\
\quad\textbf{average} & \textbf{62.50/45.00} & \textbf{70.67/59.75} & \textbf{35.00/12.50} & \textbf{54.50/37.50} & \textbf{9.17/2.50} & \textbf{38.70/30.73} & \textbf{0.83/0.00} & \textbf{20.92/16.67} \\

\multicolumn{9}{l}{\textit{Dynamic Objects}} \\
\quad\texttt{pick\_and\_place\_on\_turntable} & 0.00/32.50 & 1.00/40.00 & 9.76/35.00 & 12.00/38.00 & 0.00/5.00 & 8.80/27.30 & 0.00/4.88 & 6.00/9.76 \\
\quad\textbf{average} & \textbf{0.00/32.50} & \textbf{1.00/40.00} & \textbf{9.76/35.00} & \textbf{12.00/38.00} & \textbf{0.00/5.00} & \textbf{8.80/27.30} & \textbf{0.00/4.88} & \textbf{6.00/9.76} \\

\quad\textbf{total average} & \textbf{39.91/32.12} & \textbf{54.60/56.29} & \textbf{17.95/16.09} & \textbf{38.97/39.28} & \textbf{4.26/3.44} & \textbf{30.54/30.24} & \textbf{1.43/1.97} & \textbf{16.01/13.95} \\

\midrule

\multicolumn{9}{l}{\textbf{All}} \\
\quad\textbf{total average} & \textbf{53.90/34.26} & \textbf{67.25/56.63} & \textbf{26.12/16.13} & \textbf{46.19/37.75} & \textbf{4.85/3.45} & \textbf{34.21/32.12} & \textbf{4.08/3.20} & \textbf{17.85/14.10} \\

\bottomrule
\end{tabular}%
}
\end{table}

\subsubsection{Coding Agent as Policy}
\label{sec:exp-coding-agent}

We additionally evaluate GPT-6-Astra~\cite{openai_gpt6_astra} as an interactive robot policy to examine whether a general-purpose coding agent can translate visual observations and explicit reasoning into successful manipulation. Astra is queried with the \texttt{xHigh} reasoning configuration and iteratively produces structured end-effector pose and gripper commands, which are executed by a robot controller. We evaluate it on all 44 native OOD environments, with 10 scored episodes per environment, and aggregate the results into the same 38 tasks using the hierarchical macro-averaging protocol in Eq.~\ref{eq:sr-score}. We compare against the OOD results of wall-oss-preview reported in Tab.~\ref{tab:x2real-results}. The two methods share the benchmark metrics and task definitions, but use different policy interfaces and inference procedures.

At each decision step, Astra receives the current and previous RGB observations from the head and two wrist cameras, together with proprioceptive state, execution feedback, and remaining task budgets. Proprioceptive inputs include bimanual end-effector poses, joint positions, and gripper states. Textual observations, decisions, actions, and feedback are retained within an episode and reset between episodes. Astra does not receive depth images, camera calibration parameters, ground-truth object poses, rewards, or online progress scores. Commands are executed at 30\,Hz with translational and rotational speed limits of 0.10\,m/s and 0.50\,rad/s, respectively. Each action has an 8\,s timeout, while the episode horizon follows the native task configuration. The model-call budget is proportional to the simulated episode horizon, allowing at most $\lceil 100T/60 \rceil$ calls for a task with a time limit of $T$ seconds.

\begin{figure}[t]
    \centering
    \includegraphics[width=\linewidth]
        {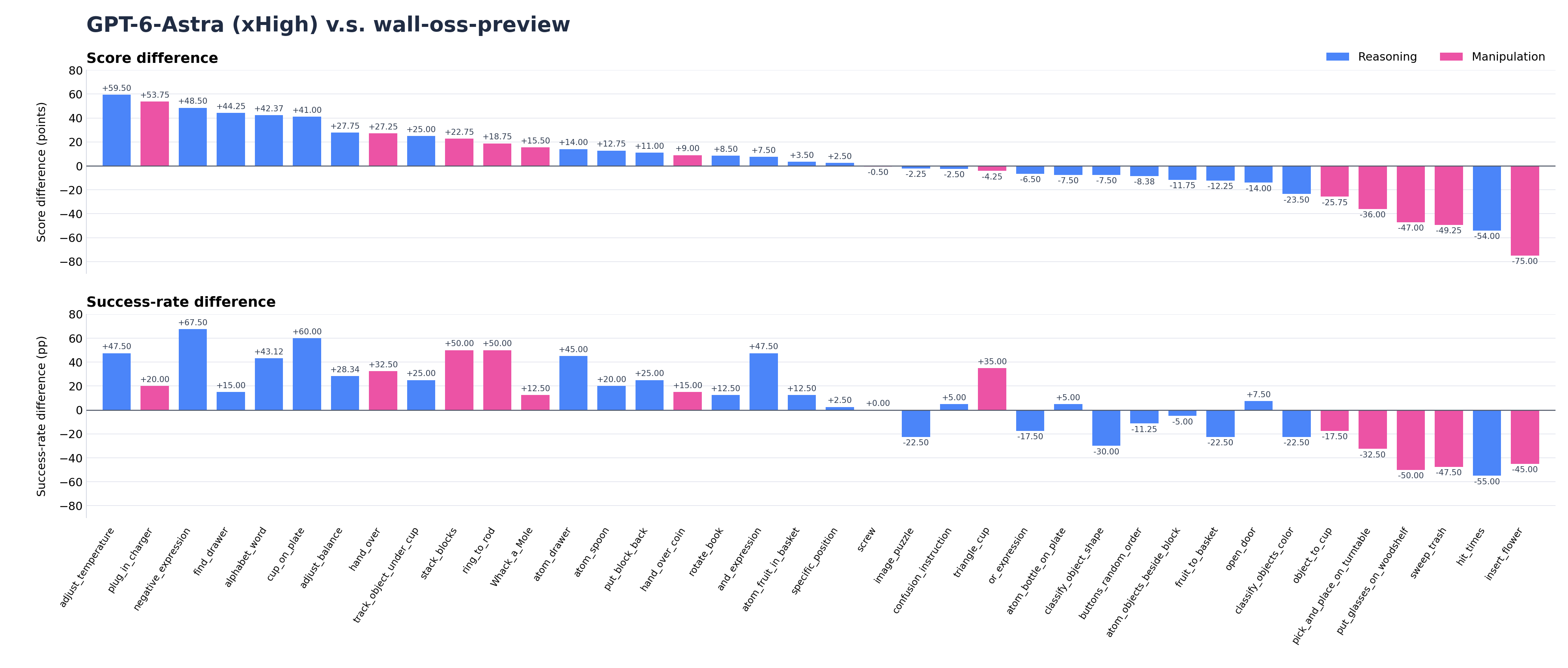}
    \caption{\textbf{Coding Agent as Policy.}
    Task-level OOD progress-score differences (top) and success-rate
    differences (bottom), computed as GPT-6-Astra minus wall-oss-preview.
    Tasks are ordered by decreasing score difference, with the same
    order in both panels. Blue denotes Reasoning tasks and pink denotes
    Manipulation tasks. Score differences use the 100-point scale;
    success-rate differences are in percentage points.}
    \label{fig:astra_bus2606_ood_compact}
\end{figure}

Overall, Astra achieves an OOD success rate of $36.15\%$, exceeding
wall-oss-preview by 1.89 percentage points, but obtains a lower progress
score of $53.74$ versus $56.63$. This comparable aggregate performance
masks distinct capability profiles. On Reasoning tasks, Astra achieves
$46.96\%$ success and a score of $64.52$, improving over wall-oss-preview
by 10.57 percentage points and 7.55 score points. On Manipulation
tasks, it achieves $25.33\%$ success and a score of $42.96$, falling
behind by 6.79 percentage points and 13.33 score points. Thus, neither
method dominates both capability dimensions or both evaluation metrics.

Figure~\ref{fig:astra_bus2606_ood_compact} illustrates the task dependence
of these differences. Astra's largest score gains occur on
\texttt{adjust\_temperature} ($+59.50$),
\texttt{plug\_in\_charger} ($+53.75$), and
\texttt{negative\_expression} ($+48.50$).
Its Fine Manipulation success rate also exceeds wall-oss-preview
($38.00\%$ versus $16.00\%$), although the corresponding score advantage
is smaller ($64.00$ versus $60.80$).
In contrast, substantial score deficits occur on
\texttt{insert\_flower} ($-75.00$),
\texttt{hit\_times} ($-54.00$), and
\texttt{sweep\_trash} ($-49.25$).
These results indicate strengths in selected reasoning-oriented tasks
and some precision-sensitive operations, alongside weaknesses in
dynamic interaction, tool use, and certain bimanual behaviors.
The pattern is not universal: some manipulation tasks favor Astra,
while some reasoning tasks remain challenging. Moreover, high progress
does not imply reliable completion; for example, Astra scores $85.00$
on \texttt{find\_drawer} but succeeds in only $20\%$ of episodes.

A practical limitation is inference latency. Recorded inference takes
more than 30\,s per decision chunk, during which the simulation is
paused. Consequently, the reported results characterize task execution
with substantial deliberation time and a suspended environment,
rather than real-time closed-loop control. Simulation-time budgets
therefore do not capture the full wall-clock cost of this approach.
Together with the limited sample size of 10 scored episodes per native
environment, this setting motivates treating the observed differences
as descriptive rather than statistically established advantages.
Deployment-oriented evaluation should additionally measure end-to-end
latency and execution reliability when the environment continues to
evolve during inference.


\subsubsection{Mobile Operation}
\label{sec:exp-mobile-operation}

We evaluate our internal model wall-oss-preview and
$\pi_{0.5}$~\cite{black2025pi05} on three Mobile Operation tasks:
\textit{Move Bottle to Bin}, \textit{Move Fruit to Basket}, and
\textit{Bottle on Shelf}. Each task comprises five sequential DAG stages:
source navigation, object picking, object transport, placement, and arm
retraction. Using Mana, we automatically synthesize approximately 1,000
successful demonstrations per task and jointly post-train each model
across the three tasks using the same datasets. We report checkpoints
at 30k training steps for wall-oss-preview and 45k for $\pi_{0.5}$.

Both models are evaluated under the same protocol, with 40 episodes per
task in each of the in-distribution (ID) and out-of-distribution (OOD)
settings. ID uses the green-booth environment, whereas OOD varies scene
backgrounds and table appearance while retaining the same object pools,
layout distributions, and task definitions. The episode time limits are
90 seconds for ID and 120 seconds for OOD.

\begin{table}[t]
    \centering
    \caption{\textbf{Mobile Operation Evaluation.}
    Each cell reports ID/OOD.}
    \label{tab:mobile-operation-results}
    \setlength{\tabcolsep}{4pt}
    \renewcommand{\arraystretch}{1.05}
    \footnotesize
    \begin{tabular}{lcccc}
        \toprule
        Task & \multicolumn{2}{c}{wall-oss-preview}
             & \multicolumn{2}{c}{$\pi_{0.5}$} \\
        \cmidrule(lr){2-3}\cmidrule(lr){4-5}
         & SR (\%) & Score & SR (\%) & Score \\
        \midrule
        \texttt{move\_bottle\_to\_bin}
            & 35/7.5 & 67/45 & 37.5/0 & 70.5/25 \\
        \texttt{move\_fruit\_to\_basket}
            & 85/2.5 & 93.5/45 & 15/0 & 38/30.5 \\
        \texttt{put\_bottle\_on\_woodshelf}
            & 55/5 & 72/42.5 & 35/2.5 & 50/19 \\
        \midrule
        \textbf{average}
            & \textbf{58.33/5.00} & \textbf{77.50/44.17}
            & \textbf{29.17/0.83} & \textbf{52.83/24.83} \\
        \bottomrule
    \end{tabular}
\end{table}

As shown in Table~\ref{tab:mobile-operation-results}, wall-oss-preview
achieves higher task-mean SR and Score on both splits, with the largest
ID success-rate difference on \textit{Move Fruit to Basket}
($85.0\%$ versus $15.0\%$). Nevertheless, both models have low OOD
success: wall-oss-preview completes 6 of 120 episodes and $\pi_{0.5}$
completes one. Their OOD Scores of $44.17$ and $24.83$, respectively,
indicate partial progress despite limited end-to-end success.

\begin{figure}[t]
    \centering
    \includegraphics[width=\linewidth]
        {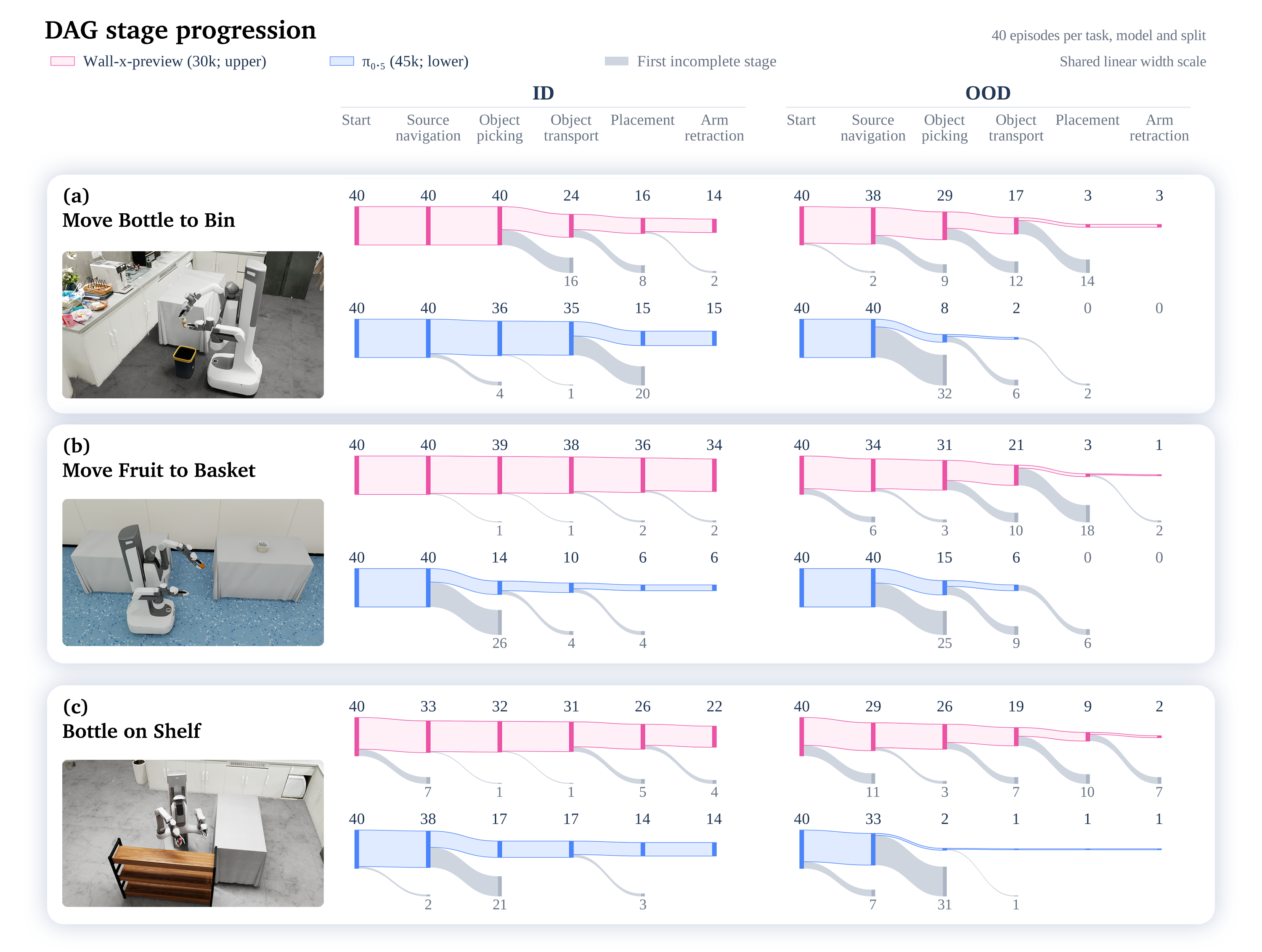}
    \caption{\textbf{DAG-based Mobile Operation Analysis.}
    ID and OOD stage progression for the three mobile tasks.
    Pink (upper) and blue (lower) ribbons represent wall-oss-preview and
    $\pi_{0.5}$, respectively. Each flow starts with 40 episodes, and all
    ribbon widths use a shared linear count scale. Colored flows indicate
    completion of successive DAG stages; gray branches count episodes
    whose first incomplete stage at termination is the corresponding
    column. Numbers denote episode counts; zero-count branches are omitted.}
    \label{fig:mobile-operation-dag-flow}
\end{figure}

Figure~\ref{fig:mobile-operation-dag-flow} reveals distinct execution
profiles behind similar task-level outcomes. On \textit{Move Bottle to Bin}
under ID, wall-oss-preview and $\pi_{0.5}$ achieve similar success rates
($35.0\%$ and $37.5\%$), but their most frequent first incomplete stages
differ: transport for wall-oss-preview (16 episodes) and placement for
$\pi_{0.5}$ (20 episodes). Under OOD, picking becomes the most frequent
first incomplete stage for $\pi_{0.5}$ (32 episodes), whereas placement
is the most frequent for wall-oss-preview (14 episodes).

For \textit{Move Fruit to Basket}, $\pi_{0.5}$ completes picking in only
14 ID and 15 OOD episodes, revealing a limitation already present on ID.
wall-oss-preview completes picking in 39 ID and 31 OOD episodes, but its
placement completion decreases from 36 to three. On
\textit{Bottle on Shelf}, nine wall-oss-preview OOD episodes complete
placement, but only two satisfy the final arm-retraction node, which
also checks that the object remains validly placed. These DAG records
distinguish early picking limitations from unfinished placement and
finalization, without attributing them to specific physical failure causes.

\input{sec/automatic_synthesis_vs_teleoperation/automatic_synthesis_vs_teleoperation}

%% file: sec/automatic_synthesis_vs_teleoperation/automatic_synthesis_vs_teleoperation.tex
\subsection{Automatic Synthesis vs. Teleoperation}
\label{sec:automatic-synthesis-vs-teleoperation}

We compare Task-DAG-guided automatic synthesis with master--slave teleoperation on the dual-arm \texttt{stack\_blocks (hard)} challenge and further study how full-task and stagewise instructions affect the two data sources. The task requires the robot to stack 12 blocks consecutively and return both arms to their home configurations; as the tower grows, increasingly precise end-effector alignment and placement are required. Crossing the two data sources with the two instruction settings yields four experimental groups. Automatic synthesis and teleoperation each provide 1,398 complete task trajectories and use the same training budget. Within each data source, the two models use the same underlying trajectories; the stagewise variant only reorganizes episodes according to the stage boundaries recorded by the Task DAG and assigns the corresponding subtask instruction to each segment. All four models are initialized from OpenPI $\pi_{0.5}$ and trained for two epochs using identical visual and proprioceptive inputs, training hyperparameters, and action representations.

For each group, we report one representative checkpoint evaluated over 20 closed-loop episodes. Each episode runs for at most 300 seconds. An episode is considered successful only if all 12 layers are completed and both arms return to their home configurations. For incomplete episodes, task progress is measured by the number of stably completed layers latched by the Task DAG.

\begin{table}[t]
  \centering
  \small
  \setlength{\tabcolsep}{6pt}
  \caption{Closed-loop performance of automatic synthesis and teleoperation under full-task and stagewise instruction settings. Each row reports one representative checkpoint evaluated over 20 episodes.}
  \label{tab:auto-vs-teleop}
  \begin{tabular}{llccc}
    \toprule
    Data source & Instruction & Checkpoint & Mean layers & Max layer \\
    \midrule
    Automatic synthesis & Full-task & 55k & 3.75 & 6 \\
    Teleoperation       & Full-task & 50k & 1.90 & 4 \\
    Automatic synthesis & Stagewise & 40k & 4.40 & 10 \\
    Teleoperation       & Stagewise & 50k & 2.05 & 4 \\
    \bottomrule
  \end{tabular}
\end{table}

\begin{figure}[t]
  \centering
  \includegraphics[width=\linewidth]{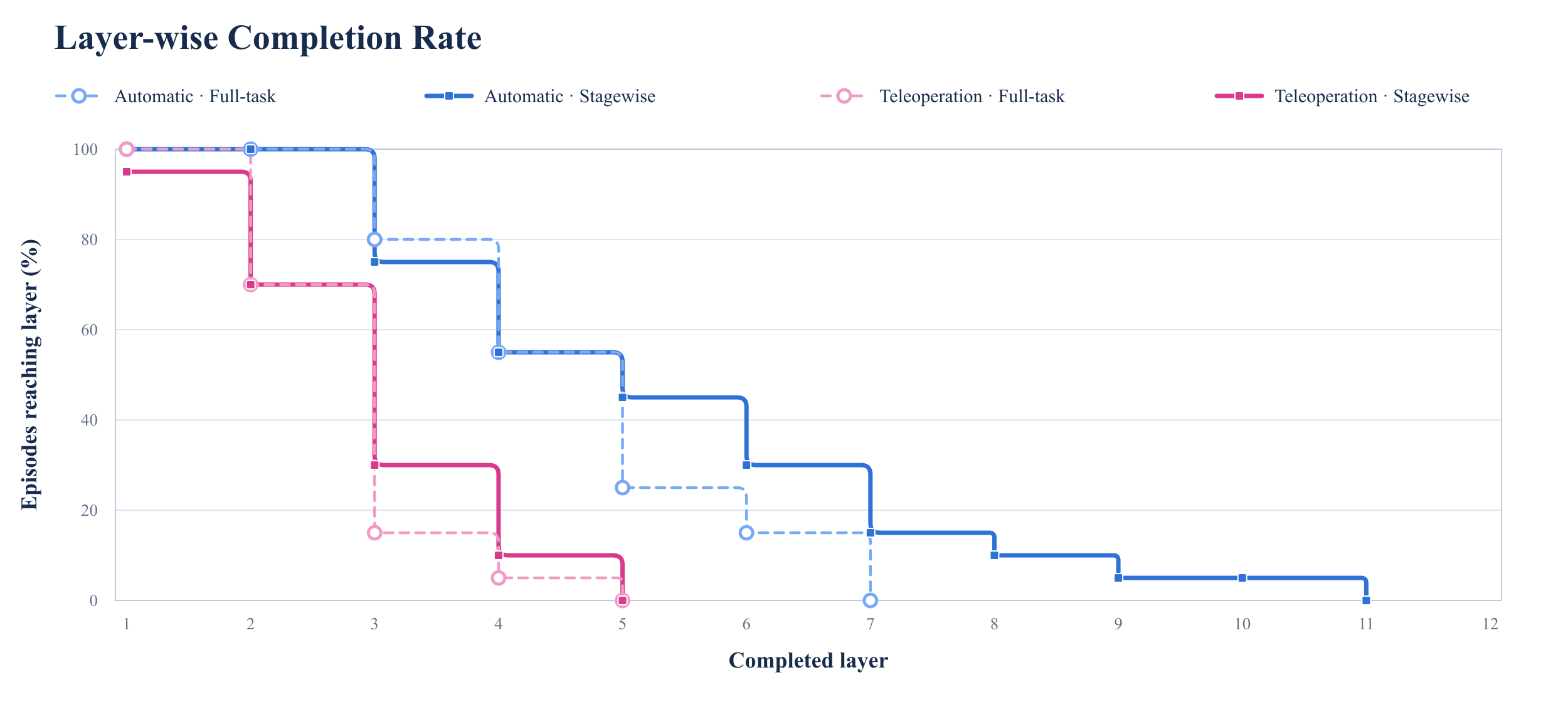}
  \caption{Layer-wise completion rates across 20 closed-loop episodes. Blue and pink denote automatic synthesis and teleoperation, respectively. Solid lines denote stagewise instructions, whereas dashed lines denote a single full-task instruction. Each curve extends to the first layer with a zero completion rate.}
  \label{fig:auto-vs-teleop-layer-completion}
\end{figure}

Under the full-task instruction setting, the automatically synthesized data yield an average of 3.75 completed layers and a maximum of 6, whereas the teleoperated data yield an average of 1.90 layers and a maximum of 4. Automatic synthesis therefore improves the mean progress by 1.85 layers and maintains a higher layer-wise completion rate beyond the third layer. On this precision-intensive, multi-stage stacking task, the result indicates that Task-DAG-guided automatic synthesis provides more effective policy-training data than master--slave teleoperation. The two demonstration sources differ substantially in their temporal structure. As the tower grows, teleoperated trajectories devote progressively more time to end-effector alignment, visual inspection, and local position correction. When fixed-length action chunks are sampled from the temporal sequence, these long, low-motion segments occupy a larger fraction of the training distribution, reducing the relative frequency of grasping and placement actions that advance the task. In contrast, automatic synthesis generates more compact trajectories at the level of Task DAG nodes and can control coverage across stages, thereby reducing the training-distribution bias caused by differences in stage duration.

The DAG-aligned stagewise setting further improves task progress. For automatic synthesis, the mean number of completed layers increases from 3.75 to 4.40 and the maximum from 6 to 10. The completion rate at layer 5 rises from 25\% to 45\%, and that at layer 6 from 15\% to 30\%; the policy progresses from never reaching layer 7 to completing as many as 10 layers, with the improvement concentrated in later stacking stages. Under a single full-task instruction, all 12 stacking stages share the same language condition even though their local objectives and action distributions differ, requiring the policy to infer the current operation primarily from visual observations. The stagewise setting instead uses Task DAG completion boundaries to organize training segments and assigns the corresponding local objective to each segment, thereby reducing the action ambiguity caused by associating one instruction with multiple operation stages. Because the automatic trajectories are themselves generated at the node level and verified by node success predicates, their stage boundaries, instruction semantics, and physical outcomes are tightly aligned. For teleoperation, the mean number of completed layers increases from 1.90 to 2.05 while the maximum remains 4, indicating that explicit stage objectives benefit both data sources, with a more pronounced gain for automatic synthesis.

%% file: sec/6_related.tex
\section{Related Work}\label{sec:related}

\subsection{Evaluating Generalist Robot Manipulation Policies}

Generalist robot policies are often trained to acquire diverse manipulation skills and generalize across multiple dimensions, such as objects, environments, language instructions, task configurations, and robot embodiments.
Representative approaches to building such policies include vision-language-action models (VLAs)~\citep{zitkovich2023rt2,kim2025openvla,black2025pi0,black2025pi05,abdolmaleki2025geminirobotics15,nvidia2026grootn17,yu2026walloss05}, as well as world action models (WAMs) and related video--action models that jointly capture visual dynamics and robot actions~\citep{li2025uva,zhu2025uwm,pai2025mimicvideo,ye2026dreamzero,li2026wallwm,zhang2026lingbotva2}.
Evaluating these policies across diverse manipulation skills and generalization dimensions therefore poses new challenges.

For single-task policies, success rates under a fixed evaluation protocol often serve as the primary measure of performance. 
For generalist policies, however, evaluation must also assess both the range of manipulation skills and their generalization beyond the objects, environments, language instructions, task configurations, and robot embodiments encountered during training. 
Evaluation therefore cannot rely solely on aggregate success rates across a small set of fixed tasks. 
This requires broad yet controlled coverage across relevant skills and generalization dimensions~\citep{gao2026taxonomy,black2025pi05,pumacay2024colosseum}.

Real-robot evaluation provides direct evidence of policy performance in physical systems, naturally incorporating sensor noise, contact dynamics, control delays, and hardware-specific effects.
However, statistically reliable evaluation requires repeated trials across different tasks and initial conditions. 
Each trial requires access to a robot and often involves manual scene setup, environment reset, safety monitoring, and success assessment, making large-scale evaluation costly and time-consuming. 
Differences in robot hardware, evaluation environments, and evaluation protocols across institutions further hinder reproducibility. 
Recent work has begun to address these challenges through automated environment reset and success detection~\citep{zhou2025autoeval}, distributed evaluation across institutions~\citep{atreya2025roboarena}, and more rigorous statistical evaluation protocols~\citep{carefulLBM2025}. 
Nevertheless, the scale of real-robot evaluation remains limited by the number of available robots and evaluation environments. 
This limited scale poses a particular challenge during generalist policy evaluation, where many training configurations, checkpoints, and hyperparameter settings must be compared.
Simulation-based evaluation offers a scalable complement to the real-robot evaluation and can partially address these limitations.

\subsection{Simulation-based Evaluation of Robot Manipulation Policies}

Simulation benchmarks provide standardized, reproducible, and scalable environments for evaluating robot policies. 
Meta-World targets multi-task and meta-reinforcement learning~\citep{yu2020metaworld}, RLBench supports vision-based manipulation under several learning settings~\citep{james2020rlbench}, and robosuite provides modular and standardized environments for robot learning~\citep{zhu2020robosuite}.
ManiSkill and ManiSkill2 progressively expand the coverage of objects, tasks, robot embodiments, and observation modalities~\citep{mu2021maniskill,gu2023maniskill2}, while ManiSkill3 further introduces GPU-parallelized physics simulation and rendering for more efficient data collection and policy evaluation~\citep{tao2025maniskill3}.
CALVIN evaluates policies on long-horizon manipulation tasks specified by natural-language instructions~\citep{mees2022calvin}, whereas LIBERO evaluates knowledge transfer across sequentially learned manipulation tasks~\citep{liu2023libero}.
More recent benchmarks, including BEHAVIOR-1K, RoboCasa, VLABench, RoboTwin 2.0, RoboVerse, MolmoSpaces, RoboCasa365, and RoboDojo, further expand evaluation to richer household environments, more diverse language-conditioned and long-horizon tasks, bimanual and mobile manipulation, and a wider range of robot embodiments~\citep{li2023behavior1k,nasiriany2024robocasa,zhang2025vlabench,chen2025robotwin2,geng2025roboverse,kim2026molmospaces,nasiriany2026robocasa365,chen2026robodojo}.

Most existing simulation benchmarks rely on physics-based simulation. Action-conditioned world models have also been studied as learned environments for robot policy evaluation by predicting future visual observations conditioned on robot actions~\citep{li2025worldeval,quevedo2025worldgym,tseng2025scalable,geminirobotics2025veo,wang2026interactiveworldsimulator,li2026dworldeval,tseng2026sc3eval}. Learned and physics-based simulators provide complementary forms of environment modeling. Because our framework requires explicit state access, state-based success detection, and systematic control over assets, tasks, and evaluation conditions, we adopt physics-based simulation.

When simulation is used as a proxy for real-robot evaluation, its validity depends on whether performance measured in simulation reflects real-robot performance. Simulators allow explicit configuration of visual conditions, dynamics, controllers, and hardware parameters, but these configurations must be calibrated against the target physical system~\citep{mehta2021calibrating}. SIMPLER studies sim--real correspondence through paired simulation and real-robot evaluations~\citep{li2025simpler}. PolaRiS and REALM extend this line of research through reconstructed real-world environments and controlled evaluation conditions, respectively~\citep{jain2026polaris,sedlacek2026realm}. VISER further studies how visual realism affects the agreement between simulation and real-world evaluation~\citep{zhu2026viser}, while related work extends sim--real policy evaluation to deformable-object interactions~\citep{zhang2026real2simsoftbody, li2026lehome}. SureSim instead combines a large number of simulation trials with a small number of paired real-robot trials to estimate real-robot performance and its confidence interval~\citep{badithela2026suresim}.

Beyond sim--real correspondence, evaluation diversity depends not only on the number of tasks, but also on broad yet controlled coverage of manipulation skills and generalization across objects, environments, language instructions, task configurations, and robot embodiments~\citep{gao2026taxonomy}. Colosseum, Colosseum V2, REALM, and RoboLab use controlled variations to analyze policy robustness and generalization~\citep{pumacay2024colosseum,morgan2026colosseumv2,sedlacek2026realm,yang2026robolab}. EBench provides capability- and generalization-level diagnosis for mobile manipulation~\citep{gao2026ebench}, while RoboDojo evaluates generalist policies across generalization, memory, precision, long-horizon execution, and open-vocabulary instruction following in both simulation and real-world settings~\citep{chen2026robodojo}.

Another concern is benchmark integrity, including the separation between training and evaluation and the complete specification of evaluation protocols. RoboLab highlights domain overlap between training and evaluation environments~\citep{yang2026robolab}. Recent benchmark audits further identify shortcut solvability, insufficient statistical evidence, gradual overfitting to established benchmarks, and dependence on training-data sources as factors that can limit the interpretation of benchmark results~\citep{jiang2026benchmarking}. Reproduction experiments conducted with vla-eval also show that undocumented termination conditions, data normalization, and other implementation details can affect evaluation results~\citep{choi2026vlaeval}. Together, these works motivate simulation benchmarks that provide evidence of sim--real correspondence, broad yet controlled coverage of relevant skills and generalization dimensions, and clearly specified training--evaluation splits and evaluation protocols.

\subsection{Simulation Data Engines for Robot Learning}

Simulation can serve not only as an environment for predefined training and evaluation tasks, but also as a system for continually producing robot learning data. In this work, we use the term \emph{simulation data engine} to denote an integrated pipeline for constructing and validating simulation-ready assets, generating executable scenes and tasks, collecting and validating demonstrations, organizing the resulting data, and maintaining a clear separation between training and evaluation distributions. A simulation benchmark primarily specifies what is evaluated and under which conditions, whereas a simulation data engine provides the underlying processes for producing and managing the assets, tasks, and data used for training and evaluation.

Simulation-ready assets form the foundation of a data engine and must support the physical interactions involved in its target tasks. SAPIEN, PartNet-Mobility, Objaverse, and GAPartNet provide important foundations for 3D objects, articulated structures, and manipulation-relevant part semantics~\citep{xiang2020sapien,deitke2023objaverse,geng2023gapartnet}. Building on these foundations, assets intended for robot simulation may require additional task-dependent information, such as metric scale, canonical orientation, part and task semantics, physically based rendering (PBR) materials, collision geometry, mass and inertia, contact properties, and executable kinematic structures. Recent generative models extend high-quality 3D generation toward simulation-ready and physically grounded asset creation~\citep{zhao2025hunyuan3d20,feng2025seed3d,cao2025physx3d}.
Articulate-Anything automates articulated-object modeling, whereas URDF-Anything+ generates executable articulated models directly from visual observations~\citep{le2025articulateanything,wu2026urdfanythingplus}. ManiTwin provides a scalable pipeline for generating simulation-ready assets enriched with physical properties, functional annotations, and language descriptions~\citep{wang2026manitwin}. AnnotateAnything generates manipulation-relevant annotations grounded in asset geometry and physical constraints~\citep{lu2026annotateanything}. VISER further develops material-aware asset preparation for visually realistic policy evaluation~\citep{zhu2026viser}. For reliable use at scale, automatically constructed assets generally require geometric, physical, and interaction-level validation. More generally, simulation readiness is task-relative: it depends on whether an asset provides the geometric, physical, semantic, and, where applicable, kinematic information required by its intended role, such as support surfaces, containment volumes, articulation parameters, and contact regions.

Given suitable assets, a data engine can expand its coverage by generating scenes, tasks, and demonstrations across diverse conditions and behaviors. GenSim uses large language models to generate simulation tasks, environments, and expert demonstrations~\citep{wang2024gensim}. RoboGen organizes task proposal, scene generation, supervision generation, and skill learning into a generative pipeline~\citep{wang2024robogen}. RoboTwin 2.0 combines multimodal language models with simulation-in-the-loop refinement to generate executable task-level programs~\citep{chen2025robotwin2}. Demonstrations can be collected through teleoperation, scripted experts, motion planning, trajectory optimization, and reinforcement learning. MimicGen and its subsequent extensions generate additional demonstrations from limited source data by adapting object-centric motion segments and extending this paradigm to skill composition, bimanual dexterous manipulation, dynamic tasks, deformable objects, and humanoid loco-manipulation~\citep{mandlekar2023mimicgen,garrett2025skillmimicgen,jiang2025dexmimicgen,pomponi2026dynamimicgen,moghani2026softmimicgen,lin2026humanoidmimicgen}. Such methods can reduce the need to collect demonstrations manually for every task condition, while the resulting coverage remains shaped by factors such as the available assets, task specifications, source demonstrations, and validation procedures.

A data engine operating at scale requires coordinated support for asset and scene construction, task and demonstration generation, simulation, and evaluation, together with unified interfaces and mechanisms for validating, organizing, and versioning the resulting data. Recent systems such as RoboTwin 2.0, SimFoundry, and Genie Sim 3.0 integrate multiple components spanning asset and scene construction, task generation, data generation, and policy evaluation~\citep{chen2025robotwin2,ranawaka2026simfoundry,yin2026geniesim30}. RoboVerse provides simulator-agnostic interfaces across multiple physics and rendering backends~\citep{geng2025roboverse}. IsaacIPC couples GPU-accelerated penetration-free contact simulation with Isaac Sim/Lab to support contact-rich rigid--deformable interactions~\citep{liang2026isaacipc}. MagicSim supports particle-based fluids, granular materials, and other physical processes, and integrates their simulation with world construction, robot execution, task evaluation, and data generation within a deterministic batched runtime~\citep{magicsim2026}. Complementary learned approaches, including DreamGen, Qwen-RobotWorld, and GigaWorld-0, use video or hybrid world models to generate synthetic visual trajectories and interaction data for policy learning~\citep{jang2025dreamgen,zhang2026qwenrobotworld,gigaworldteam2025gigaworld0}. RLDS, robomimic, LeRobot, and Robo-DM support standardized sequential-decision datasets, offline robot learning, end-to-end data collection and learning workflows, and large-scale robot data management, respectively~\citep{ramos2021rlds,mandlekar2022robomimic,cadene2026lerobot,chen2025robodm}. Integrating these components, including asset production, scene and task generation, demonstration collection, simulation infrastructure, quality control, version management, and training--evaluation separation, can support the continued expansion of training and evaluation distributions as policy capabilities evolve. Evaluation results can in turn identify underrepresented conditions and inform subsequent task and data generation.

%% file: sec/7_discussion.tex
\section{Discussion and Limitations}\label{sec:discussion}

We present X2Real simulation benchmark,  which addresses three core limitations inherent to contemporary robot manipulation simulation benchmarks, namely the simulation-to-reality gap, insufficient task diversity, and evaluative bias stemming from benchmark exploitation. To remedy these deficiencies, X2Real incorporates hardware-aligned physical calibration, a hierarchically structured multi-dimensional task set, orthogonal domain randomization, and strictly decoupled training and evaluation pipelines, yielding improved cross-domain correlation and robust quantification of generalist manipulation performance. Furthermore, the proposed Mana simulation ecosystem supports iterative benchmark update and closed-loop evaluation, alleviating the performance saturation inherent to static benchmark designs. The following discussion analyzes the key findings and inherent limitations of our framework, and highlights potential avenues for future embodied intelligence benchmarking and generalist policy development.

Despite its improved faithfulness, diversity, and fairness over existing benchmarks, X2Real still entails several limitations that point to promising future directions. First, the current benchmark tasks are restricted to rigid and articulated rigid body manipulation, which cannot fully replicate real-world scenarios involving deformable objects, fluids, and soft materials. Extending the physical simulation to support diverse material dynamics is essential for evaluating generalist policies with universal manipulation capabilities. Second, the construction of existing task suites relies on a combination of manual design and agent-assisted refinement, which limits the scalability of task quantity and scenario richness. Fully automated agent-driven task generation will further expand task diversity and continuously enrich the benchmark distribution. Third, our simulation framework currently adopts standard gripper embodiments without tactile sensing modules. Integrating dexterous hands, high-precision tactile feedback, and diverse robot hardware setups can better align with real robotic manipulation research and enhance evaluation generality. Fourth, the evaluation pipeline remains relatively static in paradigm. Future work can introduce intelligent agent-based result analysis to enable automated experimental iteration and comparative study, as well as adversarial and competitive evaluation paradigms similar to RoboArena, to achieve more dynamic, in-depth, and robust policy assessment.

%% file: sec/contributors.tex
\section{Contributors}\label{sec:contributors}

X2Real is a collaborative effort of the X Square robot team and outside collaborators. $^*$ denotes core contributors, $^\dagger$ denotes the project leader, $^\ddagger$ denotes the corresponding author.

Lian Ruan$^*\dagger$, Jade Yang$^*$, Sherphylan Gao$^*$, Felix Gao$^*$, Kyson Liang$^*$, Galen Liu$^*$, Ligo Wu$^*$, Lane Jin$^*$, Guu Gu$^*$, Bevan Xie$^*$, Cloud Yan$^*$, Zongzi Yuan$^*$, Kino Luo, Emma Chen, Shuwen Chen, Yang Ping, Miles Guo, Rain Sun, Kayden Zhang, Alex Du, Ruihai Wu, Liang Hao, Zhaoshuo Li, Roy Gan, Hao Wang$^\ddagger$, Qian Wang.

%% file: sec/appendix.tex
\section{Appendix}\label{sec:appendix}

\subsection{Task Details}
\label{app:task-details}
\input{sec/task_appendix/task_suite_details}



\input{sec/assets_scene/appendix_content}


%% file: sec/task_appendix/task_suite_details.tex
%

\providecolor{XTaskPink}{HTML}{D94F9A}

\providecommand{\XTaskEntry}[6]{%
  \par\vspace{0.2em}
  \noindent
  \begin{minipage}[t]{0.22\linewidth}
    \vspace{0pt}
    \includegraphics[width=\linewidth]{#2}
    \par\vspace{0.2em}
    {\normalsize\bfseries #1\par}
  \end{minipage}\hfill
  \begin{minipage}[t]{0.75\linewidth}
    \vspace{0pt}
    \normalsize
    #3\par
    \vspace{0.2em}
    \begin{tabularx}{\linewidth}{@{}!{\color{XTaskPink}\vrule width 2pt}@{\hspace{0.2em}}X@{}}
      #4
    \end{tabularx}
    \par\vspace{0.2em}
    {\footnotesize\color{gray}
      \textbf{Embodiment:} #5
      \hspace{0.2em}\textperiodcentered\hspace{0.2em}
      \textbf{Data:} #6
    }
  \end{minipage}
  \par\vspace{0.2em}
}


\medskip
\noindent{\bfseries Reasoning}\par
\smallskip

\noindent{\itshape Generalization}\par

\XTaskEntry
  {Open Drawer}
  {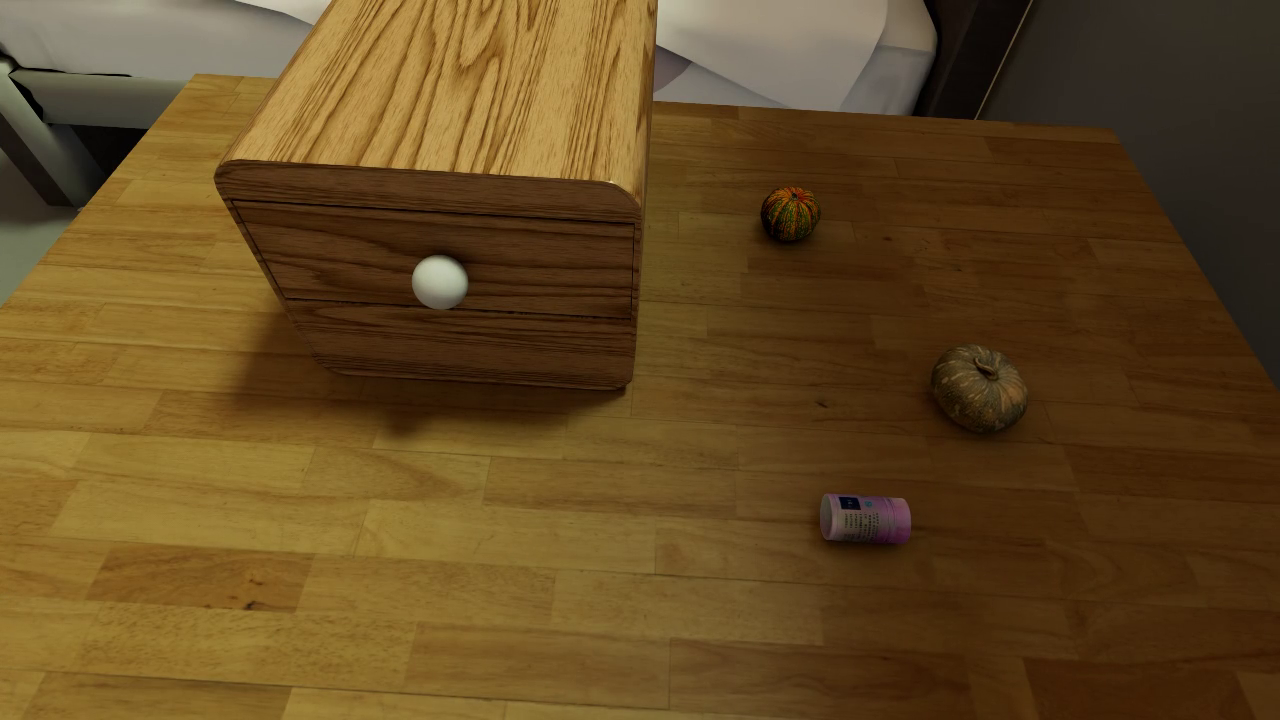}
  {Open the drawer, place the pill bottle inside, close the drawer, and return both arms to their home configuration.}
  {ID uses seen drawers and pill bottles. OOD introduces unseen pill bottles, two distractors, and visual and embodiment changes.}
  {ArtiXon Arm-6A, ARX R5}
  {Auto \textperiodcentered{} 300 demos}

\XTaskEntry
  {Fruit in Basket}
  {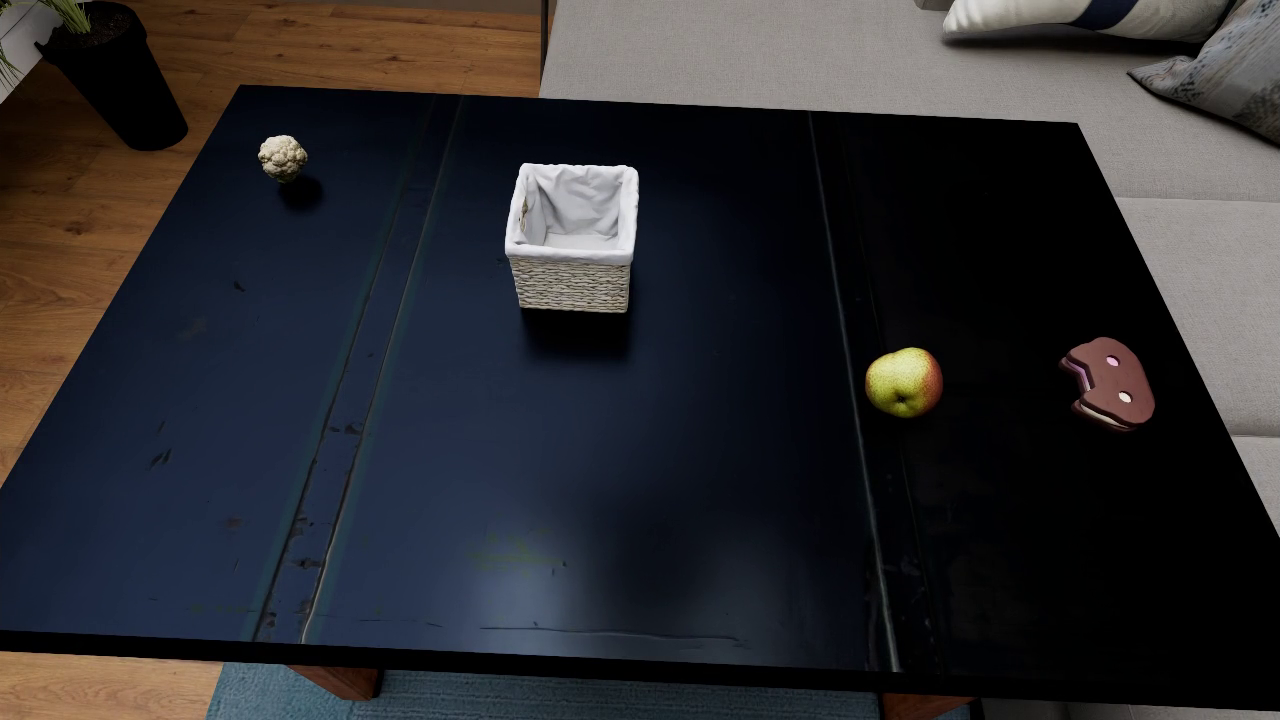}
  {Pick up the fruit, place it inside the basket, and return the arm to its home configuration.}
  {ID uses seen fruits and baskets. OOD introduces unseen fruits, two distractors, and visual and embodiment changes; the basket must remain free of distractors.}
  {ArtiXon Arm-6A, ARX R5, Franka}
  {Auto \textperiodcentered{} 300 demos}

\XTaskEntry
  {Bottle on Plate}
  {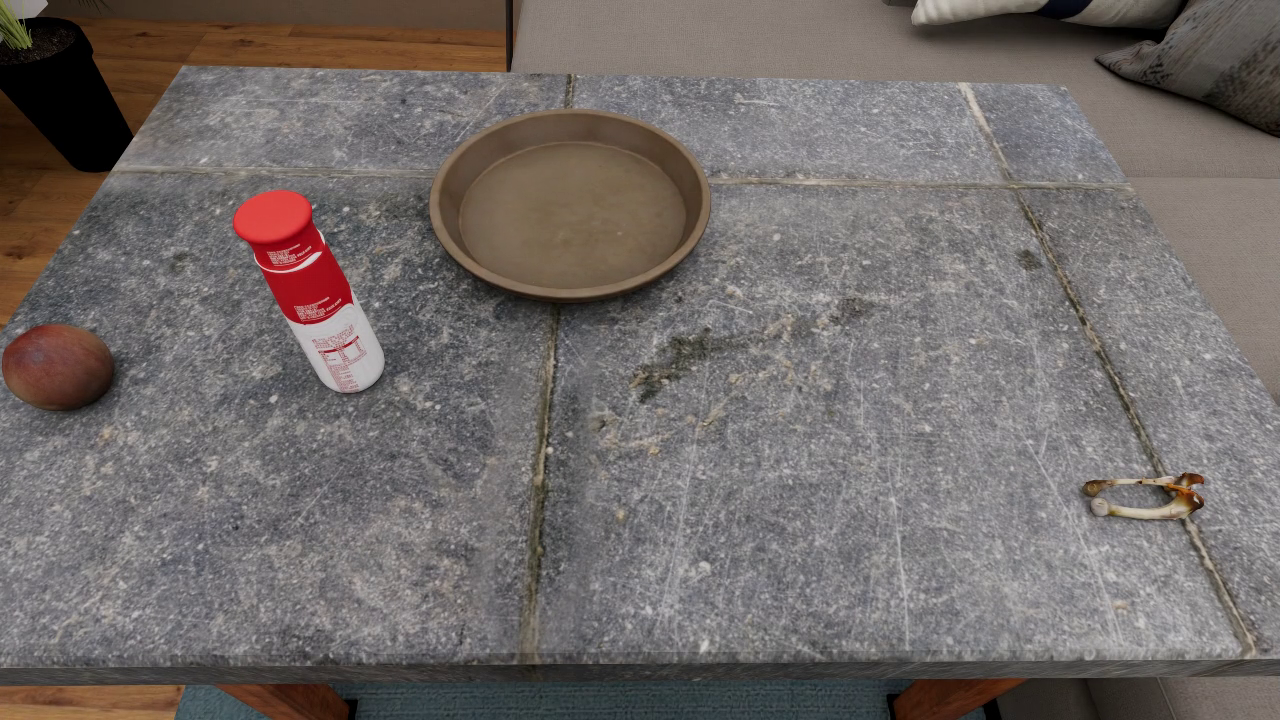}
  {Pick up the bottle, place it upright on the plate, and return the arm to its home configuration.}
  {ID uses seen bottles and plates. OOD introduces unseen bottles and plates, two distractors, and visual and embodiment changes; the bottle must remain upright and the plate free of distractors.}
  {ArtiXon Arm-6A, ARX R5, Franka}
  {Auto \textperiodcentered{} 300 demos}

\XTaskEntry
  {Object Beside Block}
  {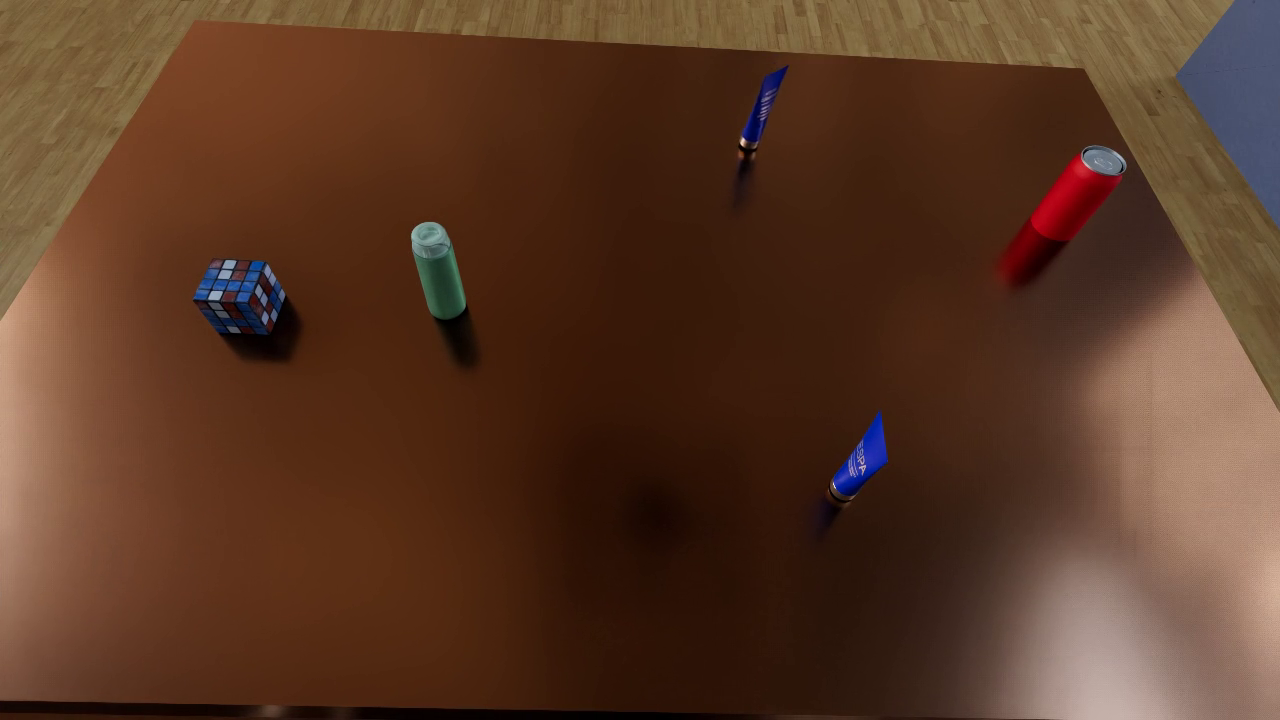}
  {Pick up the green object, place it on the left/right/below/upper side of the reference block, and return the arm to its home configuration.}
  {ID uses seen objects and blocks. OOD introduces unseen objects and blocks, three non-green distractors, and visual and embodiment changes; the instructed placement region must remain clear.}
  {ArtiXon Arm-6A, ARX R5, Franka}
  {Auto \textperiodcentered{} 300 demos}

\XTaskEntry
  {Fruit in Microwave}
  {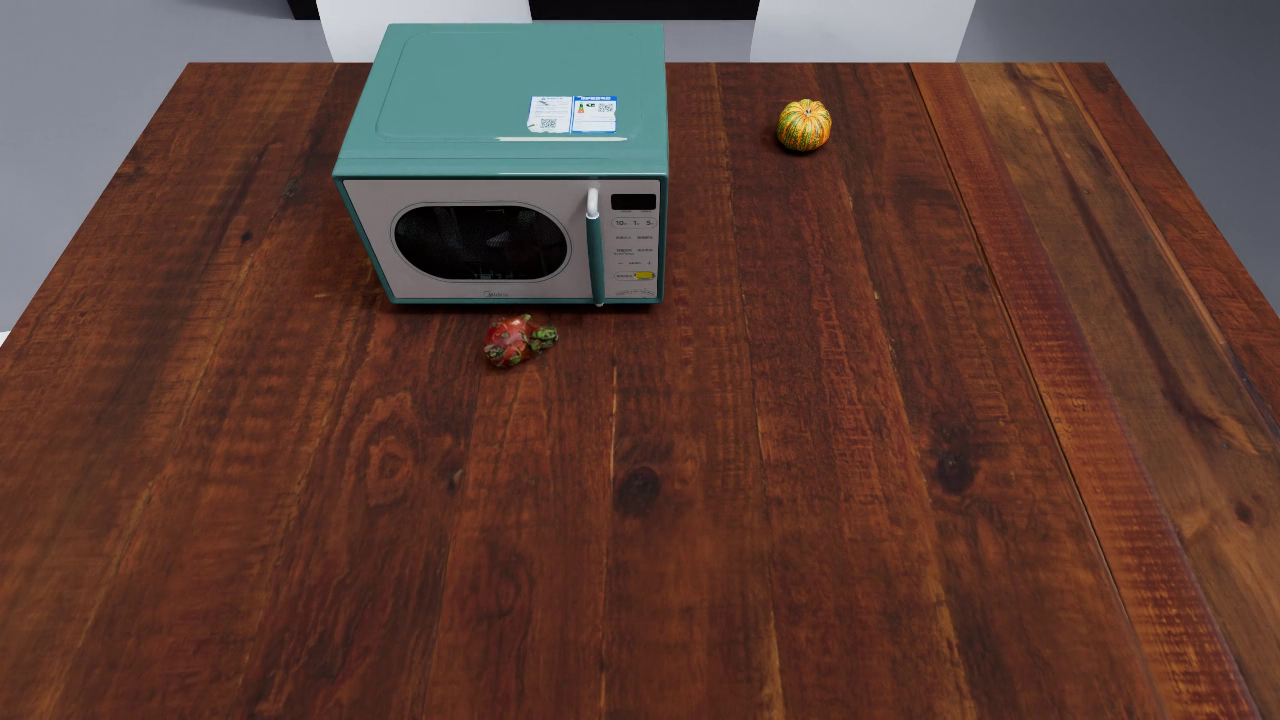}
  {Open the microwave, place the fruit inside, close the door, and return the arm to its home configuration.}
  {ID uses a fixed microwave and seen fruits. OOD introduces unseen fruits, two distractors, and visual and embodiment changes; the microwave must remain free of distractors.}
  {ArtiXon Arm-6A}
  {Teleop \textperiodcentered{} 300 demos}

\XTaskEntry
  {Cup on Plate}
  {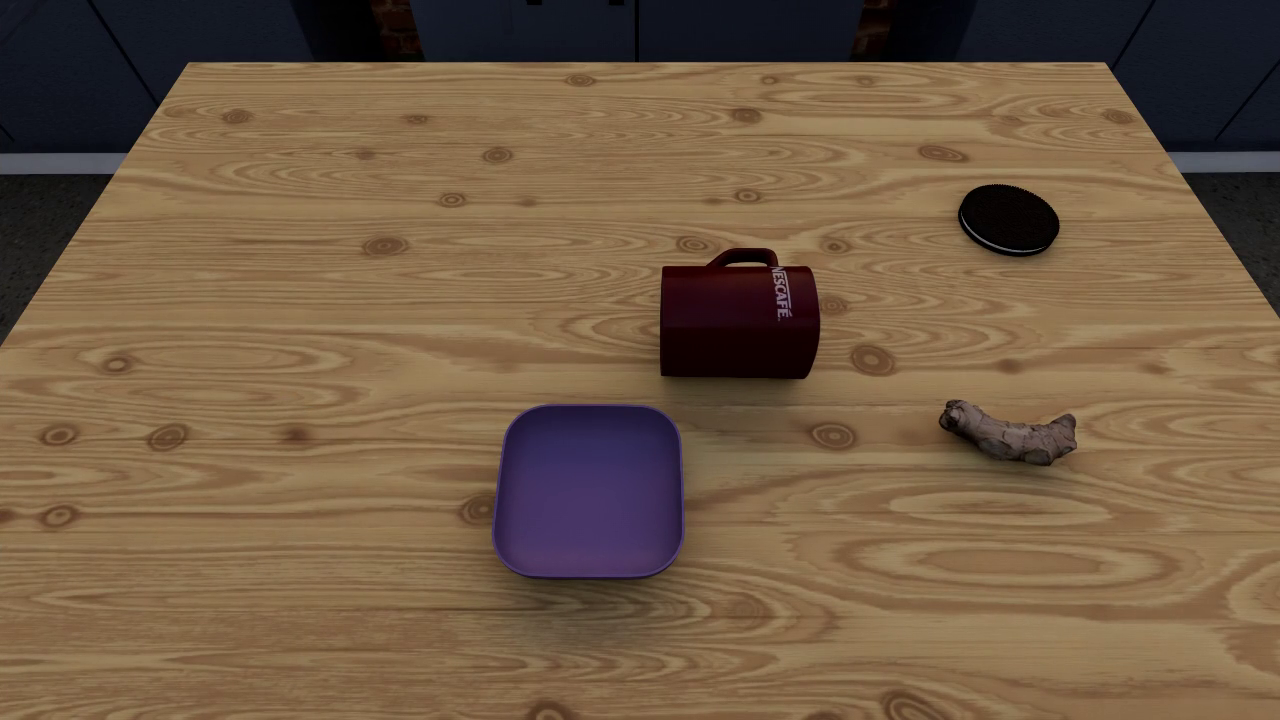}
  {Orient the cup upright, move the plate to the table center, place the cup upright on the plate, and return both arms to their home configurations.}
  {ID and OOD use the same cups and plates. OOD adds two distractors and introduces visual and embodiment changes; the cup must remain upright and the plate free of distractors.}
  {ArtiXon Arm-6A}
  {Teleop \textperiodcentered{} 300 demos}

\XTaskEntry
  {Spoon in Bowl}
  {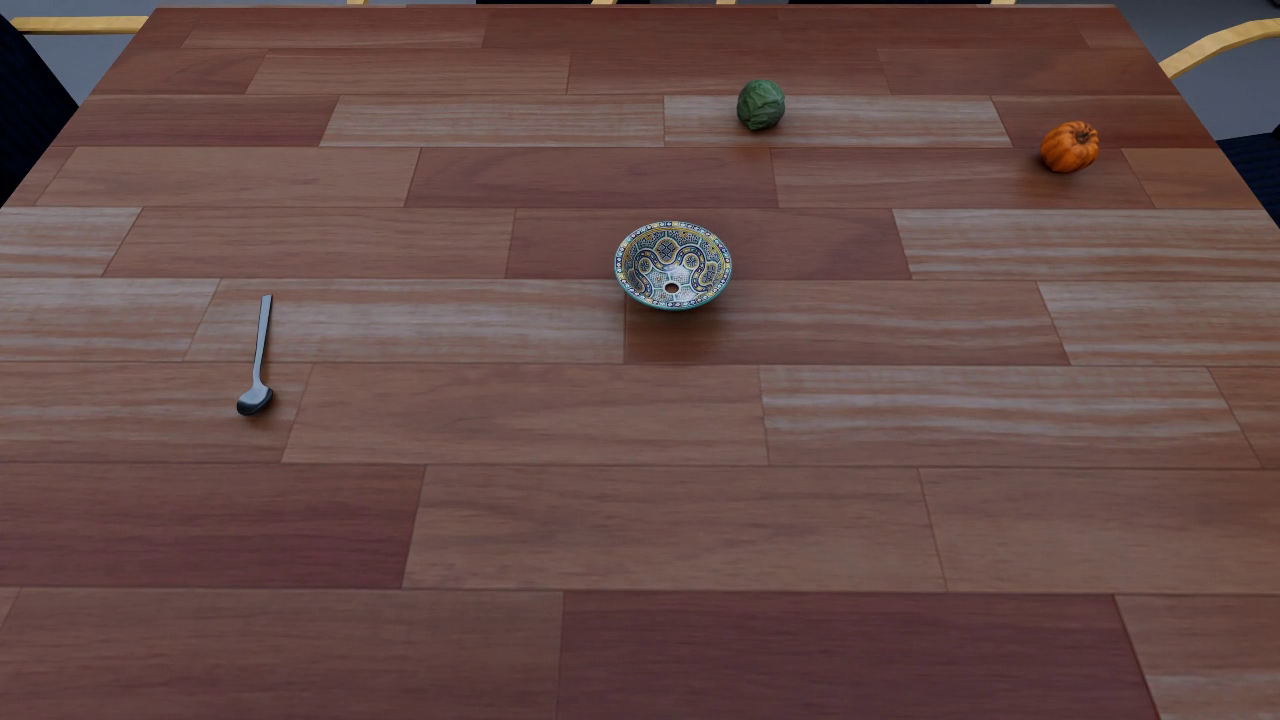}
  {Pick up the spoon, place it inside the bowl, and return the arm to its home configuration.}
  {ID uses a fixed spoon and seen bowls. OOD introduces unseen bowls, two distractors, and visual and embodiment changes; the bowl must remain free of distractors.}
  {ArtiXon Arm-6A, ARX R5, Franka}
  {Auto \textperiodcentered{} 300 demos}

\medskip
\noindent{\itshape Visual Understanding}\par

\XTaskEntry
  {Classify by Shape}
  {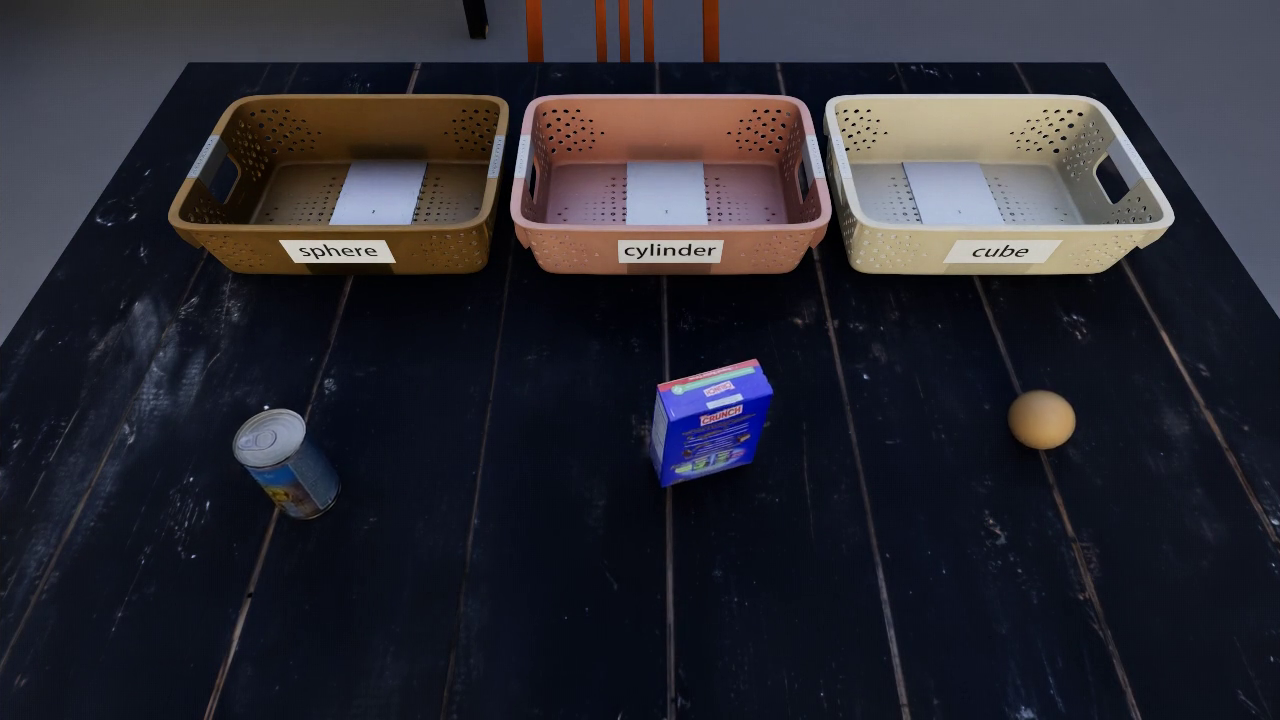}
  {Sort the cube, cylinder, and sphere into the baskets with matching shapes, then return the arm to its home configuration.}
  {ID uses seen instances of the three shapes. OOD replaces all three objects with unseen instances and introduces appearance randomization; the basket remains unchanged.}
  {ArtiXon Arm-6A}
  {Teleop \textperiodcentered{} 300 demos}

\XTaskEntry
  {Classify by Color}
  {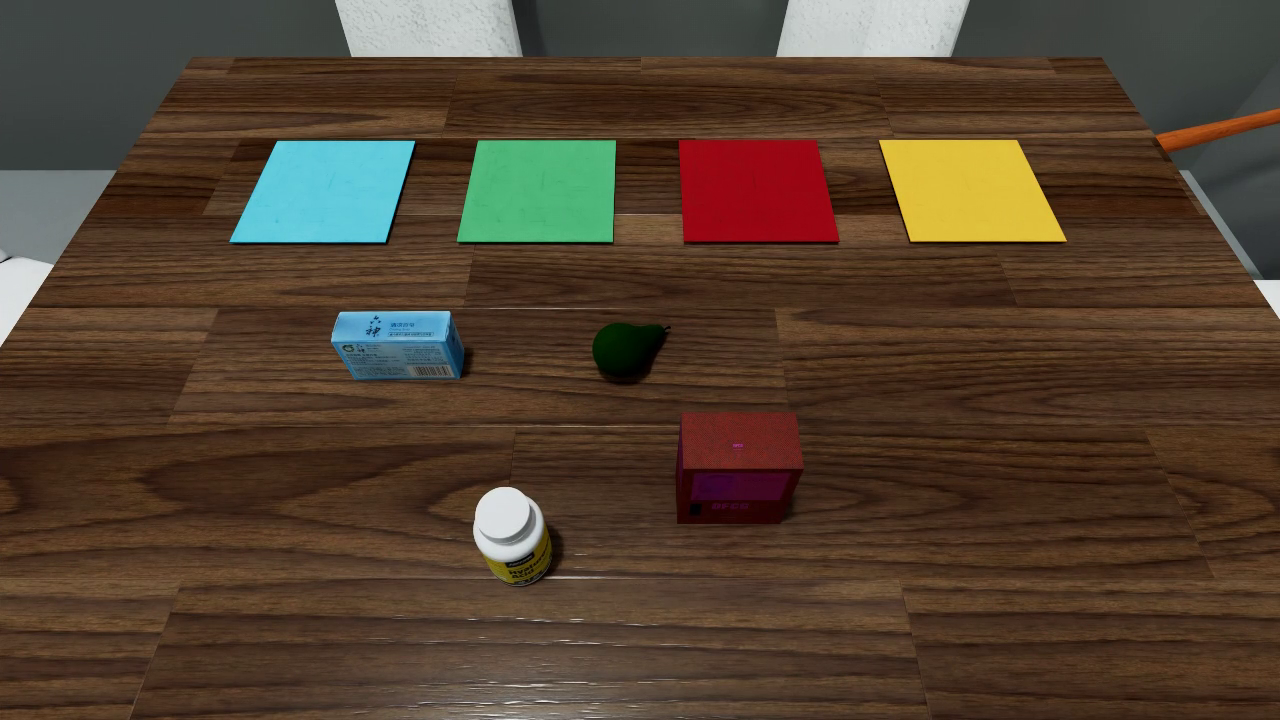}
  {Place each of the four objects on the visible region with the matching color, then return the arm to its home configuration.}
  {ID uses seen objects in blue, green, red, and yellow. OOD replaces them with unseen objects of the same colors and introduces appearance randomization.}
  {ArtiXon Arm-6A}
  {Teleop \textperiodcentered{} 300 demos}

\XTaskEntry
  {Adjust Temperature}
  {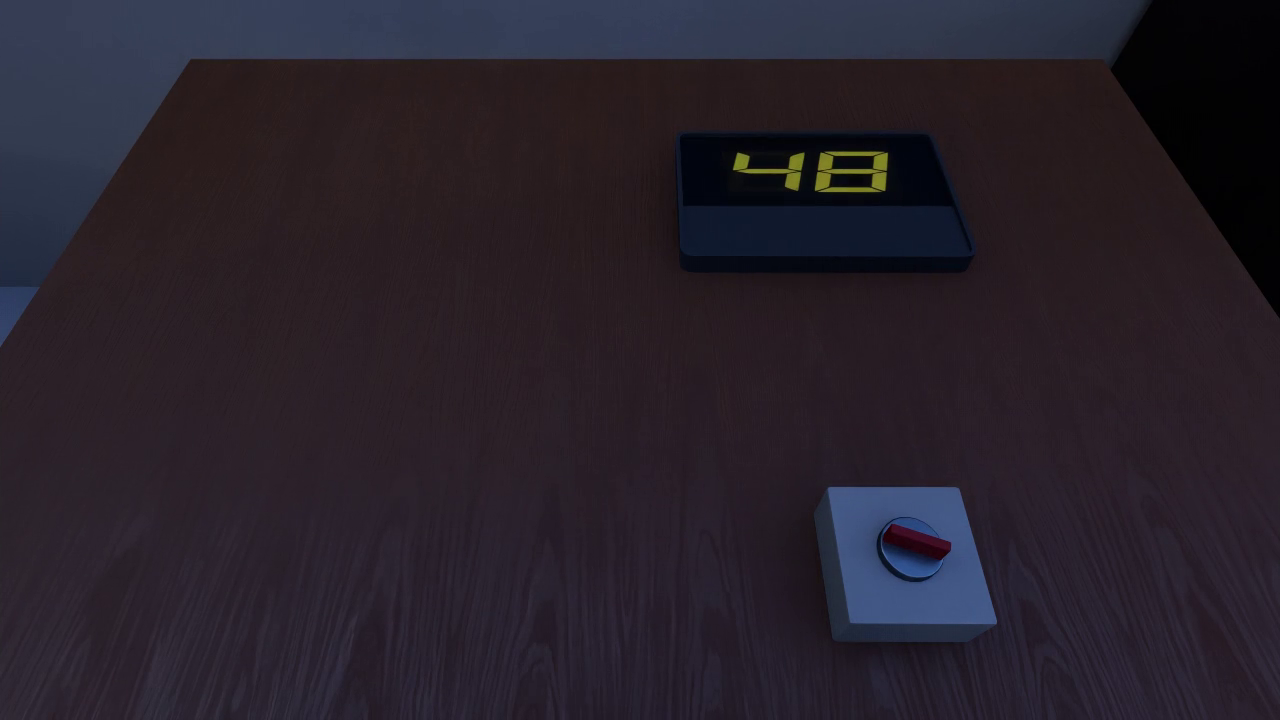}
  {Rotate the knob from the displayed initial temperature to the target temperature, then return the arm to its home configuration.}
  {ID samples temperatures from 0--60. OOD expands the range to 0--99 and introduces appearance randomization while retaining the same adjustment rule.}
  {ArtiXon Arm-6A, ARX R5, Franka}
  {Auto \textperiodcentered{} 300 demos}

\XTaskEntry
  {Image Puzzle}
  {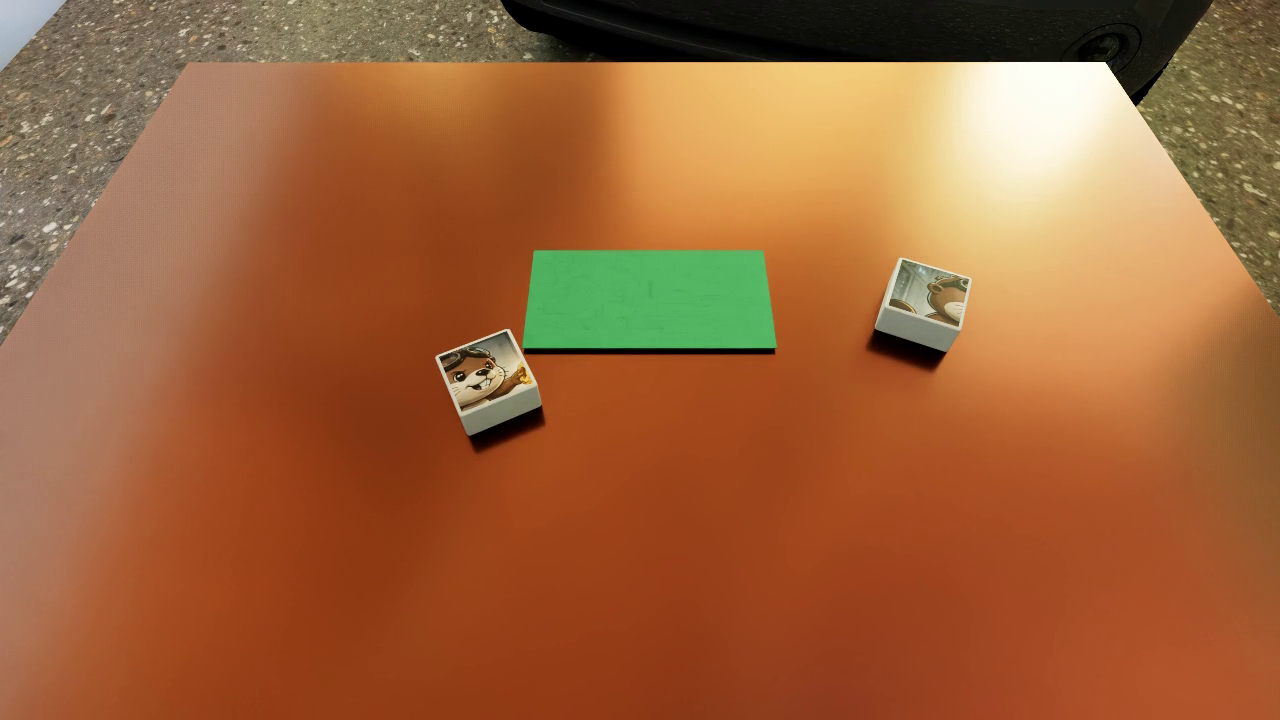}
  {Place and align the two puzzle pieces on the green region to form a complete image, then return the arm to its home configuration.}
  {ID uses seen image pairs. OOD introduces unseen image pairs together with visual and embodiment changes; the two-piece assembly rule remains unchanged.}
  {ArtiXon Arm-6A}
  {Teleop \textperiodcentered{} 300 demos}

\XTaskEntry
  {Balance Scale}
  {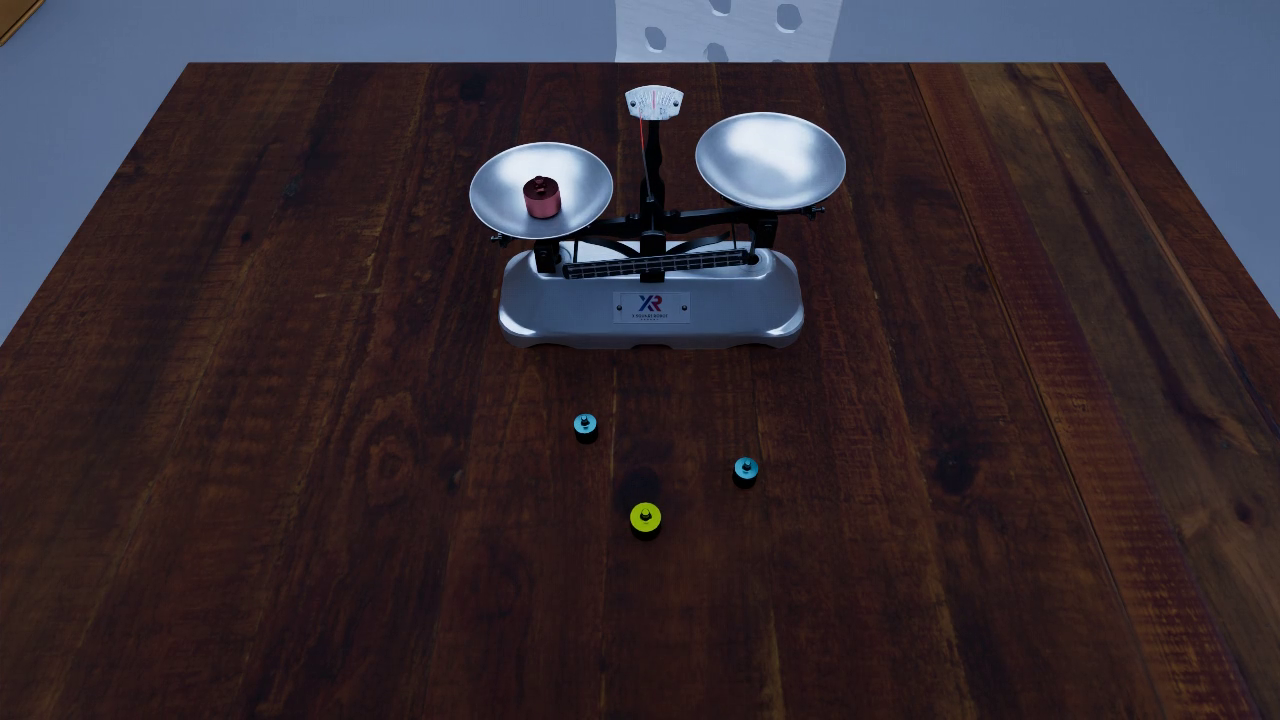}
  {Select and place the required weights from 3-4 weights onto the lighter pan until the scale is balanced, then return the arm to its home configuration.}
  {ID contains seen cases requiring one or two added weights. OOD introduces unseen weight configurations, including cases requiring three weights, together with visual and embodiment changes.}
  {ArtiXon Arm-6A}
  {Teleop \textperiodcentered{} 300 demos}

\XTaskEntry
  {Rotate Book}
  {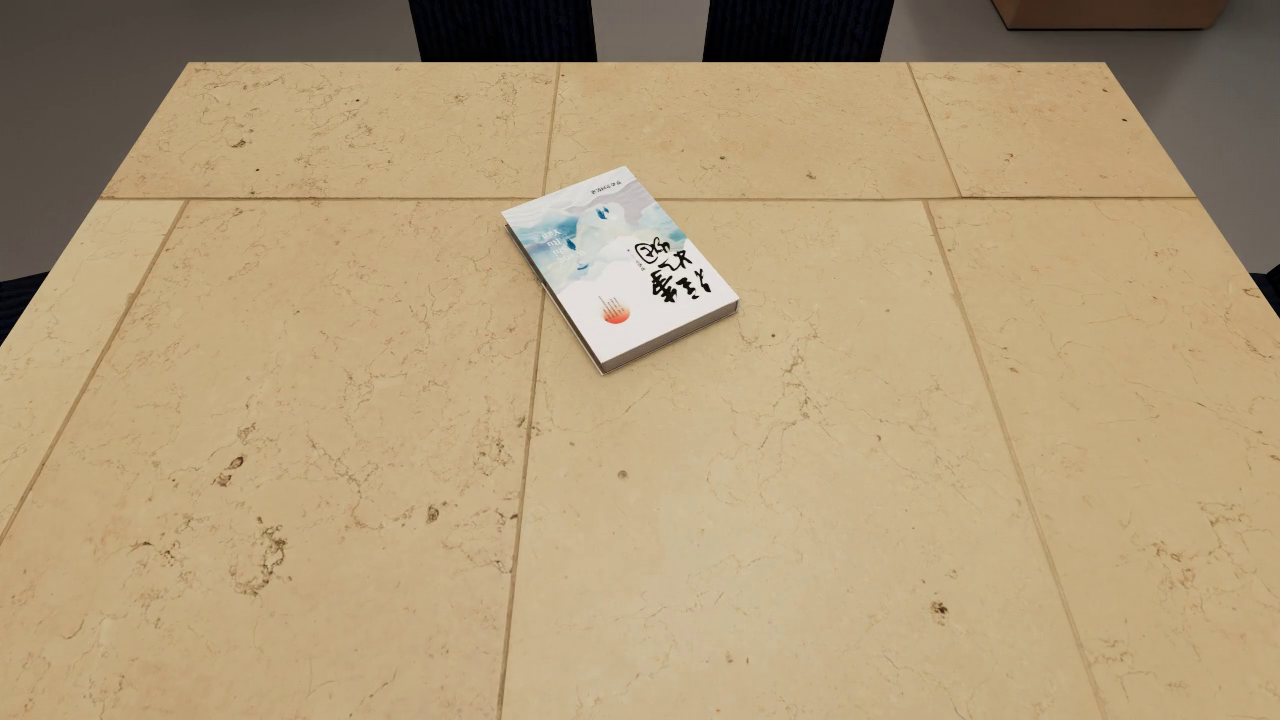}
  {Rotate the book on the table into the reader-facing orientation, then return both arms to their home configurations.}
  {ID uses seen books. OOD replaces them with unseen books and introduces appearance randomization while retaining the same target orientation.}
  {ArtiXon Arm-6A}
  {Teleop \textperiodcentered{} 300 demos}

\XTaskEntry
  {Spell Word}
  {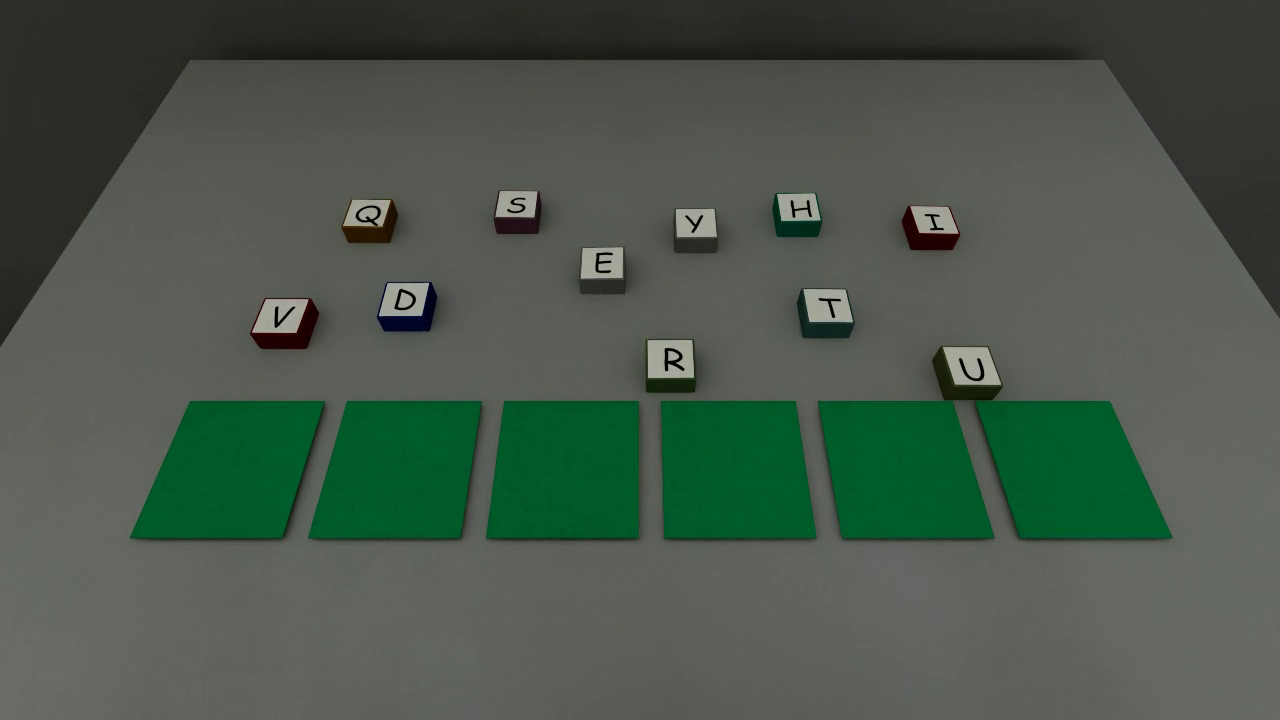}
  {Select the required letter blocks and place them in order to spell the displayed word, then return the arm to its home configuration.}
  {ID uses seen three-letter or four-letter words. OOD introduces unseen words and extends the task to five-letter and six-letter words, together with appearance randomization.}
  {ArtiXon Arm-6A}
  {Teleop \textperiodcentered{} 300 demos}

\medskip
\noindent{\itshape Language Understanding}\par

\XTaskEntry
  {Fruit to Basket}
  {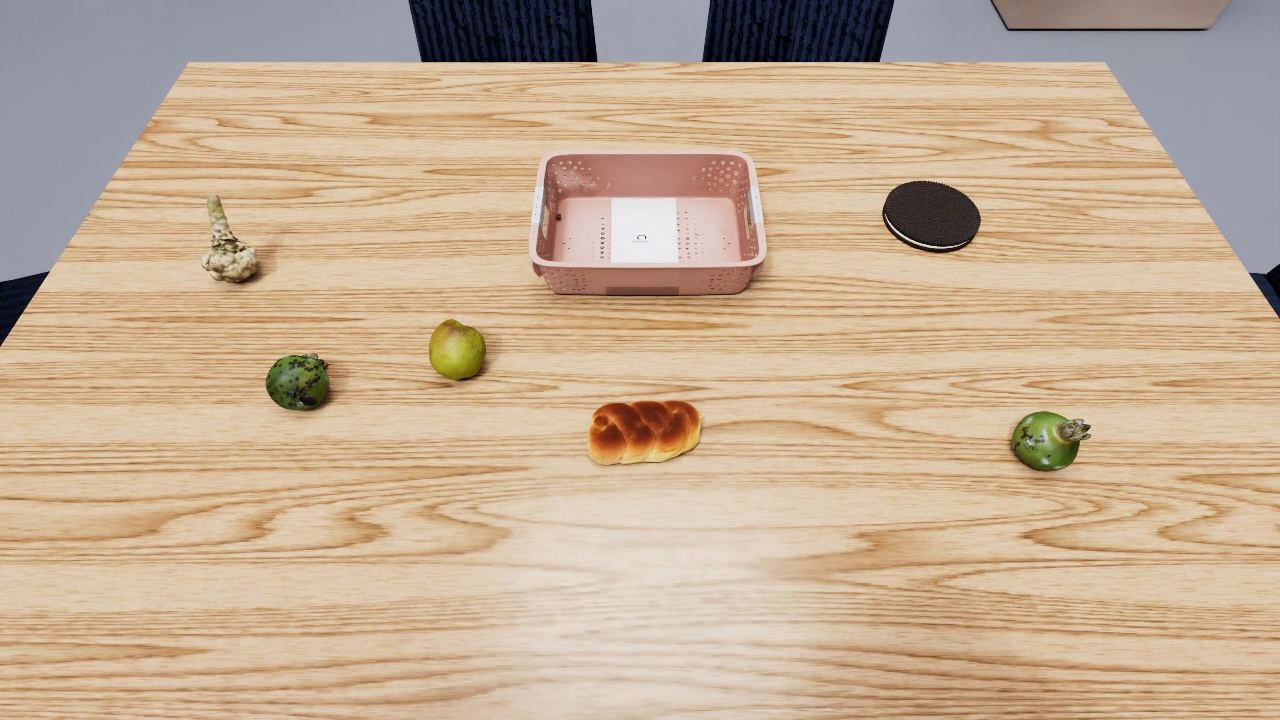}
  {Place all three fruits in the pink basket, keep the non-fruit objects outside, and return the arm to its home configuration.}
  {ID uses seen fruits and a fixed distractor pool. OOD replaces the fruits with unseen instances and introduces appearance randomization; the distractors and basket remain unchanged.}
  {ArtiXon Arm-6A, ARX R5, Franka}
  {Teleop, Auto \textperiodcentered{} 300 demos}

\XTaskEntry
  {AND Expression}
  {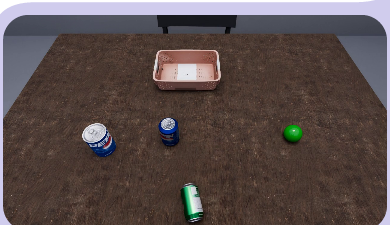}
  {Place all and only the objects satisfying both attributes in the instruction into the basket, then return the arm to its home configuration.}
  {The ID training instructions specify a single color or shape. This setting evaluates unseen conjunctive color--shape expressions, together with appearance randomization, using the same object pool.}
  {ArtiXon Arm-6A}
  {Teleop \textperiodcentered{} 300 demos}

\XTaskEntry
  {Confusing Instruction}
  {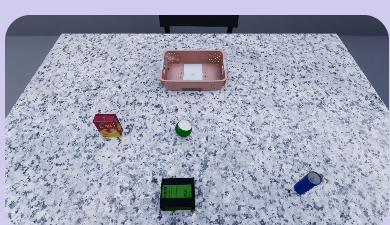}
  {Follow the operative placement instruction while ignoring the surrounding distractor sentences, then return the arm to its home configuration.}
  {The ID training instructions contain only a direct single-attribute command. This setting adds irrelevant sentences before or after that command and introduces appearance randomization.}
  {ArtiXon Arm-6A}
  {Teleop \textperiodcentered{} 300 demos}

\XTaskEntry
  {OR Expression}
  {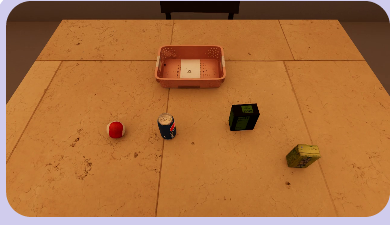}
  {Place all and only the objects satisfying either attribute in the instruction into the basket, then return the arm to its home configuration.}
  {The ID training instructions specify a single color or shape. This setting evaluates unseen disjunctive expressions over colors and shapes, together with appearance randomization, using the same object pool.}
  {ArtiXon Arm-6A}
  {Teleop \textperiodcentered{} 300 demos}

\XTaskEntry
  {NOT Expression}
  {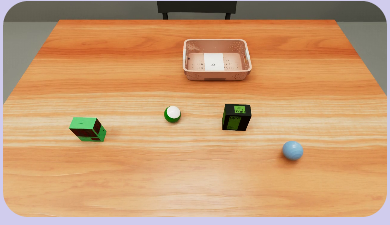}
  {Place every object except those excluded by the instruction into the basket, then return the arm to its home configuration.}
  {The ID training instructions specify a positive single attribute. This setting evaluates unseen negated or exclusion expressions, together with appearance randomization, using the same object pool.}
  {ArtiXon Arm-6A}
  {Teleop \textperiodcentered{} 300 demos}

\XTaskEntry
  {Specific Position}
  {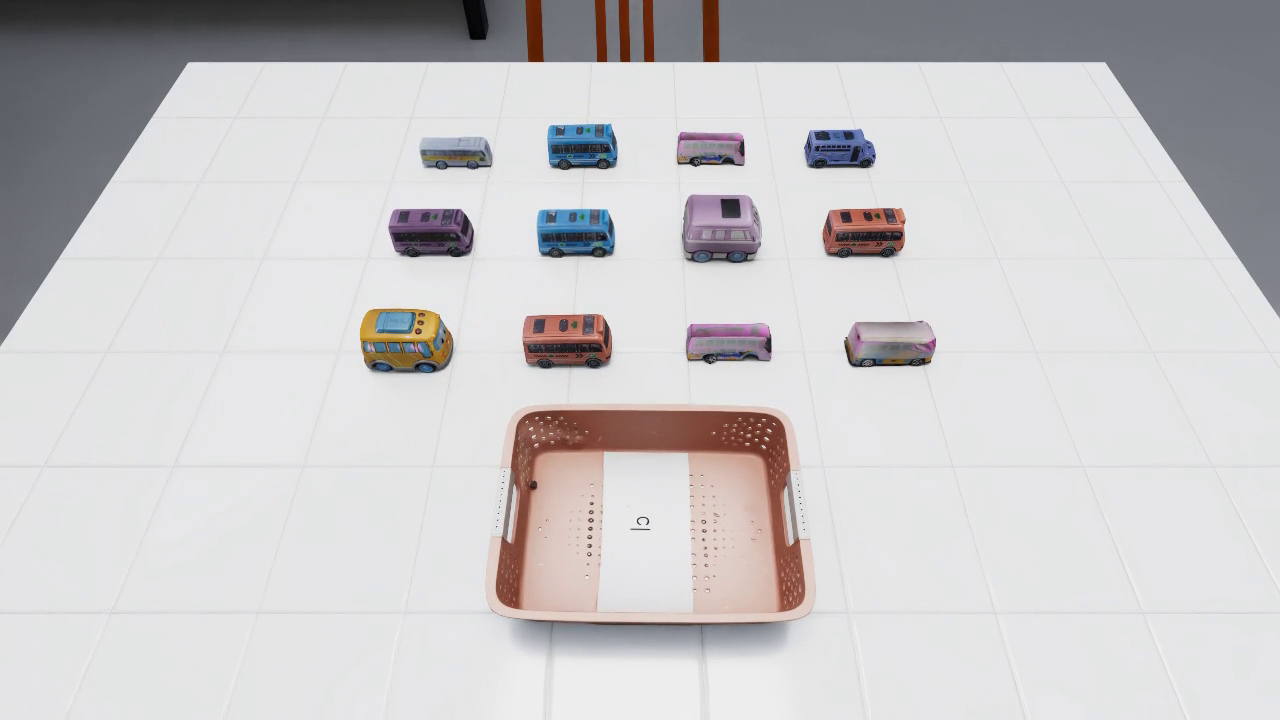}
  {Pick the object at the instructed row and column, place it in the basket, and return the arm to its home configuration.}
  {ID uses $2\times4$ and $3\times3$ object grids. OOD changes the layouts to $2\times3$ and $3\times4$ grids and introduces visual and embodiment changes while retaining the row--column instruction format.}
  {ArtiXon Arm-6A}
  {Teleop \textperiodcentered{} 300 demos}

\medskip
\noindent{\itshape Memory}\par

\XTaskEntry
  {Track Under Cup}
  {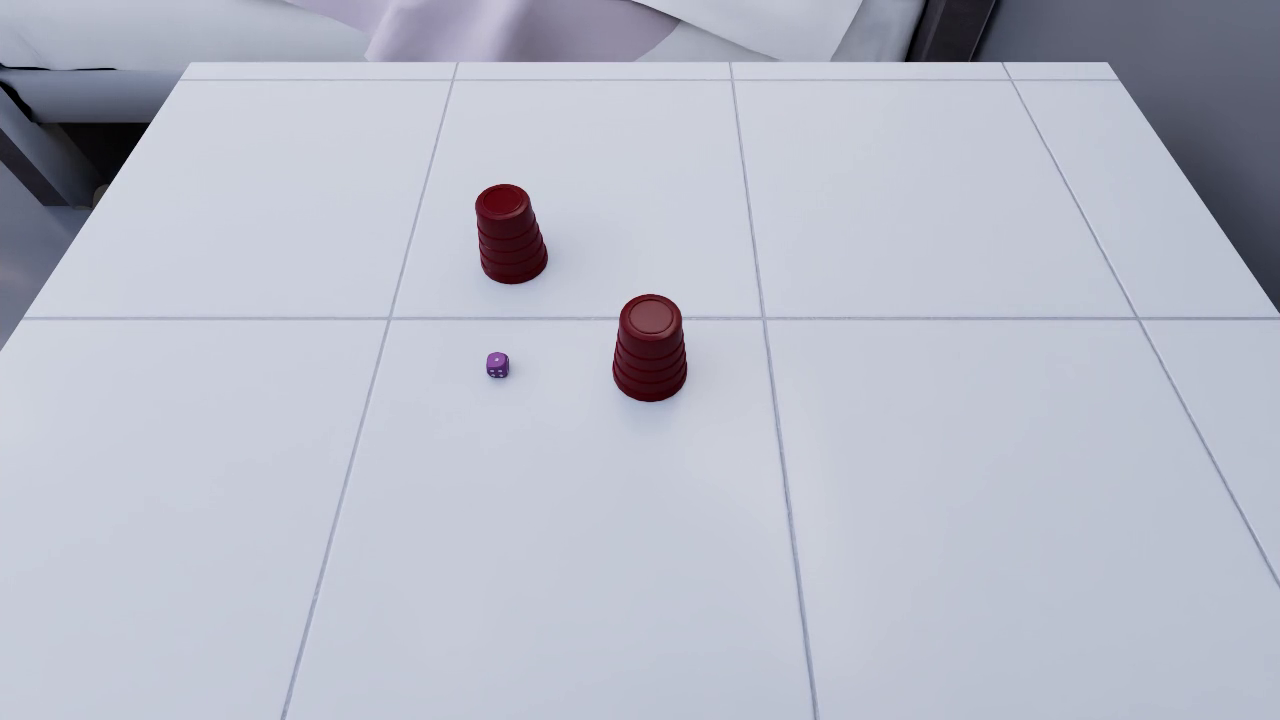}
  {Cover the die with the target cup, swap the two cups once, lift the cup that still covers the die, and return the arms home.}
  {ID uses one seen cup type. OOD replaces both cups with an unseen matched pair and introduces visual and embodiment changes while retaining the cover--swap--reveal sequence.}
  {ArtiXon Arm-6A}
  {Teleop \textperiodcentered{} 300 demos}

\XTaskEntry
  {Find Drawer}
  {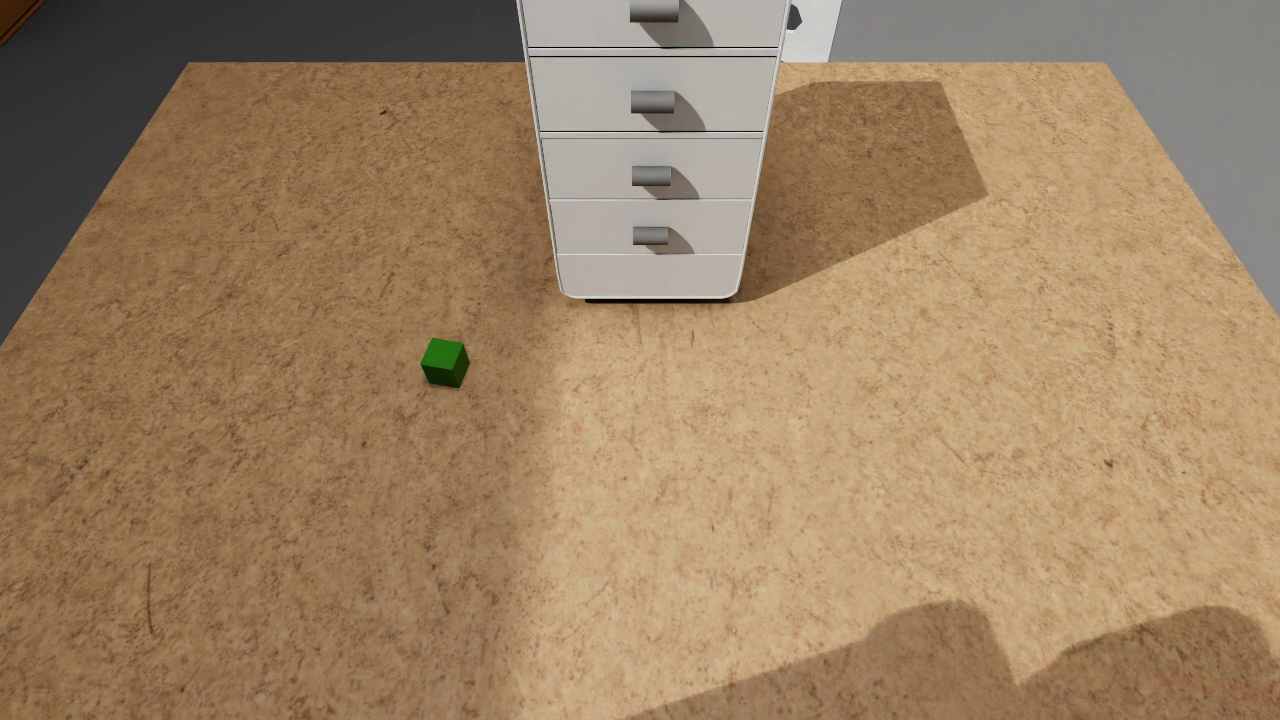}
  {Place the block in the instructed drawer, close it, wait, reopen the same drawer, retrieve the block, close it again, and return the arm home.}
  {ID uses a three-level drawer. OOD replaces it with a four-level drawer, updates the level names in the instruction, and introduces appearance randomization.}
  {ArtiXon Arm-6A}
  {Teleop \textperiodcentered{} 300 demos}

\XTaskEntry
  {Press in Order}
  {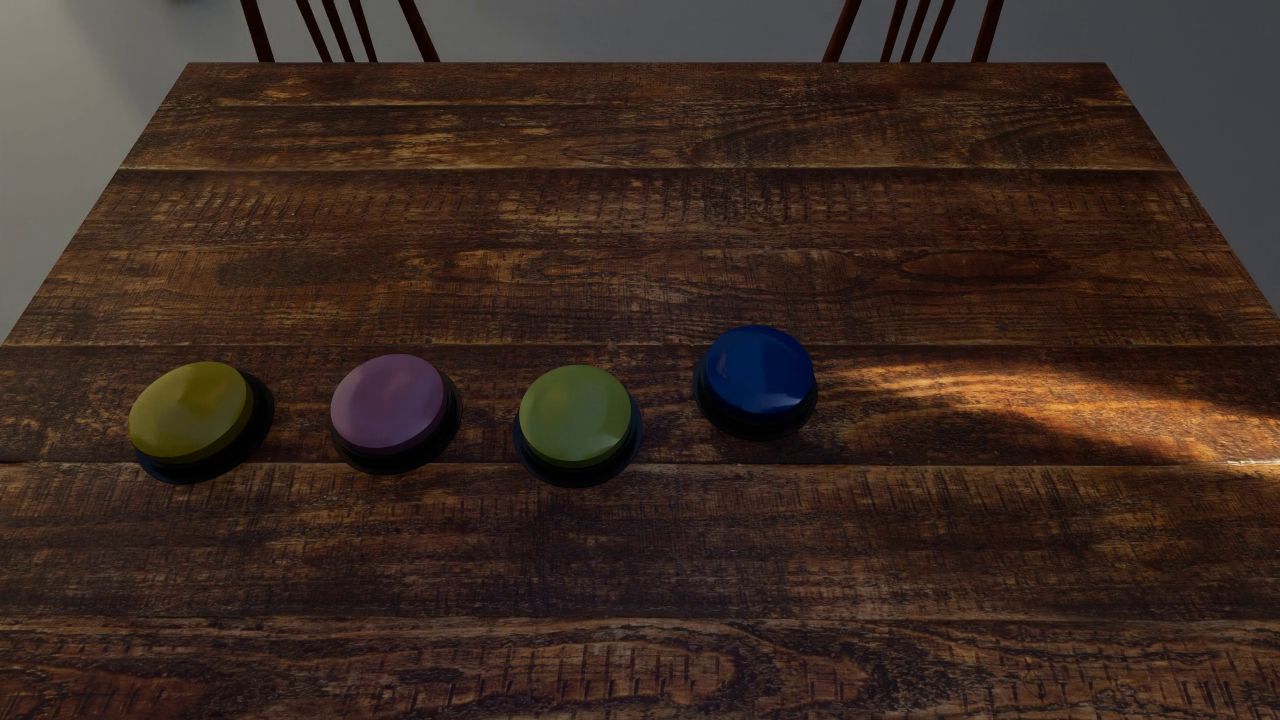}
  {Press the colored buttons in the instructed order, then return the arm to its home configuration.}
  {ID uses seen three-button and four-button sequences. OOD introduces unseen orders over the same button colors together with appearance randomization.}
  {ArtiXon Arm-6A}
  {Teleop \textperiodcentered{} 300 demos}

\XTaskEntry
  {Press N Times}
  {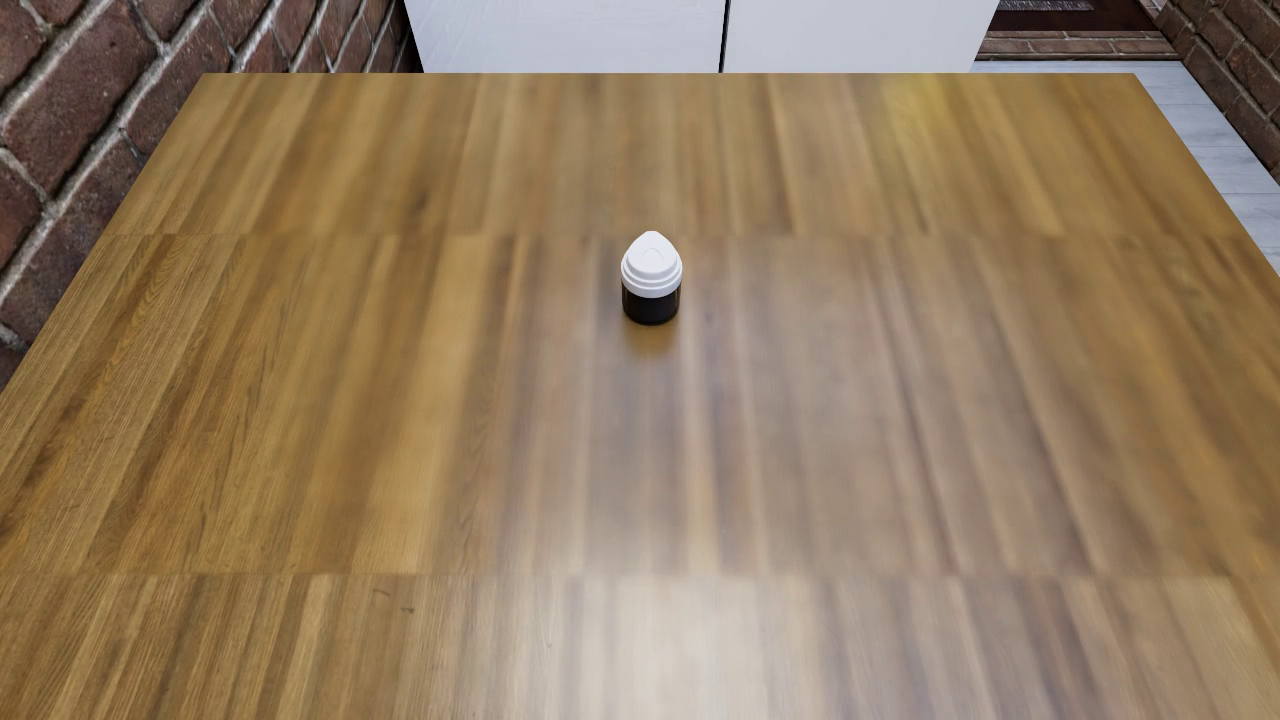}
  {Stabilize the pump bottle, press the pump head the instructed number of times, and return both arms to their home configurations.}
  {ID requests three to six presses. OOD requests two, seven, or eight presses and introduces appearance randomization while using the same bottle.}
  {ArtiXon Arm-6A, ARX R5, Franka}
  {Teleop, Auto \textperiodcentered{} 300 demos}

\XTaskEntry
  {Put Block Back}
  {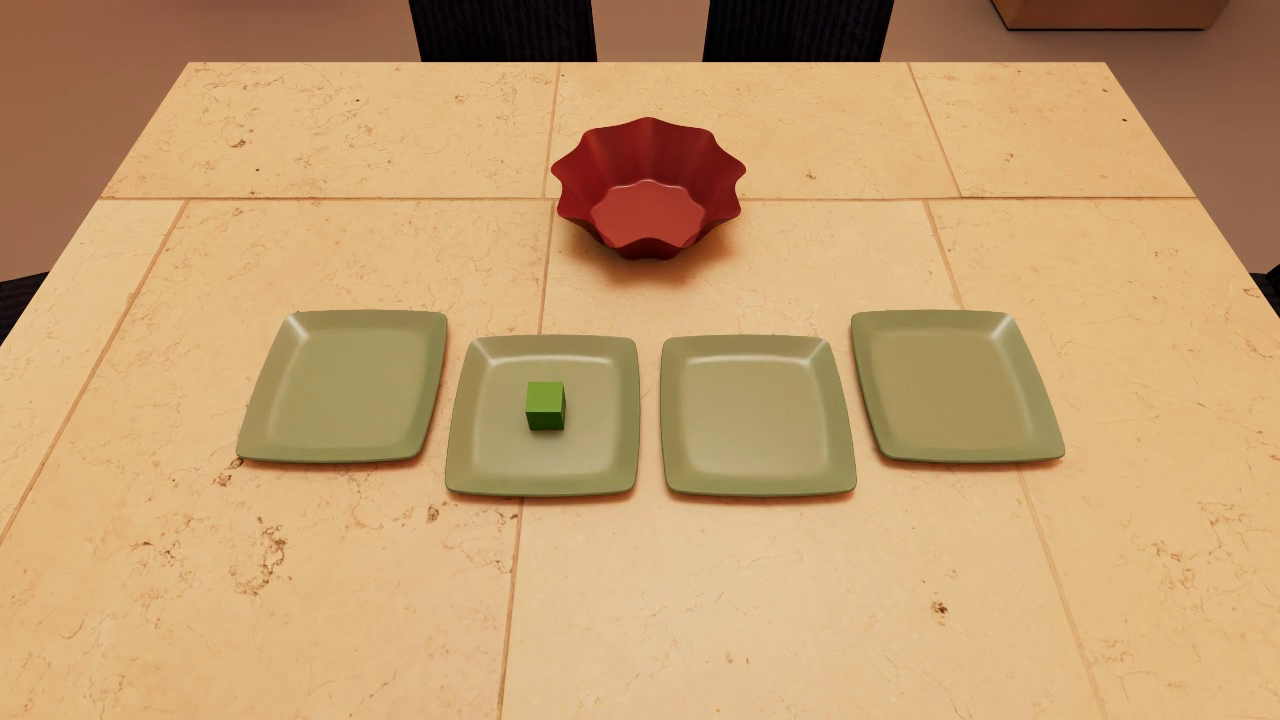}
  {Move the instructed colored block to the central plate, keep it there briefly, return it to its original plate, and return the arm home.}
  {ID uses three candidate source plates. OOD increases this set to four plates and introduces appearance randomization while retaining the round-trip requirement.}
  {ArtiXon Arm-6A}
  {Teleop \textperiodcentered{} 300 demos}

\medskip
\noindent{\bfseries Manipulation}\par
\smallskip

\noindent{\itshape Precision Operation}\par

\XTaskEntry
  {Ring onto Rod}
  {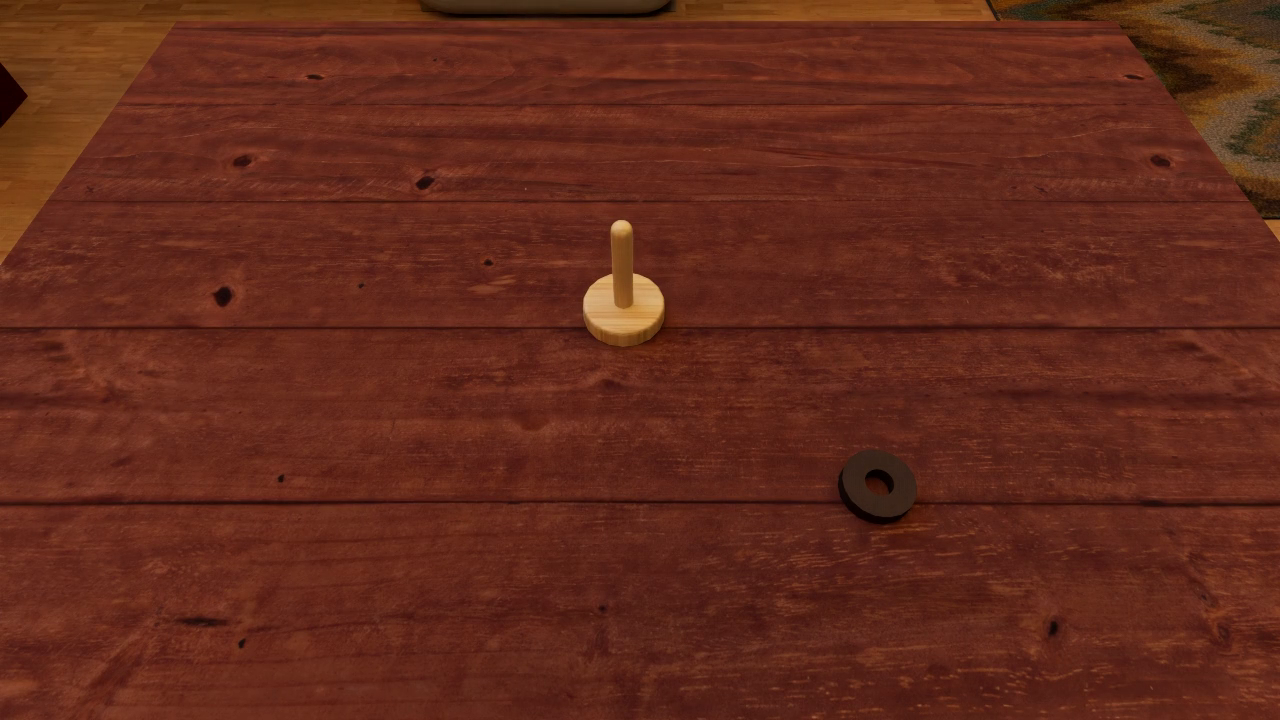}
  {Pick up the ring, place it over the vertical rod, release it securely, and return the arm to its home configuration.}
  {ID and OOD use the same rings and rod. OOD changes the visual environment while retaining the same asset.}
  {ArtiXon Arm-6A, ARX R5, Franka}
  {Auto \textperiodcentered{} 300 demos}

\XTaskEntry
  {Stack Blocks}
  {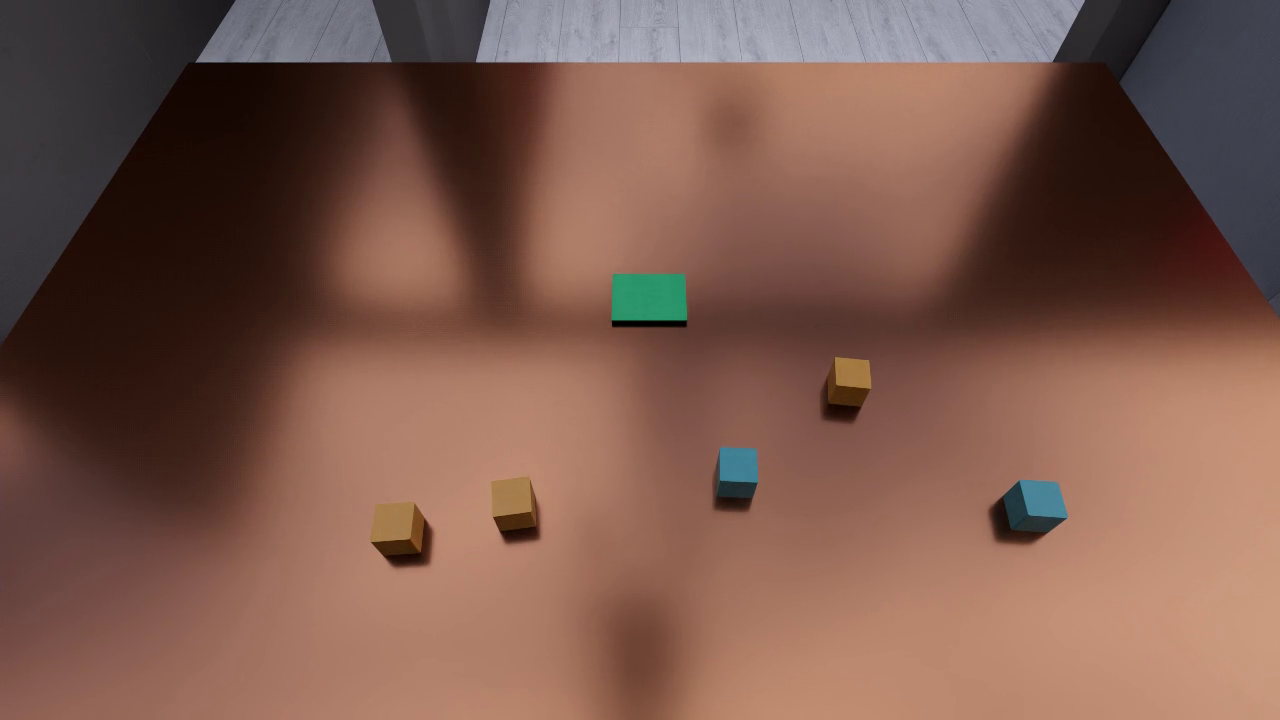}
  {Place one block on the green region, stack two more blocks to form a three-layer tower, and return the arm to its home configuration.}
  {ID and OOD use the same blocks, target region, and tower height. OOD changes only the visual environment.}
  {ArtiXon Arm-6A, ARX R5, Franka}
  {Teleop, Auto \textperiodcentered{} 300 demos}

\XTaskEntry
  {Plug in Charger}
  {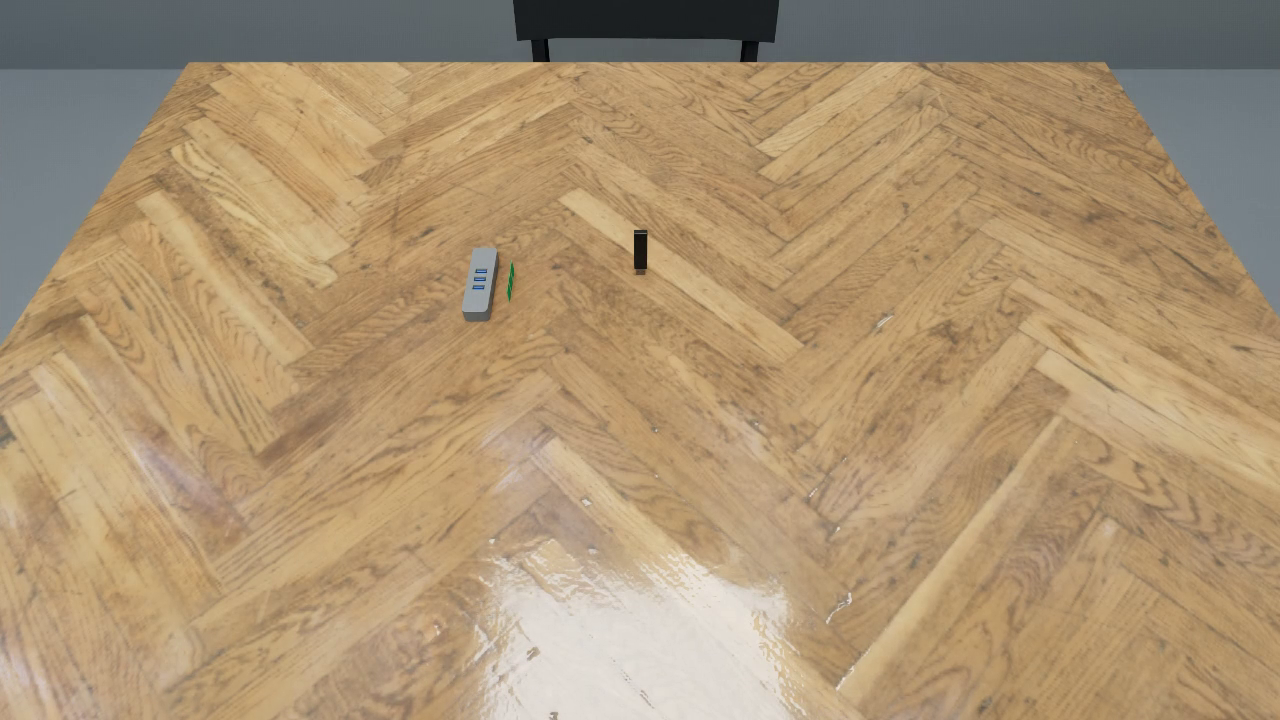}
  {Pick up the USB stick, align it with the instructed left, middle, or right port, insert it fully, and return the arm home.}
  {ID and OOD use the same USB stick, hub, ports, and instructions. OOD changes only the visual environment.}
  {ArtiXon Arm-6A, ARX R5, Franka}
  {Auto \textperiodcentered{} 300 demos}

\XTaskEntry
  {Cup Tower}
  {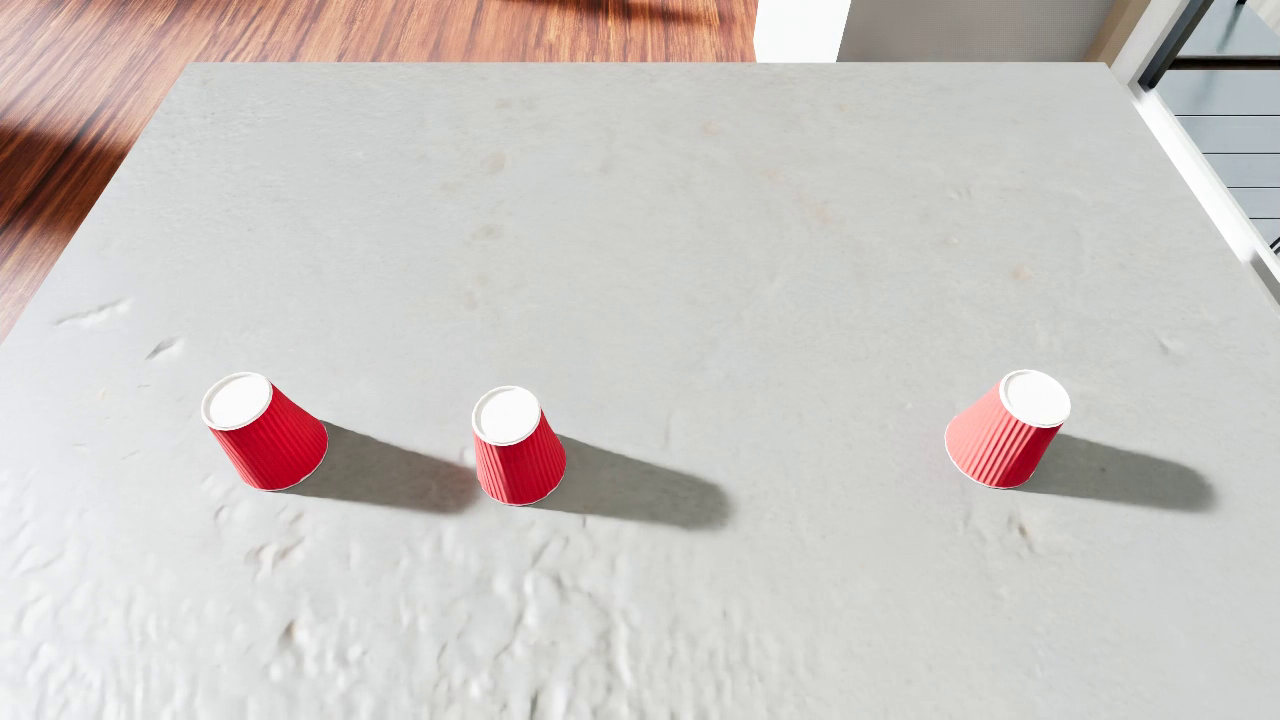}
  {Place two inverted cups side by side, stack the third cup across them to form a triangular tower, and return the arm home.}
  {ID and OOD use the same three cups and assembly order. OOD changes only the visual environment.}
  {ArtiXon Arm-6A}
  {Teleop \textperiodcentered{} 300 demos}

\XTaskEntry
  {Flowers in Vase}
  {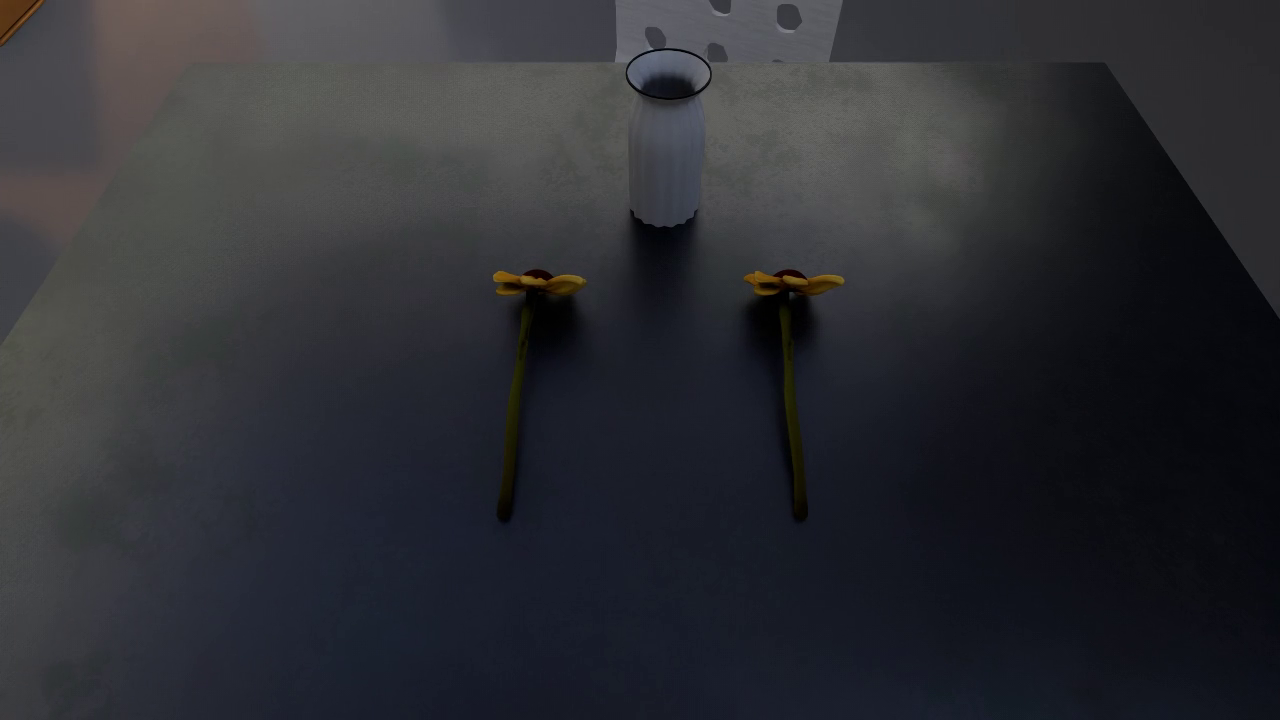}
  {Pick up both flowers by their stems, insert them upright into the vase, and return the arm to its home configuration.}
  {ID and OOD use the same flowers and vase pool. OOD changes only the visual environment.}
  {ArtiXon Arm-6A}
  {Teleop \textperiodcentered{} 300 demos}

\medskip
\noindent{\itshape Bimanual Coordination}\par

\XTaskEntry
  {Object to Cup}
  {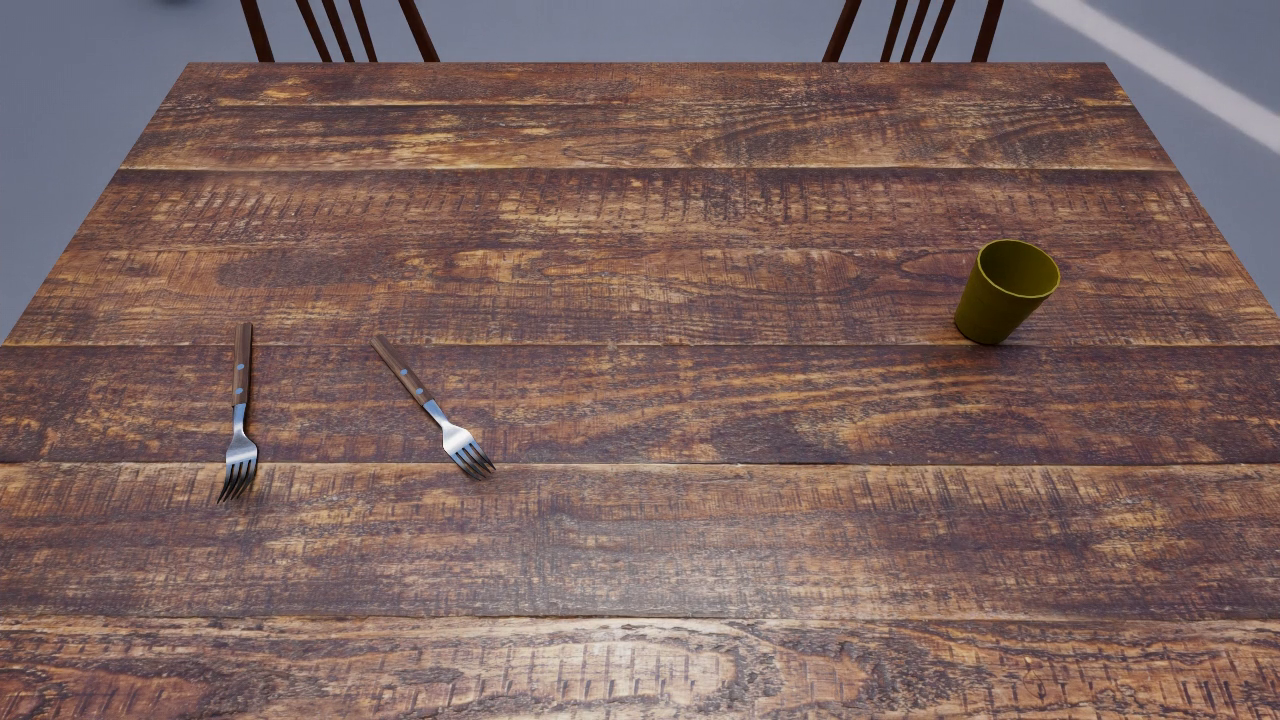}
  {Hold the cup with the right hand, place two objects into it with the left hand, set the cup back on the table, and return both arms home.}
  {ID uses seen small objects and cups. OOD replaces the inserted objects with unseen instances and introduces visual changes while retaining the same cups.}
  {ArtiXon Arm-6A}
  {Teleop \textperiodcentered{} 300 demos}

\XTaskEntry
  {Hand Over}
  {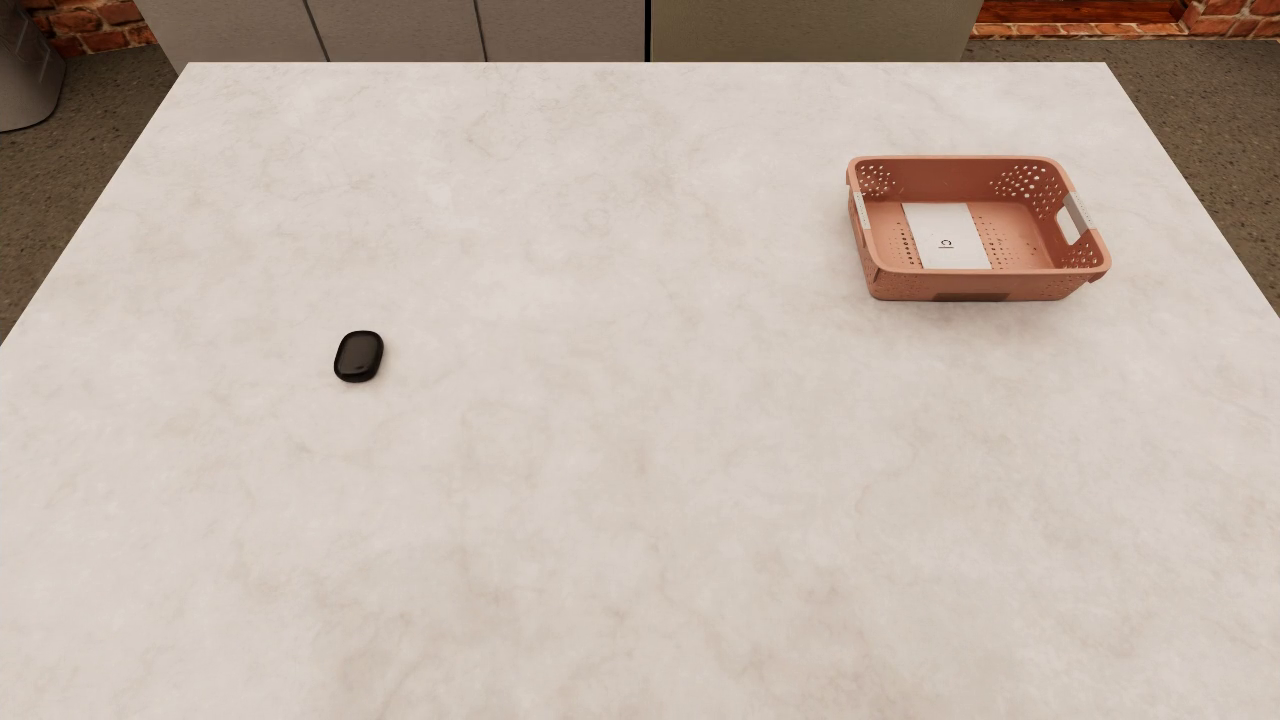}
  {Pick up the object with the left hand, transfer it to the right hand, place it in the right-side basket, and return both arms home.}
  {ID and OOD use the same object and basket pools. OOD changes only the visual environment.}
  {ArtiXon Arm-6A}
  {Teleop \textperiodcentered{} 300 demos}

\XTaskEntry
  {Glasses on Shelf}
  {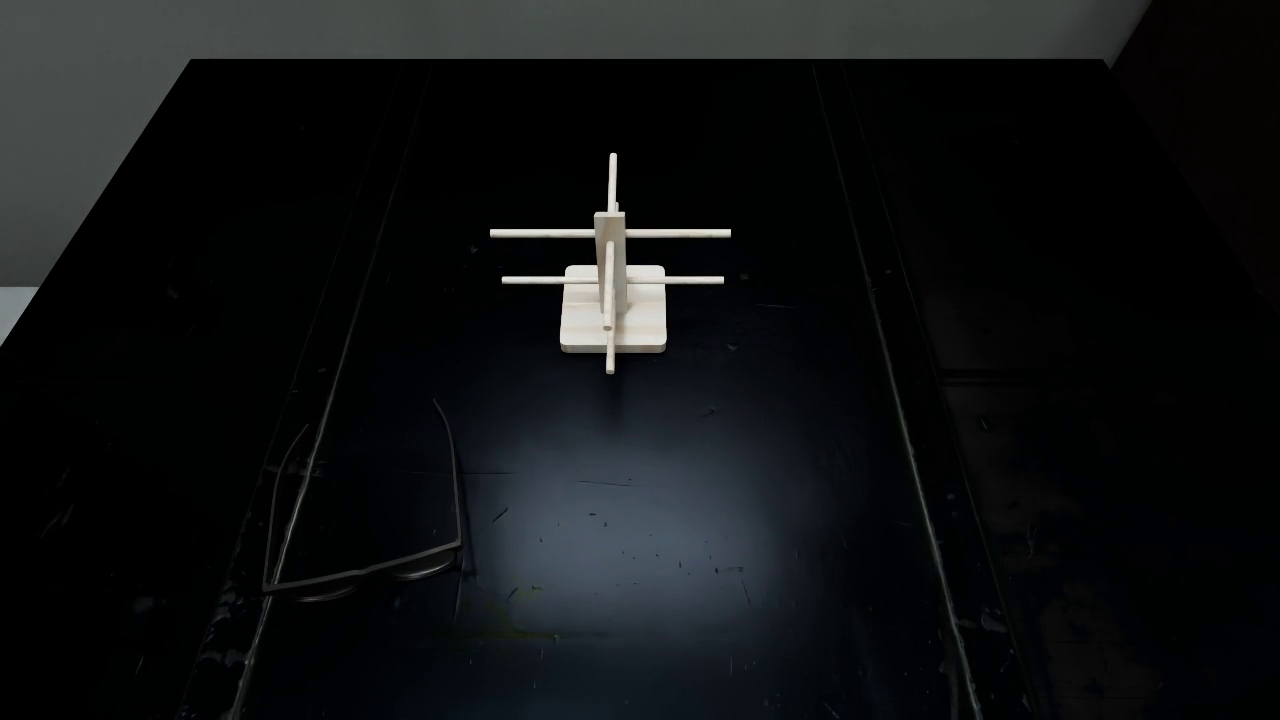}
  {Pick up and reorient the glasses, hang them securely on the rack using both arms, and return both arms to their home configurations.}
  {ID and OOD use the same glasses and rack. OOD changes only the visual environment.}
  {ArtiXon Arm-6A}
  {Teleop \textperiodcentered{} 300 demos}

\XTaskEntry
  {Hand Over Coin}
  {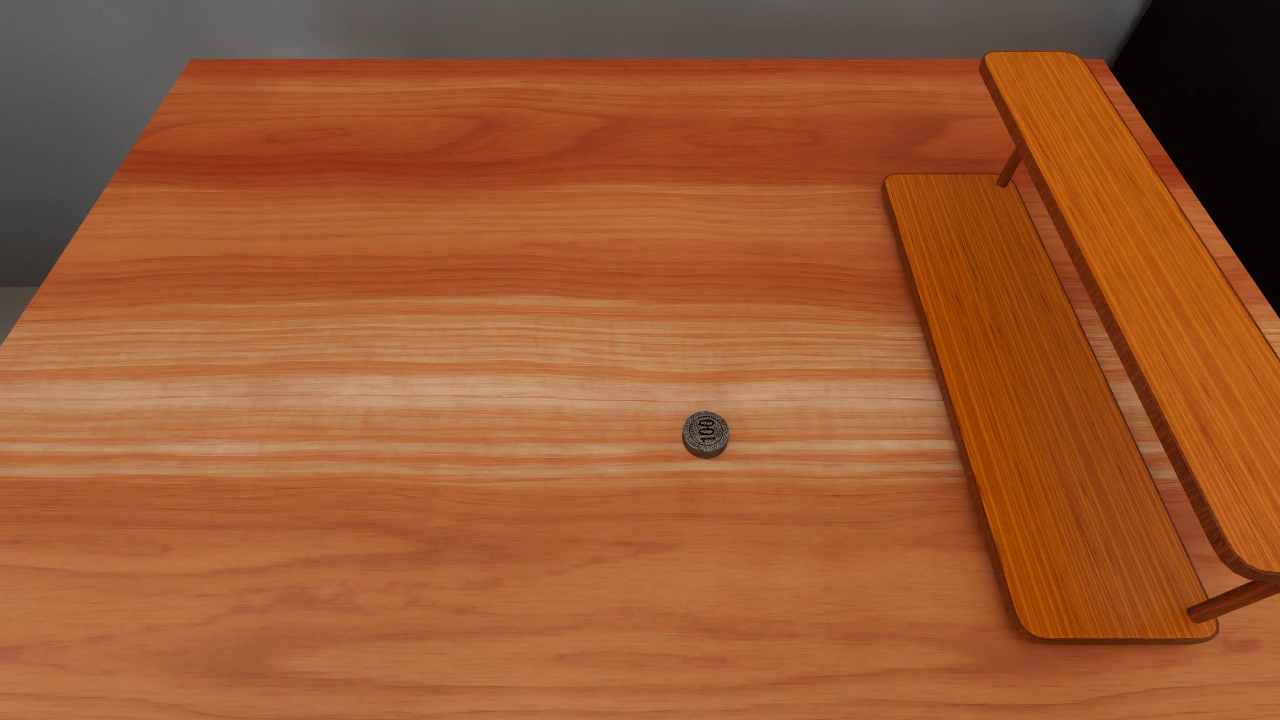}
  {Pick up the coin with one hand, transfer it to the other hand, place it on the coin rack, and return both arms home.}
  {ID and OOD use the same coin and rack, whose initial position may be left, right, or center. OOD changes only the visual environment.}
  {ArtiXon Arm-6A}
  {Teleop \textperiodcentered{} 300 demos}

\medskip
\noindent{\itshape Tool Usage}\par

\XTaskEntry
  {Nut on Screw}
  {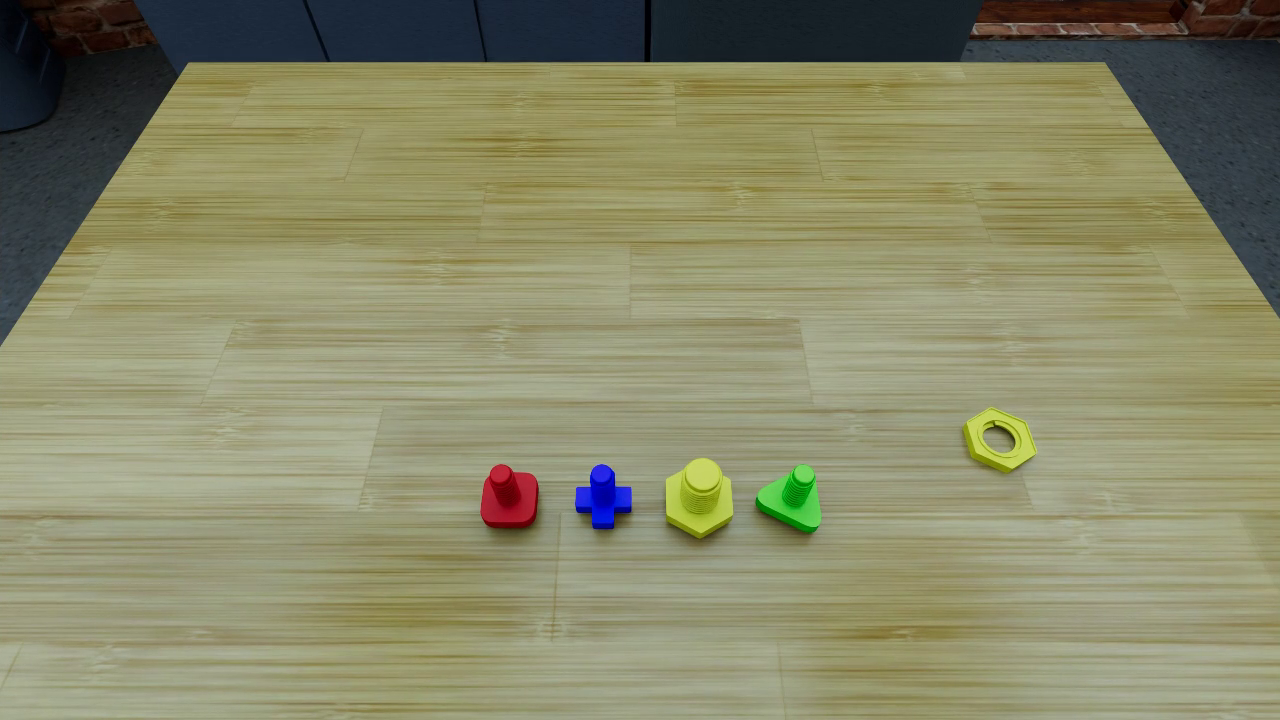}
  {Pick up the instructed nut, place it on the screw with matching color and shape, twist it into place, and return the arm home.}
  {ID and OOD use the same four nut--screw pairs and matching rule. OOD changes only the visual environment.}
  {ArtiXon Arm-6A, ARX R5}
  {Teleop \textperiodcentered{} 300 demos}

\XTaskEntry
  {Sweep Trash}
  {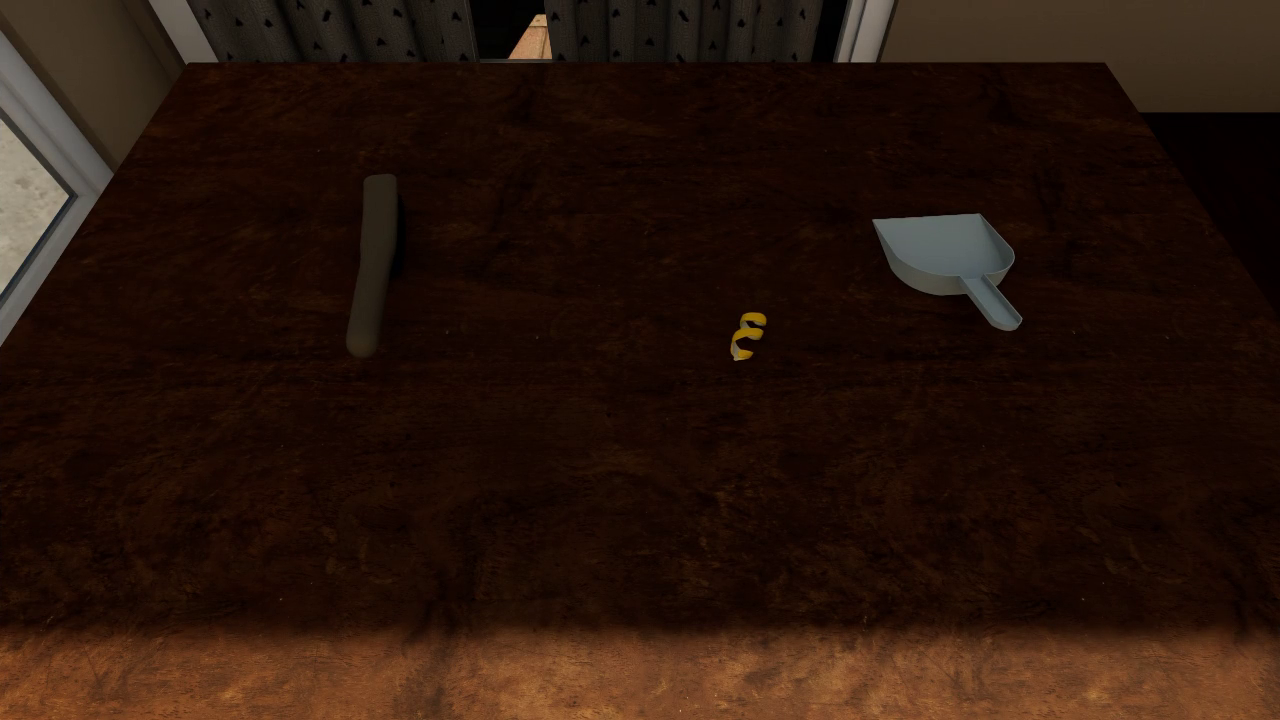}
  {Hold the dustpan with one hand and the broom with the other, sweep the trash into the dustpan, and return both arms home.}
  {ID and OOD use the same broom, dustpan, and trash pool. OOD changes only the visual environment.}
  {ArtiXon Arm-6A}
  {Teleop \textperiodcentered{} 300 demos}

\XTaskEntry
  {Whack-a-Mole}
  {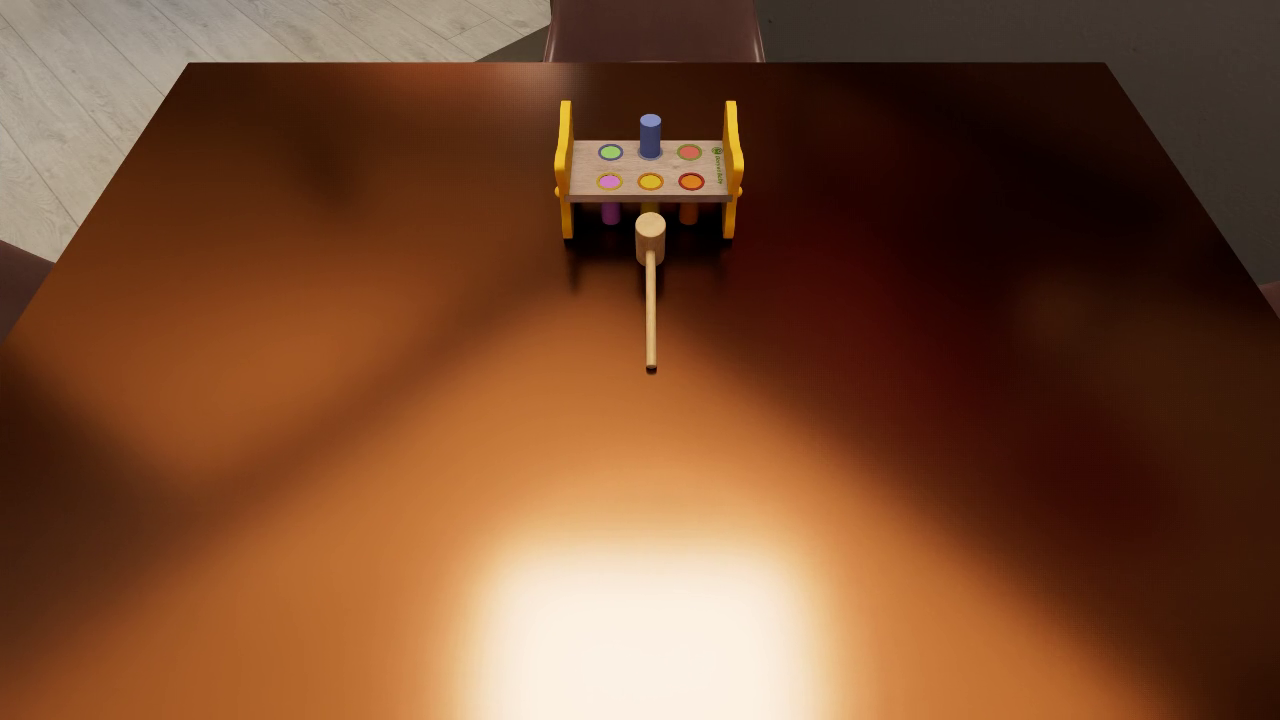}
  {Pick up the hammer, strike the instructed colored peg below the panel, return the hammer to the table, and return the arm home.}
  {ID and OOD use the same hammer, panel, and six colored pegs. OOD changes only the visual environment.}
  {ArtiXon Arm-6A}
  {Teleop \textperiodcentered{} 300 demos}

\medskip
\noindent{\itshape Dynamic Operation}\par

\XTaskEntry
  {Pick on Turntable}
  {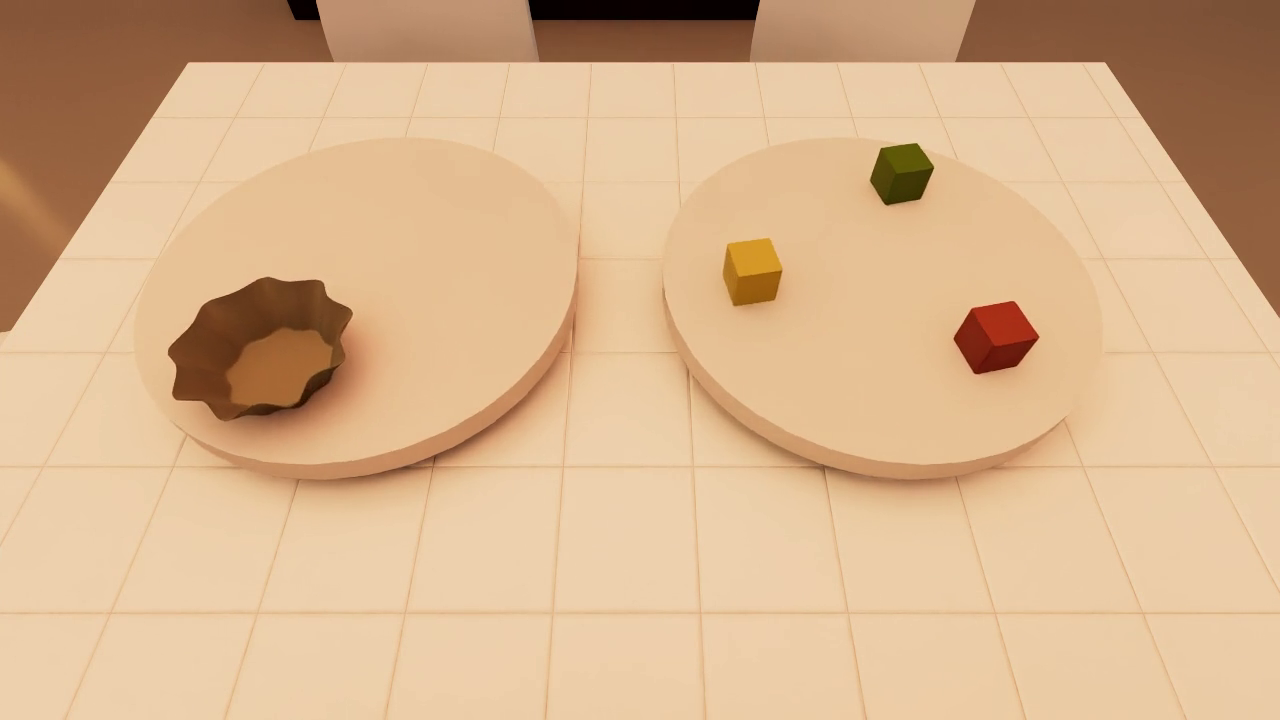}
  {Pick the instructed colored block from the rotating turntable, place it on the rotating plate, and return the arm to its home configuration.}
  {ID uses moderate turntable speeds. OOD introduces unseen slower or faster speeds together with visual changes while retaining the same blocks and plate.}
  {ArtiXon Arm-6A}
  {Teleop \textperiodcentered{} 300 demos}

\medskip
\noindent{\itshape Mobile Manipulation Tasks}\par

\XTaskEntry
  {Move Bottle to Bin}
  {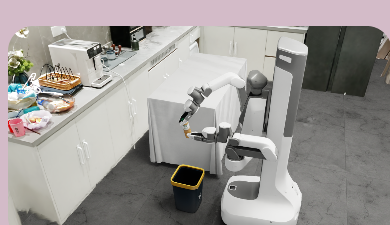}
  {Navigate to the table, grasp the bottle, carry it to the bin, place it inside, and return both arms to their home configuration.}
  {ID and OOD use the same bottle pool and bin. OOD changes only the visual environment.}
  {Quanta X1}
  {Auto \textperiodcentered{} 1,021 demos}

\XTaskEntry
  {Move Fruit to Basket}
  {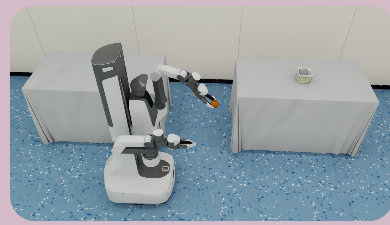}
  {Navigate to the fruit table, grasp the fruit, carry it to the second table, place it in the basket, and return both arms home.}
  {ID and OOD use the same fruit and basket pools. OOD changes only the visual environment.}
  {Quanta X1}
  {Auto \textperiodcentered{} 1,021 demos}

\XTaskEntry
  {Bottle on Shelf}
  {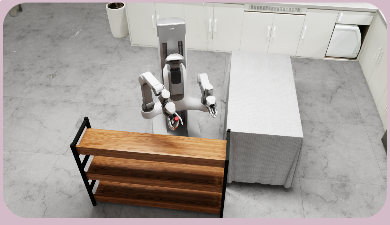}
  {Navigate to the table, grasp the bottle, carry it to the wooden shelf, place it upright on the shelf, and return both arms home.}
  {ID and OOD use the same bottle and shelf. OOD changes only the visual environment.}
  {Quanta X1}
  {Auto \textperiodcentered{} 1,027 demos}


\medskip
\noindent{\bfseries Challenge}\par
\smallskip

\XTaskEntry
  {Hanoi Tower}
  {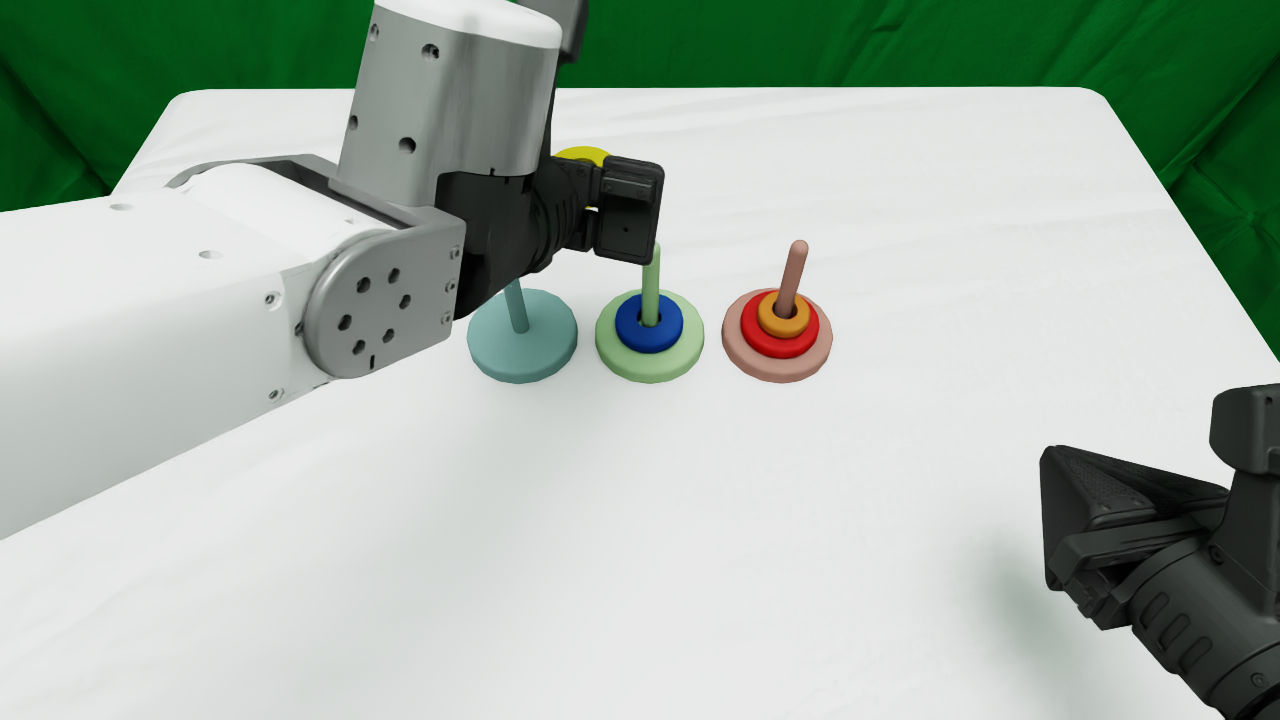}
  {Move two to five rings from the left peg to the right peg, one ring at a time, without placing a larger ring on a smaller one.}
  {The task is evaluated separately with two, three, four, or five rings and requires the prescribed legal intermediate states. It currently has no OOD evaluation.}
  {ArtiXon Arm-6A}
  {Auto \textperiodcentered{} 45 demos}

\XTaskEntry
  {Press Buttons (Hard)}
  {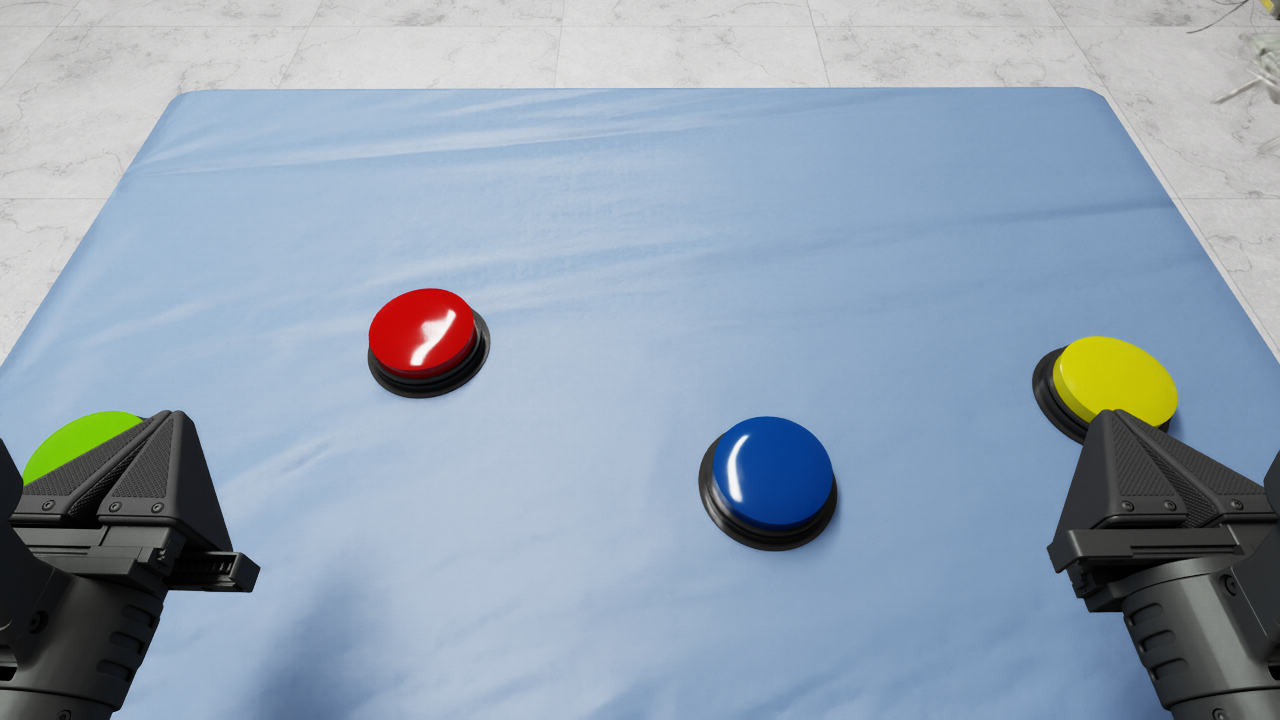}
  {Press the colored buttons in the instructed sequence, releasing between repeated presses, then return the arm to its home configuration.}
  {Sequence lengths of 4, 8, 12, 16, and 20 are evaluated separately over the same four buttons. This challenge currently has no OOD evaluation.}
  {ArtiXon Arm-6A}
  {Auto \textperiodcentered{} 4,000 demos}

\XTaskEntry
  {Stack Blocks (Hard)}
  {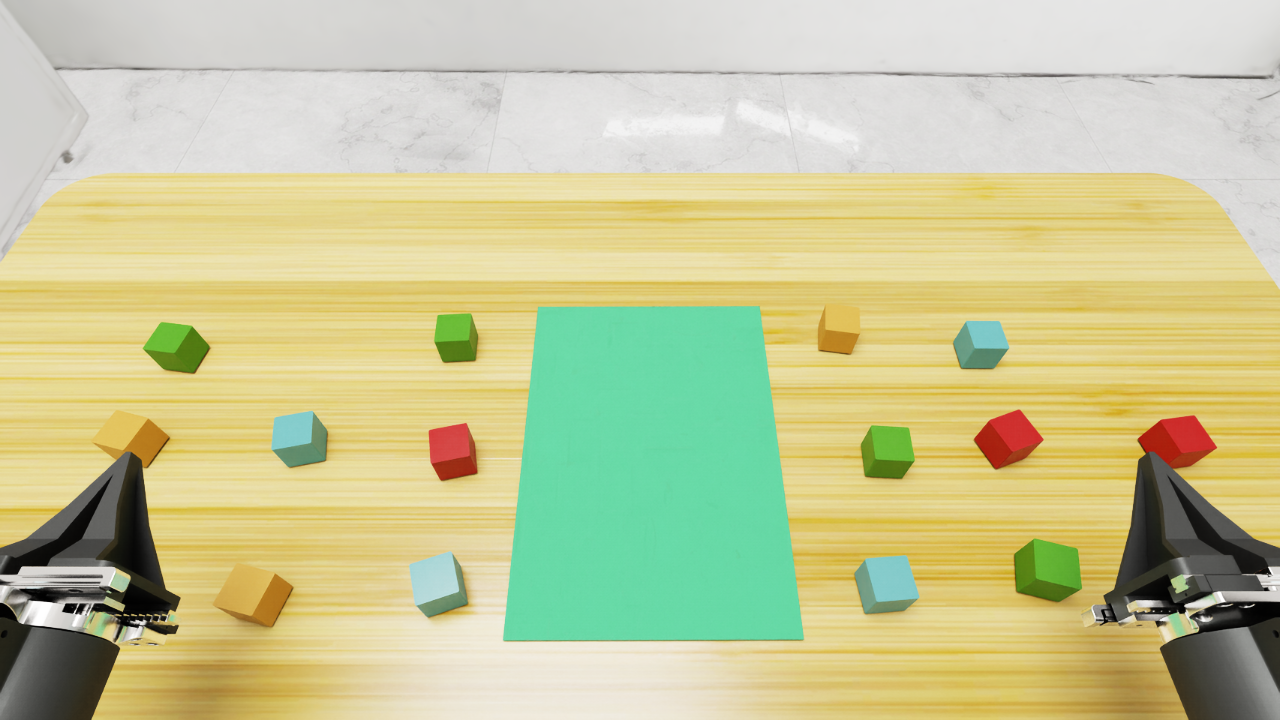}
  {Build a 12-layer block tower on the green target region and return both arms to their home configuration.}
  {Fifteen colored blocks are randomized around the target region. The task records stable completion of each layer and currently has no OOD evaluation.}
  {ArtiXon Arm-6A}
  {Auto + Teleop \textperiodcentered{} 14,600 stages + 1,398 demos}

%% file: sec/assets_scene/appendix_content.tex

\input{sec/assets_scene/appendices/10_asset_details}
\input{sec/assets_scene/appendices/20_scene_details}
\input{sec/assets_scene/appendices/30_runtime_replay_details}

%% file: sec/assets_scene/appendices/10_asset_details.tex
\subsection{Asset Preparation and AIG3D Implementation}
\label{app:asset-preparation-details}

\FloatBarrier
\begin{table}[!htbp]
  \caption{\textbf{Asset collection scope and provenance.}
  Counts use package-level units, so geometrically similar content from different
  sources may remain distinct packages.  AIG3D additions are included in the
  tabletop rigid row, while deformable assets form a separate collection outside
  the rigid and articulated totals.}
  \label{tab:tabletop-asset-statistics}
  \centering
  \footnotesize
  \setlength{\tabcolsep}{3.2pt}
  \renewcommand{\arraystretch}{1.12}
  \begin{tabularx}{\linewidth}{@{}
    >{\raggedright\arraybackslash}p{0.16\linewidth}
    >{\raggedleft\arraybackslash}p{0.08\linewidth}
    >{\raggedright\arraybackslash}p{0.30\linewidth}
    >{\raggedright\arraybackslash}X@{}}
    \toprule
    \textbf{Collection} & \textbf{Count} &
    \textbf{Collection role} & \textbf{Coverage and major sources} \\
    \midrule
    Indexed rigid catalog &
    \textbf{96,480} &
    Source records indexed for candidate retrieval &
    Tabletop objects, furniture, fixtures, beds, and major appliances. \\
    \addlinespace
    Tabletop rigid release &
    \textbf{12,485} &
    Packages prepared and accepted for tabletop task roles &
    44 leaf categories and 10 semantic families.  Major sources:
    Objaverse/CanoVerse (7,946), Omni6DPose (2,090), and
    BlenderKit (786). \\
    \addlinespace
    Articulated release &
    \textbf{670} &
    Packages prepared and accepted for articulated task roles &
    9 interaction groups.  HSSD-derived (607), task-specific additions (61), and
    ReplicaCAD (2). \\
    \bottomrule
  \end{tabularx}
\end{table}
\FloatBarrier

\subsubsection{Source, Geometry, and Appearance Implementation}

Format adapters convert GLB, Blender, USDZ, URDF, MJCF, and existing USD into a
common OpenUSD representation while preserving recoverable hierarchy,
transforms, geometry, texture and material bindings, source unit and axis
evidence, and link--joint topology when present.

Geometry processing applies observe--canonicalize--re-observe at per-mesh and
merged-object levels.  It records hierarchy and transforms, bakes child-node
transforms into a stable object frame, maps axes to canonical $z$-up, and converts
reliable source units to meters.  Checks cover metric extent, native and
triangulated complexity, degenerate elements, open boundaries, non-manifold
connectivity, thin structures, disconnected fragments, normals, and winding.
Transform, orientation, topology, or resolution repairs are followed by
world-space rechecks of bounds, origin, orientation, extent, and component
structure.
Temporary bounding-box centering and scale normalization affect only the
annotation frame, not the metric asset or its size estimate.

Source material networks are normalized by optical behavior: standard opaque
materials use OmniPBR's metallic--roughness representation, transmissive
materials use OmniGlass, and coated materials use a clearcoat model.  The
mapping reconciles diffuse--specular and base-color parameterizations,
roughness/metallic ranges, OpenGL/DirectX normal conventions, packed
occlusion--roughness--metallic channels, color spaces, UV transforms, alpha
masking, and transmission.  Trustworthy textures and scalar parameters are
preserved, and material bindings are checked together with rendered whole-object
appearance.

Generative PBR recovery uses available image, text, and geometry cues when
curated candidates are insufficient
\cite{xin2025dreampbr,hunyuan3d2025hunyuan3d21}, and rendered candidates are
evaluated as complete objects.  A Hunyuan3D-family model supports the single-
and multiview reconstruction branch \cite{zhao2025hunyuan3d2}; its conversion
path covers simulation-oriented remeshing, retopology, and PBR regeneration on
the final topology.

\subsubsection{Semantic, Physical, and Acceptance Implementation}

\noindent\textbf{Object-level semantic and metric grounding.}
Four standardized RGB views span azimuth and modest elevation variation in the
normalized annotation frame under neutral background and lighting.  We use a
Gemini~3.1-family multimodal model to produce the instance description and
retrieval record \cite{googledeepmind2026gemini31}; controlled vocabularies
support role matching, while free-form text preserves long-tail distinctions.

Semantic orientation follows CanoVerse's normalized-pose formulation
\cite{jin2026canoverse}.  Metadata, category context, and normalized views define
discrete rotations relative to a category reference pose; reliable source
orientation narrows the hypothesis set.  When source scale is missing or
untrustworthy, category- and function-matched anchors from GSO, ABO, and YCB
\cite{downs2022google,collins2022abo,calli2017ycb} condition relative-scale
estimates.  Category ranges and anchor--target co-renderings validate
relative-scale estimates; inconsistent cases are re-estimated or reviewed.

\noindent\textbf{Part and task-conditioned interaction grounding.}
PartSAM generates candidate 3D regions from surface prompts
\cite{zhu2025partsam}.  Confidence filtering and non-maximum suppression remove
weak or redundant proposals; ShapeNet-Part, 3DCoMPaT, and PartNet provide
reference part taxonomies and examples
\cite{yi2016shapenetpart,li20223dcompat,mo2019partnet}.  Whole-object views,
projected masks, local crops, the instance description, and part priors condition
cross-view labeling.  Equivalent parts are merged and reconciled using 3D
containment and relative order along front--back and vertical axes, with
selective review for residual ambiguity.  Candidate tasks condition final
interaction regions, and unsupported task--part assignments are left unassigned.

\noindent\textbf{Physical, kinematic, and contact-material authoring.}
Collision proxies progress from analytic boxes, spheres, and capsules to a
convex hull and then V-HACD/CoACD decomposition
\cite{mamou2009approximate,wei2022coacd}.  Candidate proxies are compared with
render geometry and checked for valid collider composition.  Mass and inertia
use watertight-mesh volume when supported, convex-hull volume next, and oriented
or axis-aligned bounds as conservative fallbacks; reliable source inertials are
preserved.

For articulated assets, link--joint topology retained from URDF, MJCF, and USD is
normalized into one object frame.  Joint frames, signed axes, angular and linear
limits, link collision shapes, inertial properties, and drives are authored
together.

Multiview appearance, PBR evidence, category, and part semantics determine a
coarse physical-material class for each relevant part.  Texture-region statistics
are used when maps exist, with authored constants as fallback.  The class selects
simulator-specific material-pair priors, including static and dynamic friction
relative to the contact counterpart.  Records are attached to corresponding
collision parts, and low-confidence parameters remain bounded for downstream
physical randomization.

\noindent\textbf{Role-specific simulator acceptance.}
Common prechecks cover metric size, render--collision alignment, collider
validity, and settling behavior.  Manipulable rigid objects are grasped, lifted
and held, transported, released, and observed through landing and post-placement
stabilization; checks include residual penetration, pose and velocity stability,
grasp retention, release separation, and support contact.  Articulated objects
and fixtures additionally exercise authored joints against declared axes and
limits and inspect collision behavior and task-relevant contacts.  Failed cases
return to targeted geometry, collision, material, scale, or kinematic repair and
repeat the applicable suite;
unresolved cases are excluded.

%% file: sec/assets_scene/appendices/20_scene_details.tex

%% file: sec/assets_scene/appendices/30_runtime_replay_details.tex
\subsection{Runtime Sampling and Replay Implementation Details}
\label{app:runtime-replay-details}

\subsubsection{Candidate Generation, Coupling, and Selection}

\paragraph{Candidate generation and geometric signatures.}
For each shared support region, the compiler constructs candidates with
four complementary heuristic families: grid-family enumeration, free-rectangle
search, shelf packing, and guillotine partitioning.  Strip and point placements
serve as fallbacks for domains not covered by the primary families.  Every
proposed cell is clipped against the footprint-aware feasible
domain and checked against support boundaries, polygon holes, obstacles, and
pairwise spacing.  Exact geometric duplicates are removed with a deterministic
signature comprising the sorted object names and each cell's four planar bounds,
rounded to $10^{-4}$ m.

\paragraph{Quality and diversity selection.}
Candidate quality combines total and minimum cell area, minimum short-edge
length, mean aspect score, minimum pairwise and obstacle gaps, and penalties for
fallback placements.  Selection begins with the highest-quality case, removes
candidates below a quality floor, and fills the remaining pool by
balancing quality against geometric distance from already selected cases.  The
distance normalizes corresponding cell-center displacements by the support
diagonal and also accounts for layout-partition topology, preventing near-identical
layouts from dominating the retained pool.

\paragraph{Overlap coupling.}
At runtime-plan construction, placement scopes are grouped by support surface
and coordinate frame.  Two scopes with disjoint object sets are connected when
their footprint-expanded placement bounds overlap.  Each nontrivial connected
component is compiled into one joint case pool, so a reset selects a compatible
assignment for all coupled objects rather than sampling them independently.
Scopes outside these components retain their existing local domains.

\paragraph{Reset-time plan interpretation.}
The \texttt{RuntimeSamplingPlan} contains coupled cases,
object-local domains, owner-first dependencies, and linked asset,
paired-material, and lighting variants.  At reset, the runtime
selects a coupled case, samples dependent local poses, applies compatible
variants, and instantiates the episode context.

\paragraph{Paired material variants.}
Each material-class variant couples a rendering-material template with
an authored physical-material record and the bodies, collision parts, or links
to which it applies.  Only complete, compatible rendering--physical pairs are
available for reset-time sampling.  The runtime applies each pair as one update:
dynamic bodies receive
density-consistent mass and inertia where applicable, affected contact parts
receive the associated material-pair friction parameters, and static or
kinematic supports update their applicable contact properties.
Identity-preserving appearance recipes remain visual and leave the
physical-material record unchanged.

\subsubsection{Replay Envelope and Stable-placement Details}

\paragraph{Motion proxies and temporal refinement.}
Replay reconstructs robot-link motion by forward kinematics and follows task
objects with their recorded rigid transforms.  Dynamic bodies are approximated
by sphere proxies.  Each trajectory interval is probed at its quarter, midpoint,
and three-quarter times and recursively subdivided until proxy deviation, joint
motion, root translation, and root rotation remain within tolerance.
The retained proxy paths are compressed with an error-bounded
Douglas--Peucker procedure \cite{douglas1973algorithms}.  Each line segment is
then converted to a capsule whose radius includes the proxy radius,
interpolation and simplification tolerances, and a clearance margin.

\paragraph{Stable-placement priors.}
Candidate orientations and support offsets are estimated offline with isolated
PhysX drop--settle trials under small initial-orientation perturbations.  A pose
is retained when repeated trials show stable orientation, acceptable ground gap
and penetration, limited drift, and consistent outcomes.  Stable-orientation and
support-offset priors are recomputed when the referenced asset geometry changes;
assets without a valid prior use the aligned default pose.